\documentclass[10pt,twocolumn,letterpaper]{article}

\usepackage{iccv}              %

\makeatletter
\renewcommand\footnoterule{%
  \kern 4pt
  \hrule width .4\columnwidth height .35pt
  \kern 1pt}
\renewcommand{\footnotesize}{\@setfontsize\footnotesize{8}{9}}
\makeatother

\usepackage[accsupp]{axessibility}  %

\usepackage{makecell} %

\usepackage[table]{xcolor} %
\usepackage[x11names]{xcolor} %

\usepackage{multirow}
\usepackage[most]{tcolorbox}
\newcounter{obsboxcounter}
\renewcommand{\theobsboxcounter}{\thesection.\arabic{obsboxcounter}}

\newtcolorbox[use counter=obsboxcounter, number within=section]{obsbox}{%
  colback=gray!5!white,      %
  colframe=gray!75!black,    %
  boxrule=0.5pt,             %
  arc=4pt,                   %
  left=4pt, right=4pt, top=4pt, bottom=4pt, %
  fonttitle=\bfseries,       %
  coltitle=white,            %
  title=Observation~\theobsboxcounter %
}

\newcommand{\sttdr}{$S^{2}3DR$}

\newcommand\mypara[1]{\vspace{1mm}\noindent\textbf{#1}}

\newcommand\mysubpara[1]{\vspace{1mm}\textbf{#1}}

\DeclareMathOperator*{\argmin}{arg\,min} %
\DeclareMathOperator{\E}{\mathbb{E}}

\usepackage{pgfplots}
\usepackage{amsmath}
\usepackage{tikz}
\usepackage{soul}

\definecolor{iccvblue}{rgb}{0.21,0.49,0.74}
\usepackage[pagebackref,breaklinks,colorlinks,allcolors=iccvblue]{hyperref}

\def\paperID{13483}
\def\confName{ICCV}
\def\confYear{2025}

\title{Explaining Human Preferences via Metrics for Structured 3D Reconstruction}

\author{
Jack Langerman$^{1}$\footnotemark[1]\quad
Denys Rozumnyi$^{2,3}$\footnotemark[2]\quad
Yuzhong Huang$^{4}$\quad
Dmytro Mishkin$^{3,4}$\\
$^{1}$Independent Researcher, USA\quad
$^{2}$ETH Zurich, Switzerland\\
$^{3}$Faculty of Electrical Engineering, Czech Technical University in Prague, Czech Republic\quad
$^{4}$Hover Inc., USA\\
{\tt\small jack@jackml.com\quad
rozumden@gmail.com\quad
yuzhong.huang@hover.to\quad
dmytro.mishkin@hover.to}
}

\begin{document}

\makeatletter
\renewcommand{\thefootnote}{\fnsymbol{footnote}}
\makeatother

\maketitle

\footnotetext[1]{Now at Apple.}
\footnotetext[2]{Now at Meta.}

\setcounter{footnote}{0}
\makeatletter
\renewcommand{\thefootnote}{\arabic{footnote}}
\makeatother

\begin{abstract}
"What cannot be measured cannot be improved" while likely never uttered by Lord Kelvin, summarizes effectively the driving force behind this work. This paper presents a detailed discussion of automated metrics for evaluating structured 3D reconstructions. Pitfalls of each metric are discussed, and an analysis through the lens of expert 3D modelers' preferences is presented. A set of systematic "unit tests" are proposed to empirically verify desirable properties, and context aware recommendations regarding which metric to use depending on application are provided. Finally, a learned metric distilled from human expert judgments is proposed and analyzed. The source code is available at ~\url{https://github.com/s23dr/wireframe-metrics-iccv2025}.
\end{abstract}
\vspace{-1em}

\section{Introduction}
\label{sec:intro}
Benchmarks have been key drivers of progress in computer vision; the canonical example is certainly ImageNet~\cite{ImageNet}, but prominent examples abound beyond image classification: object tracking~\cite{VOT, MOT}, image retrieval~\cite{GLD, rOxford5k}, image matching~\cite{IMC}, 6D pose estimation~\cite{BOP}, optical flow~\cite{KITTY}, \etc. Benchmarks have three main components -- the data, the protocol, and the metrics. 
While the data is the single most important component, progress is hard without being able to answer the question, "progress on what?" Good metrics are the quantitative answer to this question. While metrics do not need to be perfect, their gradient should point progress in the right direction.
\begin{figure}
    \resizebox{0.23\linewidth}{!}{
    \begin{tikzpicture}[scale=1]
\node[align=center, font=\Huge] at (5.000, 10.500) {GT};
\coordinate (V0) at (2.750, 0.500);
\filldraw[fill=gray] (V0) circle [radius=0.1];
\coordinate (V1) at (7.250, 0.500);
\filldraw[fill=gray] (V1) circle [radius=0.1];
\coordinate (V2) at (7.250, 9.500);
\filldraw[fill=gray] (V2) circle [radius=0.1];
\coordinate (V3) at (2.750, 9.500);
\filldraw[fill=gray] (V3) circle [radius=0.1];
\coordinate (V4) at (5.000, 2.300);
\filldraw[fill=gray] (V4) circle [radius=0.1];
\coordinate (V5) at (5.000, 7.700);
\filldraw[fill=gray] (V5) circle [radius=0.1];
\draw[very thick, color=black] (V0) -- (V1);
\draw[very thick, color=black] (V1) -- (V2);
\draw[very thick, color=black] (V2) -- (V3);
\draw[very thick, color=black] (V3) -- (V0);
\draw[very thick, color=black] (V0) -- (V4);
\draw[very thick, color=black] (V1) -- (V4);
\draw[very thick, color=black] (V2) -- (V5);
\draw[very thick, color=black] (V3) -- (V5);
\draw[very thick, color=black] (V4) -- (V5);
\end{tikzpicture}
    }
    \resizebox{0.23\linewidth}{!}{
    \begin{tikzpicture}[scale=1]
\node[align=center, font=\Huge] at (5.000, 10.500) {WF1};
\coordinate (V0) at (2.750, 0.500);
\filldraw[fill=gray] (V0) circle [radius=0.1];
\coordinate (V1) at (7.250, 0.500);
\filldraw[fill=gray] (V1) circle [radius=0.1];
\coordinate (V2) at (7.250, 9.500);
\filldraw[fill=gray] (V2) circle [radius=0.1];
\coordinate (V3) at (2.750, 9.500);
\filldraw[fill=gray] (V3) circle [radius=0.1];
\coordinate (V4) at (5.000, 2.300);
\filldraw[fill=gray] (V4) circle [radius=0.1];
\coordinate (V5) at (5.000, 7.700);
\filldraw[fill=gray] (V5) circle [radius=0.1];
\coordinate (V6) at (2.752, 0.500);
\filldraw[fill=gray] (V6) circle [radius=0.1];
\coordinate (V7) at (2.750, 6.601);
\filldraw[fill=gray] (V7) circle [radius=0.1];
\coordinate (V8) at (2.750, 9.201);
\filldraw[fill=gray] (V8) circle [radius=0.1];
\coordinate (V9) at (2.886, 0.609);
\filldraw[fill=gray] (V9) circle [radius=0.1];
\coordinate (V10) at (7.250, 4.079);
\filldraw[fill=gray] (V10) circle [radius=0.1];
\coordinate (V11) at (7.250, 4.445);
\filldraw[fill=gray] (V11) circle [radius=0.1];
\coordinate (V12) at (6.772, 0.882);
\filldraw[fill=gray] (V12) circle [radius=0.1];
\coordinate (V13) at (3.418, 9.500);
\filldraw[fill=gray] (V13) circle [radius=0.1];
\coordinate (V14) at (6.211, 8.669);
\filldraw[fill=gray] (V14) circle [radius=0.1];
\coordinate (V15) at (5.813, 8.351);
\filldraw[fill=gray] (V15) circle [radius=0.1];
\coordinate (V16) at (3.310, 9.052);
\filldraw[fill=gray] (V16) circle [radius=0.1];
\coordinate (V17) at (3.346, 9.023);
\filldraw[fill=gray] (V17) circle [radius=0.1];
\coordinate (V18) at (4.593, 8.026);
\filldraw[fill=gray] (V18) circle [radius=0.1];
\coordinate (V19) at (5.000, 3.815);
\filldraw[fill=gray] (V19) circle [radius=0.1];
\draw[very thick, color=black] (V0) -- (V6);
\draw[very thick, color=black] (V0) -- (V7);
\draw[very thick, color=black] (V0) -- (V9);
\draw[very thick, color=black] (V1) -- (V6);
\draw[very thick, color=black] (V1) -- (V10);
\draw[very thick, color=black] (V1) -- (V12);
\draw[very thick, color=black] (V2) -- (V11);
\draw[very thick, color=black] (V2) -- (V13);
\draw[very thick, color=black] (V2) -- (V14);
\draw[very thick, color=black] (V3) -- (V8);
\draw[very thick, color=black] (V3) -- (V13);
\draw[very thick, color=black] (V3) -- (V16);
\draw[very thick, color=black] (V4) -- (V9);
\draw[very thick, color=black] (V4) -- (V12);
\draw[very thick, color=black] (V4) -- (V19);
\draw[very thick, color=black] (V5) -- (V15);
\draw[very thick, color=black] (V5) -- (V18);
\draw[very thick, color=black] (V5) -- (V19);
\draw[very thick, color=black] (V7) -- (V8);
\draw[very thick, color=black] (V10) -- (V11);
\draw[very thick, color=black] (V14) -- (V15);
\draw[very thick, color=black] (V16) -- (V17);
\draw[very thick, color=black] (V17) -- (V18);
\end{tikzpicture}
    }
    \resizebox{0.23\linewidth}{!}{
    \begin{tikzpicture}[scale=1]
\node[align=center, font=\Huge] at (5.000, 10.500) {WF2};
\coordinate (V0) at (2.750, 0.500);
\filldraw[fill=gray] (V0) circle [radius=0.1];
\coordinate (V1) at (7.250, 0.500);
\filldraw[fill=gray] (V1) circle [radius=0.1];
\coordinate (V2) at (7.250, 9.500);
\filldraw[fill=gray] (V2) circle [radius=0.1];
\coordinate (V3) at (2.750, 9.500);
\filldraw[fill=gray] (V3) circle [radius=0.1];
\coordinate (V4) at (5.000, 7.700);
\filldraw[fill=gray] (V4) circle [radius=0.1];
\draw[very thick, color=black] (V0) -- (V1);
\draw[very thick, color=black] (V1) -- (V2);
\draw[very thick, color=black] (V2) -- (V3);
\draw[very thick, color=black] (V3) -- (V0);
\draw[very thick, color=black] (V4) -- (V3);
\end{tikzpicture}
    }
\resizebox{0.245\linewidth}{!}{
    \begin{tikzpicture}[scale=1]
\node[align=center, font=\Huge] at (5.000, 9.9) {WF3};
\coordinate (V0) at (5.000, 2.975);
\filldraw[fill=gray] (V0) circle [radius=0.1];
\coordinate (V1) at (2.750, 0.500);
\filldraw[fill=gray] (V1) circle [radius=0.1];
\coordinate (V2) at (7.250, 2.750);
\filldraw[fill=gray] (V2) circle [radius=0.1];
\coordinate (V3) at (5.000, 2.750);
\filldraw[fill=gray] (V3) circle [radius=0.1];
\coordinate (V4) at (5.000, 7.500);
\filldraw[fill=gray] (V4) circle [radius=0.1];
\draw[very thick, color=black] (V3) -- (V4);
\end{tikzpicture}
    } \\
\vspace{0.5em}
\footnotesize
\begin{tabular}{lcrrr}
\toprule
Metric/wireframe &  GT & WF1 & WF2 & WF3  \\ 
\cmidrule(r){1-5}
 Vertex F1 $\uparrow$  & 1.00 & \textcolor{red}{0.56} & 0.91 & 0.18 \\
 Edge F1 $\uparrow$  & 1.00 & \textcolor{red}{0.19} & 0.71 & 0.00 \\
 Jaccard Distance $\downarrow$  & 0.00 & 0.00 & 0.33 & 1.00 \\
 WED $\downarrow$  & 0.00 & \textcolor{red}{2.32} & 0.52 & 1.82 \\
 WED S23DR $\downarrow$  & 0.00 & \textcolor{red}{2.69} & 0.59 & 1.63 \\
 Graph Spectral $\downarrow$  & 0.00 & \textcolor{red}{377.93} & 577.49 & 1603.32 \\
\bottomrule
\end{tabular}
\vspace{-1em}
  \caption{A motivating example for this work.
  While humans tend to sort the wireframes from best to worst in the presented order, popular metrics (defined in Sec~\ref{sec:metrics}) sort them differently, sometimes completely inverting the order.
Top, left to right: GT -- ground truth wireframe, WF1 -- wireframe with edges split into several segments, maintaining geometrical and topological accuracy, WF2 -- wireframe with missing vertices and edges,  WF3 -- wireframe with only one correct vertex.
Bottom: distances between GT and respective wireframe. Numbers that change sorting are in \textcolor{red}{red}.}
\label{fig:enter-label}
\vspace{-2em}
\end{figure}
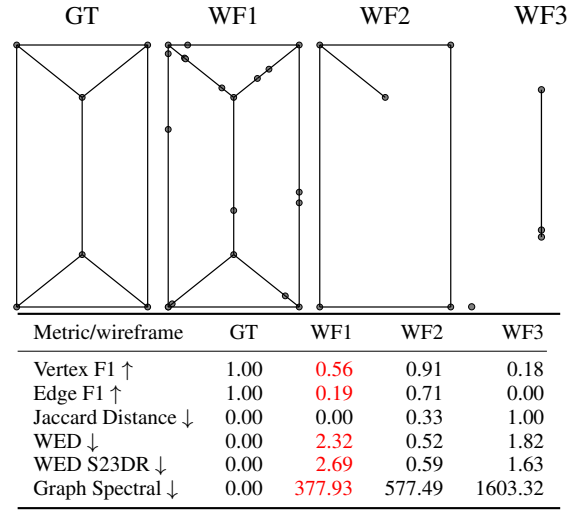%

We consider an area of structured and semi-structured reconstruction, which has recently gained popularity~\cite{USM3D2024}~\cite{AppleRoomPlan}~\cite{SceneScript2024}.
Given a set or sequence of sensory data, such as ground images~\cite{S23DR}, satellite images~\cite{HEAT}, or aerial LiDAR~\cite{Building3D}, the goal is to produce a wireframe or a CAD model of a building or other structure. 
The modeling output is presented as a spatial graph with vertices (such as roof apex point, \etc) and edges (ridge line, \etc). Several datasets have been recently proposed~\cite{Building3D, S23DR} for the task.
The issue is that hardly a pair of publications in the area use the same metric to evaluate quantitative results. 
Some use recognition metrics such as precision and recall on vertices and edges~\cite{luo2022LC2WF} or more enhanced versions such as Structured Average Precision~\cite{zhou2019end, ma2022HoW3D}. 
Others opt for graph-based metrics, such as the Wireframe Edit Distance (WED)~\cite{liu2021_pc2wf, WireframeNet, luo2022LC2WF}. 
Finally, some methods treat the problem similarly to point cloud registration and report Chamfer Distance (CD)~\cite{Huang_2024_CVPR, SepicNet, APC2Mesh}.
Other related fields, like structure from motion, use downstream metrics, such as image generation quality~\cite{acezero2024} and camera pose accuracy~\cite{IMC}.
One aspect of the difficulty is related to the fact that structured reconstructions often have different goals. 
For instance, one purpose of the reconstructed wireframes is to represent building plans and answer questions such as "what is the area of the bedroom?" 
Within this formulation, a black-box model, \eg visual-and-language model (VLM), which takes an image as an input and outputs a correct estimate, would be a perfect match. 
On the other hand, a floorplan or a blueprint has value for record-keeping, planning, and other applications that cannot be easily replaced with a black-box model.

Finally, many existing metrics, while useful, often fail to deliver value in practice when comparing two imperfect estimations, and designing metrics that effectively compare a pair of "very good" and a pair of "very bad" solutions at the same time is also non-trivial.

Moreover, there is evidence~\cite{S23DR} that some metrics can be "hacked" or exploited in such a way that obviously bad solutions have better scores than flawed but ultimately quite reasonable solutions. Such examples include a number of corner cases, which can cause existing metrics to become useless in practical scenarios. For instance, in the case where long edges are split into smaller, yet perfectly collinear, pieces, \eg Fig.~\ref{fig:enter-label}, most commonly used metrics, such as edge F1, vertex F1, and Wireframe Edit Distance, fail completely and prefer much worse solutions.

This paper makes the following contributions:
\begin{enumerate}[itemsep=0.0pt,topsep=3pt,leftmargin=*,label=\textbf{(\arabic*)}]
\item Measure the perceived quality of reconstructions by human domain experts, and infer a global ranking of all structured reconstructions. This ranking is then compared to the ranking given by all metrics.
\item Show how well existing metrics agree with human preferences and how much they correlate with each other.
\item Propose a set of "unit-tests" for testing the properties of the metrics.
\item Introduce a simple learned metric, which correlates well with human judgment.
\item Make recommendations of which metrics to use depending on the use case.
\end{enumerate}

\section{What do we actually need from wireframe comparison metrics?}
\label{sec:what-we-need}
We consider semi-automated 3D modeling as a target task with enough generality to drive progress that can smoothly scale to fully-automated modeling, but is still tractable and reliable for commercial applications today.

For this task, the wireframe representation is estimated from some "raw" inputs such as images or a LiDAR point cloud and then transferred to human experts to (1) approve it, (2) correct it and then approve, (3) reject and create the model from scratch manually. 

We argue that such task formulation creates an implicit usefulness ranking over reconstructions (and thereby over reconstruction methods). %
Otherwise, without considering human involvement, everything becomes binary -- either the model is good enough to be used (\eg for 3D printing, measurement extraction), or it is not.

\begin{figure}[tb]
    \centering
    \includegraphics[width=0.95\linewidth]{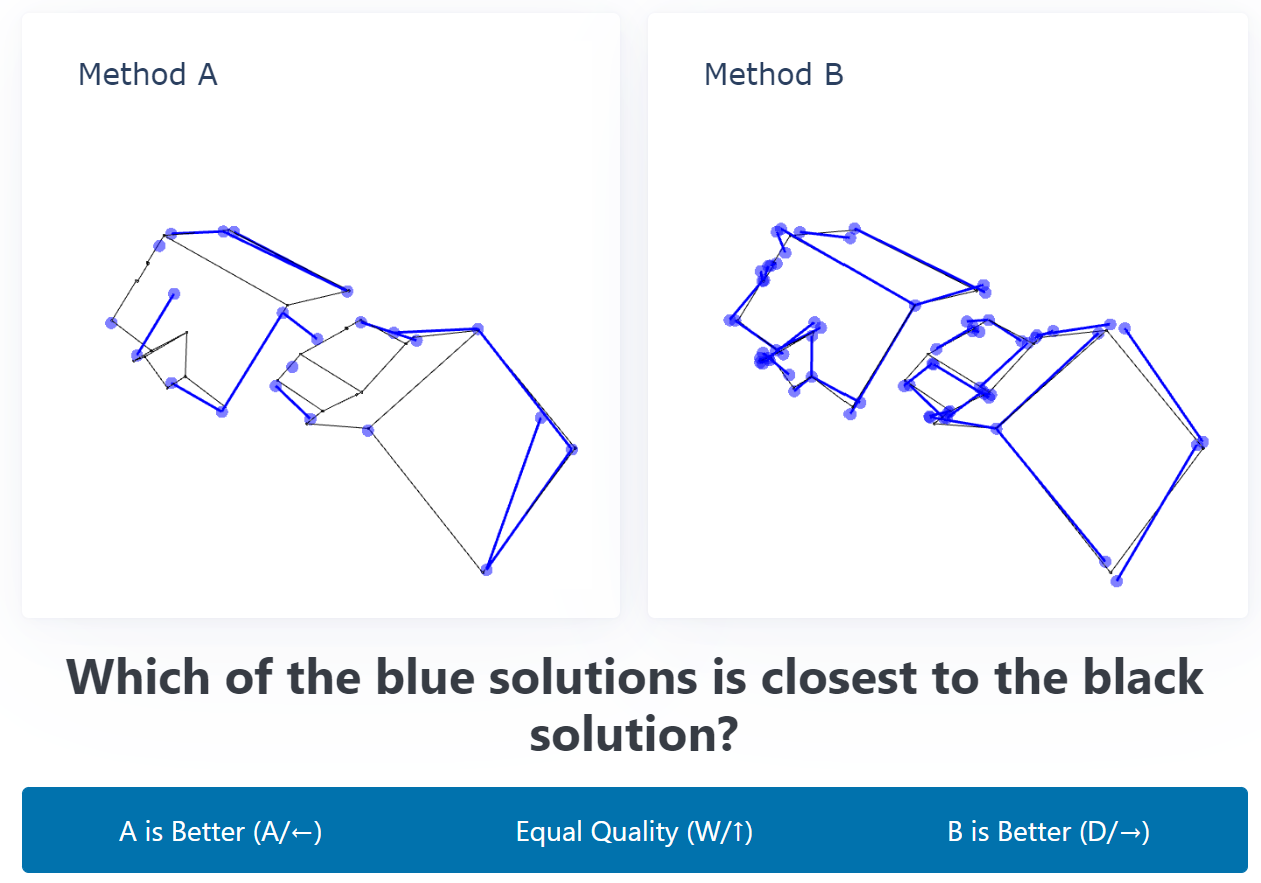}
    \caption{Wireframe ranking interface for human annotators. }
    \label{fig:ranking-app}
\vspace{-1.5em}
\end{figure}

For this reason, we consider the following experimental setup to benchmark wireframe comparison metrics.
A pool of professional 3D modelers, whose everyday job is creating CAD-like models from raw data such as images, is asked to rank pairs of wireframes.
An example of the ranking setup is shown in Fig.~\ref{fig:ranking-app}. All wireframes are superimposed with ground truth models, and annotators are provided tools allowing them to translate, scale, and rotate the wireframes in 3D. 

The wireframes are drawn from two pools, as described below. 
We then evaluate how the existing metrics agree with the judgment of professional human 3D modeling experts.

\mypara{Pool1-- \sttdr.} We acquire a representative set of $S^{2}3DR$ challenge entries~\cite{S23DR} as well as a PC2WF~\cite{liu2021_pc2wf} baseline. 
These wireframes were algorithmically (and with the help of deep-learning models) reconstructed from multiview inputs with the goal of minimizing a variant of WED. 
We include the top-10 entries with team names used as identifiers.  
The ground truth models were created by human experts and have undergone significant validation. 
The input data were captured by users on mobile phones in North America.

\begin{figure}[tb]
    \centering
        \resizebox{0.19\linewidth}{!}{
    \begin{tikzpicture}[scale=1]
\coordinate (V0) at (0.500, 4.605);
\filldraw[fill=gray] (V0) circle [radius=0.1];
\coordinate (V1) at (1.680, 4.219);
\filldraw[fill=gray] (V1) circle [radius=0.1];
\coordinate (V2) at (3.749, 4.811);
\filldraw[fill=gray] (V2) circle [radius=0.1];
\coordinate (V3) at (5.067, 3.111);
\filldraw[fill=gray] (V3) circle [radius=0.1];
\coordinate (V4) at (6.963, 2.491);
\filldraw[fill=gray] (V4) circle [radius=0.1];
\coordinate (V5) at (7.487, 4.092);
\filldraw[fill=gray] (V5) circle [radius=0.1];
\coordinate (V6) at (1.024, 6.207);
\filldraw[fill=gray] (V6) circle [radius=0.1];
\coordinate (V7) at (4.862, 2.483);
\filldraw[fill=gray] (V7) circle [radius=0.1];
\coordinate (V8) at (3.168, 3.037);
\filldraw[fill=gray] (V8) circle [radius=0.1];
\coordinate (V9) at (1.474, 3.591);
\filldraw[fill=gray] (V9) circle [radius=0.1];
\coordinate (V10) at (8.011, 5.694);
\filldraw[fill=gray] (V10) circle [radius=0.1];
\coordinate (V11) at (6.985, 6.030);
\filldraw[fill=gray] (V11) circle [radius=0.1];
\coordinate (V12) at (4.661, 5.415);
\filldraw[fill=gray] (V12) circle [radius=0.1];
\coordinate (V13) at (3.149, 7.285);
\filldraw[fill=gray] (V13) circle [radius=0.1];
\coordinate (V14) at (1.548, 7.809);
\filldraw[fill=gray] (V14) circle [radius=0.1];
\coordinate (V15) at (2.685, 7.099);
\filldraw[fill=gray] (V15) circle [radius=0.1];
\coordinate (V16) at (2.730, 7.235);
\filldraw[fill=gray] (V16) circle [radius=0.1];
\coordinate (V17) at (2.894, 7.181);
\filldraw[fill=gray] (V17) circle [radius=0.1];
\coordinate (V18) at (2.850, 7.045);
\filldraw[fill=gray] (V18) circle [radius=0.1];
\coordinate (V19) at (3.401, 8.055);
\filldraw[fill=gray] (V19) circle [radius=0.1];
\coordinate (V20) at (5.319, 7.427);
\filldraw[fill=gray] (V20) circle [radius=0.1];
\coordinate (V21) at (7.236, 6.800);
\filldraw[fill=gray] (V21) circle [radius=0.1];
\coordinate (V22) at (7.360, 4.076);
\filldraw[fill=gray] (V22) circle [radius=0.1];
\coordinate (V23) at (6.894, 2.654);
\filldraw[fill=gray] (V23) circle [radius=0.1];
\coordinate (V24) at (7.002, 2.609);
\filldraw[fill=gray] (V24) circle [radius=0.1];
\coordinate (V25) at (9.032, 1.945);
\filldraw[fill=gray] (V25) circle [radius=0.1];
\coordinate (V26) at (9.500, 3.376);
\filldraw[fill=gray] (V26) circle [radius=0.1];
\draw[very thick, color=black] (V0) -- (V1);
\draw[very thick, color=black] (V1) -- (V2);
\draw[very thick, color=black] (V2) -- (V3);
\draw[very thick, color=black] (V3) -- (V4);
\draw[very thick, color=black] (V4) -- (V5);
\draw[very thick, color=black] (V5) -- (V6);
\draw[very thick, color=black] (V6) -- (V0);
\draw[very thick, color=black] (V7) -- (V3);
\draw[very thick, color=black] (V2) -- (V8);
\draw[very thick, color=black] (V8) -- (V7);
\draw[very thick, color=black] (V1) -- (V9);
\draw[very thick, color=black] (V9) -- (V8);
\draw[very thick, color=black] (V10) -- (V11);
\draw[very thick, color=black] (V11) -- (V12);
\draw[very thick, color=black] (V12) -- (V13);
\draw[very thick, color=black] (V13) -- (V14);
\draw[very thick, color=black] (V14) -- (V6);
\draw[very thick, color=black] (V5) -- (V10);
\draw[very thick, color=black] (V15) -- (V16);
\draw[very thick, color=black] (V16) -- (V17);
\draw[very thick, color=black] (V17) -- (V18);
\draw[very thick, color=black] (V18) -- (V15);
\draw[very thick, color=black] (V19) -- (V13);
\draw[very thick, color=black] (V12) -- (V20);
\draw[very thick, color=black] (V20) -- (V19);
\draw[very thick, color=black] (V11) -- (V21);
\draw[very thick, color=black] (V21) -- (V20);
\draw[very thick, color=black] (V22) -- (V23);
\draw[very thick, color=black] (V24) -- (V25);
\draw[very thick, color=black] (V25) -- (V26);
\draw[very thick, color=black] (V26) -- (V22);
\end{tikzpicture}
    }
            \resizebox{0.19\linewidth}{!}{
    \input{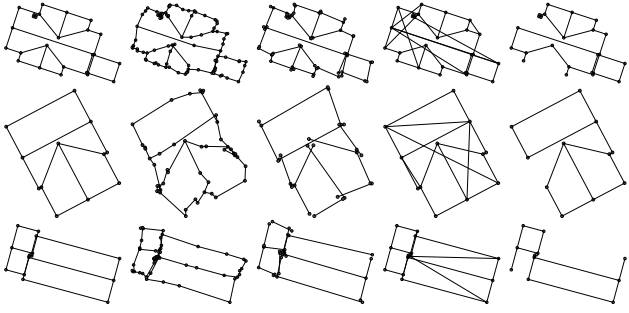}
    }
               \resizebox{0.19\linewidth}{!}{
    \begin{tikzpicture}[scale=1]
\coordinate (V0) at (1.469, 3.559);
\filldraw[fill=gray] (V0) circle [radius=0.1];
\coordinate (V1) at (6.990, 6.003);
\filldraw[fill=gray] (V1) circle [radius=0.1];
\coordinate (V2) at (2.682, 7.074);
\filldraw[fill=gray] (V2) circle [radius=0.1];
\coordinate (V3) at (1.451, 7.883);
\filldraw[fill=gray] (V3) circle [radius=0.1];
\coordinate (V4) at (1.678, 7.783);
\filldraw[fill=gray] (V4) circle [radius=0.1];
\coordinate (V5) at (3.132, 7.182);
\filldraw[fill=gray] (V5) circle [radius=0.1];
\coordinate (V6) at (3.283, 7.360);
\filldraw[fill=gray] (V6) circle [radius=0.1];
\coordinate (V7) at (3.152, 7.212);
\filldraw[fill=gray] (V7) circle [radius=0.1];
\coordinate (V8) at (5.418, 7.507);
\filldraw[fill=gray] (V8) circle [radius=0.1];
\coordinate (V9) at (5.297, 7.380);
\filldraw[fill=gray] (V9) circle [radius=0.1];
\coordinate (V10) at (5.433, 7.393);
\filldraw[fill=gray] (V10) circle [radius=0.1];
\coordinate (V11) at (3.395, 7.938);
\filldraw[fill=gray] (V11) circle [radius=0.1];
\coordinate (V12) at (3.327, 8.059);
\filldraw[fill=gray] (V12) circle [radius=0.1];
\coordinate (V13) at (1.541, 4.054);
\filldraw[fill=gray] (V13) circle [radius=0.1];
\coordinate (V14) at (1.651, 4.173);
\filldraw[fill=gray] (V14) circle [radius=0.1];
\coordinate (V15) at (1.589, 4.096);
\filldraw[fill=gray] (V15) circle [radius=0.1];
\coordinate (V16) at (7.331, 6.902);
\filldraw[fill=gray] (V16) circle [radius=0.1];
\coordinate (V17) at (7.211, 6.644);
\filldraw[fill=gray] (V17) circle [radius=0.1];
\coordinate (V18) at (4.956, 3.200);
\filldraw[fill=gray] (V18) circle [radius=0.1];
\coordinate (V19) at (5.201, 3.036);
\filldraw[fill=gray] (V19) circle [radius=0.1];
\coordinate (V20) at (5.086, 2.981);
\filldraw[fill=gray] (V20) circle [radius=0.1];
\coordinate (V21) at (7.989, 5.540);
\filldraw[fill=gray] (V21) circle [radius=0.1];
\coordinate (V22) at (8.018, 5.558);
\filldraw[fill=gray] (V22) circle [radius=0.1];
\coordinate (V23) at (7.475, 4.121);
\filldraw[fill=gray] (V23) circle [radius=0.1];
\coordinate (V24) at (7.522, 4.011);
\filldraw[fill=gray] (V24) circle [radius=0.1];
\coordinate (V25) at (7.533, 4.082);
\filldraw[fill=gray] (V25) circle [radius=0.1];
\coordinate (V26) at (7.097, 2.694);
\filldraw[fill=gray] (V26) circle [radius=0.1];
\coordinate (V27) at (7.433, 4.106);
\filldraw[fill=gray] (V27) circle [radius=0.1];
\coordinate (V28) at (7.335, 4.181);
\filldraw[fill=gray] (V28) circle [radius=0.1];
\coordinate (V29) at (1.157, 6.275);
\filldraw[fill=gray] (V29) circle [radius=0.1];
\coordinate (V30) at (0.991, 6.094);
\filldraw[fill=gray] (V30) circle [radius=0.1];
\coordinate (V31) at (0.904, 6.273);
\filldraw[fill=gray] (V31) circle [radius=0.1];
\coordinate (V32) at (3.838, 4.747);
\filldraw[fill=gray] (V32) circle [radius=0.1];
\coordinate (V33) at (3.869, 4.734);
\filldraw[fill=gray] (V33) circle [radius=0.1];
\coordinate (V34) at (3.739, 4.826);
\filldraw[fill=gray] (V34) circle [radius=0.1];
\coordinate (V35) at (9.383, 3.458);
\filldraw[fill=gray] (V35) circle [radius=0.1];
\coordinate (V36) at (9.500, 3.482);
\filldraw[fill=gray] (V36) circle [radius=0.1];
\coordinate (V37) at (2.794, 7.072);
\filldraw[fill=gray] (V37) circle [radius=0.1];
\coordinate (V38) at (2.745, 7.340);
\filldraw[fill=gray] (V38) circle [radius=0.1];
\coordinate (V39) at (2.964, 7.103);
\filldraw[fill=gray] (V39) circle [radius=0.1];
\coordinate (V40) at (2.759, 7.238);
\filldraw[fill=gray] (V40) circle [radius=0.1];
\coordinate (V41) at (6.845, 2.322);
\filldraw[fill=gray] (V41) circle [radius=0.1];
\coordinate (V42) at (7.086, 2.397);
\filldraw[fill=gray] (V42) circle [radius=0.1];
\coordinate (V43) at (4.868, 2.515);
\filldraw[fill=gray] (V43) circle [radius=0.1];
\coordinate (V44) at (4.812, 2.367);
\filldraw[fill=gray] (V44) circle [radius=0.1];
\coordinate (V45) at (2.954, 7.010);
\filldraw[fill=gray] (V45) circle [radius=0.1];
\coordinate (V46) at (2.810, 6.979);
\filldraw[fill=gray] (V46) circle [radius=0.1];
\coordinate (V47) at (3.284, 2.919);
\filldraw[fill=gray] (V47) circle [radius=0.1];
\coordinate (V48) at (3.075, 2.874);
\filldraw[fill=gray] (V48) circle [radius=0.1];
\coordinate (V49) at (3.165, 3.089);
\filldraw[fill=gray] (V49) circle [radius=0.1];
\coordinate (V50) at (4.798, 5.408);
\filldraw[fill=gray] (V50) circle [radius=0.1];
\coordinate (V51) at (4.644, 5.345);
\filldraw[fill=gray] (V51) circle [radius=0.1];
\coordinate (V52) at (4.652, 5.393);
\filldraw[fill=gray] (V52) circle [radius=0.1];
\coordinate (V53) at (6.821, 2.623);
\filldraw[fill=gray] (V53) circle [radius=0.1];
\coordinate (V54) at (0.500, 4.579);
\filldraw[fill=gray] (V54) circle [radius=0.1];
\coordinate (V55) at (0.562, 4.709);
\filldraw[fill=gray] (V55) circle [radius=0.1];
\coordinate (V56) at (8.920, 1.996);
\filldraw[fill=gray] (V56) circle [radius=0.1];
\coordinate (V57) at (9.070, 1.941);
\filldraw[fill=gray] (V57) circle [radius=0.1];
\draw[very thick, color=black] (V0) -- (V15);
\draw[very thick, color=black] (V0) -- (V47);
\draw[very thick, color=black] (V1) -- (V16);
\draw[very thick, color=black] (V1) -- (V21);
\draw[very thick, color=black] (V1) -- (V50);
\draw[very thick, color=black] (V2) -- (V37);
\draw[very thick, color=black] (V2) -- (V45);
\draw[very thick, color=black] (V3) -- (V7);
\draw[very thick, color=black] (V4) -- (V30);
\draw[very thick, color=black] (V5) -- (V51);
\draw[very thick, color=black] (V6) -- (V11);
\draw[very thick, color=black] (V8) -- (V52);
\draw[very thick, color=black] (V9) -- (V12);
\draw[very thick, color=black] (V10) -- (V17);
\draw[very thick, color=black] (V13) -- (V54);
\draw[very thick, color=black] (V14) -- (V33);
\draw[very thick, color=black] (V18) -- (V34);
\draw[very thick, color=black] (V19) -- (V41);
\draw[very thick, color=black] (V20) -- (V44);
\draw[very thick, color=black] (V22) -- (V25);
\draw[very thick, color=black] (V23) -- (V42);
\draw[very thick, color=black] (V24) -- (V31);
\draw[very thick, color=black] (V26) -- (V56);
\draw[very thick, color=black] (V27) -- (V53);
\draw[very thick, color=black] (V28) -- (V36);
\draw[very thick, color=black] (V29) -- (V55);
\draw[very thick, color=black] (V32) -- (V48);
\draw[very thick, color=black] (V35) -- (V57);
\draw[very thick, color=black] (V38) -- (V40);
\draw[very thick, color=black] (V39) -- (V46);
\draw[very thick, color=black] (V43) -- (V49);
\end{tikzpicture}
    }
                  \resizebox{0.19\linewidth}{!}{
    \begin{tikzpicture}[scale=1]
\coordinate (V0) at (0.500, 4.605);
\filldraw[fill=gray] (V0) circle [radius=0.1];
\coordinate (V1) at (1.680, 4.219);
\filldraw[fill=gray] (V1) circle [radius=0.1];
\coordinate (V2) at (3.749, 4.811);
\filldraw[fill=gray] (V2) circle [radius=0.1];
\coordinate (V3) at (5.067, 3.111);
\filldraw[fill=gray] (V3) circle [radius=0.1];
\coordinate (V4) at (6.963, 2.491);
\filldraw[fill=gray] (V4) circle [radius=0.1];
\coordinate (V5) at (7.487, 4.092);
\filldraw[fill=gray] (V5) circle [radius=0.1];
\coordinate (V6) at (1.024, 6.207);
\filldraw[fill=gray] (V6) circle [radius=0.1];
\coordinate (V7) at (4.862, 2.483);
\filldraw[fill=gray] (V7) circle [radius=0.1];
\coordinate (V8) at (3.168, 3.037);
\filldraw[fill=gray] (V8) circle [radius=0.1];
\coordinate (V9) at (1.474, 3.591);
\filldraw[fill=gray] (V9) circle [radius=0.1];
\coordinate (V10) at (8.011, 5.694);
\filldraw[fill=gray] (V10) circle [radius=0.1];
\coordinate (V11) at (6.985, 6.030);
\filldraw[fill=gray] (V11) circle [radius=0.1];
\coordinate (V12) at (4.661, 5.415);
\filldraw[fill=gray] (V12) circle [radius=0.1];
\coordinate (V13) at (3.149, 7.285);
\filldraw[fill=gray] (V13) circle [radius=0.1];
\coordinate (V14) at (1.548, 7.809);
\filldraw[fill=gray] (V14) circle [radius=0.1];
\coordinate (V15) at (2.685, 7.099);
\filldraw[fill=gray] (V15) circle [radius=0.1];
\coordinate (V16) at (2.730, 7.235);
\filldraw[fill=gray] (V16) circle [radius=0.1];
\coordinate (V17) at (2.894, 7.181);
\filldraw[fill=gray] (V17) circle [radius=0.1];
\coordinate (V18) at (2.850, 7.045);
\filldraw[fill=gray] (V18) circle [radius=0.1];
\coordinate (V19) at (3.401, 8.055);
\filldraw[fill=gray] (V19) circle [radius=0.1];
\coordinate (V20) at (5.319, 7.427);
\filldraw[fill=gray] (V20) circle [radius=0.1];
\coordinate (V21) at (7.236, 6.800);
\filldraw[fill=gray] (V21) circle [radius=0.1];
\coordinate (V22) at (7.360, 4.076);
\filldraw[fill=gray] (V22) circle [radius=0.1];
\coordinate (V23) at (6.894, 2.654);
\filldraw[fill=gray] (V23) circle [radius=0.1];
\coordinate (V24) at (7.002, 2.609);
\filldraw[fill=gray] (V24) circle [radius=0.1];
\coordinate (V25) at (9.032, 1.945);
\filldraw[fill=gray] (V25) circle [radius=0.1];
\coordinate (V26) at (9.500, 3.376);
\filldraw[fill=gray] (V26) circle [radius=0.1];
\draw[very thick, color=black] (V0) -- (V1);
\draw[very thick, color=black] (V0) -- (V6);
\draw[very thick, color=black] (V0) -- (V13);
\draw[very thick, color=black] (V1) -- (V2);
\draw[very thick, color=black] (V1) -- (V9);
\draw[very thick, color=black] (V2) -- (V3);
\draw[very thick, color=black] (V2) -- (V8);
\draw[very thick, color=black] (V3) -- (V4);
\draw[very thick, color=black] (V3) -- (V7);
\draw[very thick, color=black] (V4) -- (V5);
\draw[very thick, color=black] (V4) -- (V6);
\draw[very thick, color=black] (V5) -- (V6);
\draw[very thick, color=black] (V5) -- (V10);
\draw[very thick, color=black] (V6) -- (V14);
\draw[very thick, color=black] (V6) -- (V19);
\draw[very thick, color=black] (V6) -- (V26);
\draw[very thick, color=black] (V7) -- (V8);
\draw[very thick, color=black] (V8) -- (V9);
\draw[very thick, color=black] (V8) -- (V14);
\draw[very thick, color=black] (V10) -- (V11);
\draw[very thick, color=black] (V10) -- (V21);
\draw[very thick, color=black] (V11) -- (V12);
\draw[very thick, color=black] (V11) -- (V21);
\draw[very thick, color=black] (V12) -- (V13);
\draw[very thick, color=black] (V12) -- (V20);
\draw[very thick, color=black] (V13) -- (V14);
\draw[very thick, color=black] (V13) -- (V19);
\draw[very thick, color=black] (V15) -- (V16);
\draw[very thick, color=black] (V15) -- (V18);
\draw[very thick, color=black] (V15) -- (V26);
\draw[very thick, color=black] (V16) -- (V17);
\draw[very thick, color=black] (V17) -- (V18);
\draw[very thick, color=black] (V19) -- (V20);
\draw[very thick, color=black] (V20) -- (V21);
\draw[very thick, color=black] (V22) -- (V23);
\draw[very thick, color=black] (V22) -- (V26);
\draw[very thick, color=black] (V24) -- (V25);
\draw[very thick, color=black] (V25) -- (V26);
\end{tikzpicture}
    }
                      \resizebox{0.19\linewidth}{!}{
    \begin{tikzpicture}[scale=1]
\coordinate (V0) at (0.500, 4.605);
\filldraw[fill=gray] (V0) circle [radius=0.1];
\coordinate (V1) at (1.680, 4.219);
\filldraw[fill=gray] (V1) circle [radius=0.1];
\coordinate (V2) at (3.749, 4.811);
\filldraw[fill=gray] (V2) circle [radius=0.1];
\coordinate (V3) at (5.067, 3.111);
\filldraw[fill=gray] (V3) circle [radius=0.1];
\coordinate (V4) at (6.963, 2.491);
\filldraw[fill=gray] (V4) circle [radius=0.1];
\coordinate (V5) at (7.487, 4.092);
\filldraw[fill=gray] (V5) circle [radius=0.1];
\coordinate (V6) at (1.024, 6.207);
\filldraw[fill=gray] (V6) circle [radius=0.1];
\coordinate (V7) at (4.862, 2.483);
\filldraw[fill=gray] (V7) circle [radius=0.1];
\coordinate (V8) at (1.474, 3.591);
\filldraw[fill=gray] (V8) circle [radius=0.1];
\coordinate (V9) at (8.011, 5.694);
\filldraw[fill=gray] (V9) circle [radius=0.1];
\coordinate (V10) at (6.985, 6.030);
\filldraw[fill=gray] (V10) circle [radius=0.1];
\coordinate (V11) at (4.661, 5.415);
\filldraw[fill=gray] (V11) circle [radius=0.1];
\coordinate (V12) at (3.149, 7.285);
\filldraw[fill=gray] (V12) circle [radius=0.1];
\coordinate (V13) at (2.685, 7.099);
\filldraw[fill=gray] (V13) circle [radius=0.1];
\coordinate (V14) at (2.730, 7.235);
\filldraw[fill=gray] (V14) circle [radius=0.1];
\coordinate (V15) at (2.894, 7.181);
\filldraw[fill=gray] (V15) circle [radius=0.1];
\coordinate (V16) at (2.850, 7.045);
\filldraw[fill=gray] (V16) circle [radius=0.1];
\coordinate (V17) at (3.401, 8.055);
\filldraw[fill=gray] (V17) circle [radius=0.1];
\coordinate (V18) at (5.319, 7.427);
\filldraw[fill=gray] (V18) circle [radius=0.1];
\coordinate (V19) at (7.236, 6.800);
\filldraw[fill=gray] (V19) circle [radius=0.1];
\coordinate (V20) at (7.360, 4.076);
\filldraw[fill=gray] (V20) circle [radius=0.1];
\coordinate (V21) at (6.894, 2.654);
\filldraw[fill=gray] (V21) circle [radius=0.1];
\coordinate (V22) at (7.002, 2.609);
\filldraw[fill=gray] (V22) circle [radius=0.1];
\coordinate (V23) at (9.032, 1.945);
\filldraw[fill=gray] (V23) circle [radius=0.1];
\coordinate (V24) at (9.500, 3.376);
\filldraw[fill=gray] (V24) circle [radius=0.1];
\draw[very thick, color=black] (V0) -- (V1);
\draw[very thick, color=black] (V0) -- (V6);
\draw[very thick, color=black] (V1) -- (V2);
\draw[very thick, color=black] (V1) -- (V8);
\draw[very thick, color=black] (V2) -- (V3);
\draw[very thick, color=black] (V3) -- (V4);
\draw[very thick, color=black] (V3) -- (V7);
\draw[very thick, color=black] (V4) -- (V5);
\draw[very thick, color=black] (V5) -- (V6);
\draw[very thick, color=black] (V5) -- (V9);
\draw[very thick, color=black] (V9) -- (V10);
\draw[very thick, color=black] (V10) -- (V11);
\draw[very thick, color=black] (V10) -- (V19);
\draw[very thick, color=black] (V11) -- (V12);
\draw[very thick, color=black] (V11) -- (V18);
\draw[very thick, color=black] (V12) -- (V17);
\draw[very thick, color=black] (V13) -- (V14);
\draw[very thick, color=black] (V13) -- (V16);
\draw[very thick, color=black] (V14) -- (V15);
\draw[very thick, color=black] (V15) -- (V16);
\draw[very thick, color=black] (V17) -- (V18);
\draw[very thick, color=black] (V18) -- (V19);
\draw[very thick, color=black] (V20) -- (V21);
\draw[very thick, color=black] (V20) -- (V24);
\draw[very thick, color=black] (V22) -- (V23);
\draw[very thick, color=black] (V23) -- (V24);
\end{tikzpicture}
    }\\
            \resizebox{0.19\linewidth}{!}{
    \begin{tikzpicture}[scale=1]
\coordinate (V0) at (0.946, 6.913);
\filldraw[fill=gray] (V0) circle [radius=0.1];
\coordinate (V1) at (2.117, 4.695);
\filldraw[fill=gray] (V1) circle [radius=0.1];
\coordinate (V2) at (7.017, 7.282);
\filldraw[fill=gray] (V2) circle [radius=0.1];
\coordinate (V3) at (5.845, 9.500);
\filldraw[fill=gray] (V3) circle [radius=0.1];
\coordinate (V4) at (7.958, 4.943);
\filldraw[fill=gray] (V4) circle [radius=0.1];
\coordinate (V5) at (4.685, 5.712);
\filldraw[fill=gray] (V5) circle [radius=0.1];
\coordinate (V6) at (6.812, 1.684);
\filldraw[fill=gray] (V6) circle [radius=0.1];
\coordinate (V7) at (9.054, 2.868);
\filldraw[fill=gray] (V7) circle [radius=0.1];
\coordinate (V8) at (3.474, 2.575);
\filldraw[fill=gray] (V8) circle [radius=0.1];
\coordinate (V9) at (4.570, 0.500);
\filldraw[fill=gray] (V9) circle [radius=0.1];
\coordinate (V10) at (8.188, 5.064);
\filldraw[fill=gray] (V10) circle [radius=0.1];
\coordinate (V11) at (3.289, 2.477);
\filldraw[fill=gray] (V11) circle [radius=0.1];
\draw[very thick, color=black] (V0) -- (V1);
\draw[very thick, color=black] (V1) -- (V2);
\draw[very thick, color=black] (V2) -- (V3);
\draw[very thick, color=black] (V3) -- (V0);
\draw[very thick, color=black] (V4) -- (V5);
\draw[very thick, color=black] (V5) -- (V6);
\draw[very thick, color=black] (V6) -- (V7);
\draw[very thick, color=black] (V7) -- (V4);
\draw[very thick, color=black] (V5) -- (V8);
\draw[very thick, color=black] (V8) -- (V9);
\draw[very thick, color=black] (V9) -- (V6);
\draw[very thick, color=black] (V4) -- (V10);
\draw[very thick, color=black] (V10) -- (V2);
\draw[very thick, color=black] (V1) -- (V11);
\draw[very thick, color=black] (V11) -- (V8);
\end{tikzpicture}
    }
            \resizebox{0.19\linewidth}{!}{
    \begin{tikzpicture}[scale=1]
\coordinate (V0) at (0.958, 7.066);
\filldraw[fill=gray] (V0) circle [radius=0.1];
\coordinate (V1) at (2.185, 4.621);
\filldraw[fill=gray] (V1) circle [radius=0.1];
\coordinate (V2) at (7.020, 7.232);
\filldraw[fill=gray] (V2) circle [radius=0.1];
\coordinate (V3) at (6.019, 9.472);
\filldraw[fill=gray] (V3) circle [radius=0.1];
\coordinate (V4) at (7.773, 5.457);
\filldraw[fill=gray] (V4) circle [radius=0.1];
\coordinate (V5) at (4.708, 5.880);
\filldraw[fill=gray] (V5) circle [radius=0.1];
\coordinate (V6) at (6.100, 2.209);
\filldraw[fill=gray] (V6) circle [radius=0.1];
\coordinate (V7) at (9.000, 3.029);
\filldraw[fill=gray] (V7) circle [radius=0.1];
\coordinate (V8) at (2.878, 2.652);
\filldraw[fill=gray] (V8) circle [radius=0.1];
\coordinate (V9) at (4.769, 0.500);
\filldraw[fill=gray] (V9) circle [radius=0.1];
\coordinate (V10) at (8.271, 4.847);
\filldraw[fill=gray] (V10) circle [radius=0.1];
\coordinate (V11) at (2.956, 2.372);
\filldraw[fill=gray] (V11) circle [radius=0.1];
\coordinate (V12) at (1.672, 5.583);
\filldraw[fill=gray] (V12) circle [radius=0.1];
\coordinate (V13) at (1.859, 5.407);
\filldraw[fill=gray] (V13) circle [radius=0.1];
\coordinate (V14) at (1.876, 5.317);
\filldraw[fill=gray] (V14) circle [radius=0.1];
\coordinate (V15) at (3.750, 8.810);
\filldraw[fill=gray] (V15) circle [radius=0.1];
\coordinate (V16) at (5.061, 9.258);
\filldraw[fill=gray] (V16) circle [radius=0.1];
\coordinate (V17) at (5.957, 9.340);
\filldraw[fill=gray] (V17) circle [radius=0.1];
\coordinate (V18) at (2.888, 4.961);
\filldraw[fill=gray] (V18) circle [radius=0.1];
\coordinate (V19) at (3.929, 5.623);
\filldraw[fill=gray] (V19) circle [radius=0.1];
\coordinate (V20) at (6.933, 7.726);
\filldraw[fill=gray] (V20) circle [radius=0.1];
\coordinate (V21) at (2.515, 4.176);
\filldraw[fill=gray] (V21) circle [radius=0.1];
\coordinate (V22) at (5.798, 9.500);
\filldraw[fill=gray] (V22) circle [radius=0.1];
\coordinate (V23) at (7.281, 5.982);
\filldraw[fill=gray] (V23) circle [radius=0.1];
\coordinate (V24) at (8.001, 5.278);
\filldraw[fill=gray] (V24) circle [radius=0.1];
\coordinate (V25) at (8.202, 4.988);
\filldraw[fill=gray] (V25) circle [radius=0.1];
\coordinate (V26) at (6.225, 5.488);
\filldraw[fill=gray] (V26) circle [radius=0.1];
\coordinate (V27) at (5.893, 5.464);
\filldraw[fill=gray] (V27) circle [radius=0.1];
\coordinate (V28) at (8.317, 4.941);
\filldraw[fill=gray] (V28) circle [radius=0.1];
\coordinate (V29) at (8.459, 4.726);
\filldraw[fill=gray] (V29) circle [radius=0.1];
\coordinate (V30) at (9.042, 4.068);
\filldraw[fill=gray] (V30) circle [radius=0.1];
\coordinate (V31) at (7.527, 5.247);
\filldraw[fill=gray] (V31) circle [radius=0.1];
\coordinate (V32) at (5.796, 3.588);
\filldraw[fill=gray] (V32) circle [radius=0.1];
\coordinate (V33) at (6.389, 2.926);
\filldraw[fill=gray] (V33) circle [radius=0.1];
\coordinate (V34) at (3.494, 3.635);
\filldraw[fill=gray] (V34) circle [radius=0.1];
\coordinate (V35) at (3.452, 3.807);
\filldraw[fill=gray] (V35) circle [radius=0.1];
\coordinate (V36) at (3.155, 2.830);
\filldraw[fill=gray] (V36) circle [radius=0.1];
\coordinate (V37) at (6.625, 2.105);
\filldraw[fill=gray] (V37) circle [radius=0.1];
\coordinate (V38) at (6.911, 1.839);
\filldraw[fill=gray] (V38) circle [radius=0.1];
\coordinate (V39) at (5.613, 1.452);
\filldraw[fill=gray] (V39) circle [radius=0.1];
\coordinate (V40) at (5.455, 1.702);
\filldraw[fill=gray] (V40) circle [radius=0.1];
\coordinate (V41) at (4.759, 0.952);
\filldraw[fill=gray] (V41) circle [radius=0.1];
\coordinate (V42) at (2.929, 2.183);
\filldraw[fill=gray] (V42) circle [radius=0.1];
\coordinate (V43) at (2.733, 2.656);
\filldraw[fill=gray] (V43) circle [radius=0.1];
\coordinate (V44) at (2.712, 2.284);
\filldraw[fill=gray] (V44) circle [radius=0.1];
\coordinate (V45) at (2.863, 2.296);
\filldraw[fill=gray] (V45) circle [radius=0.1];
\draw[very thick, color=black] (V0) -- (V12);
\draw[very thick, color=black] (V0) -- (V15);
\draw[very thick, color=black] (V1) -- (V14);
\draw[very thick, color=black] (V1) -- (V18);
\draw[very thick, color=black] (V1) -- (V21);
\draw[very thick, color=black] (V2) -- (V20);
\draw[very thick, color=black] (V2) -- (V22);
\draw[very thick, color=black] (V2) -- (V23);
\draw[very thick, color=black] (V3) -- (V17);
\draw[very thick, color=black] (V3) -- (V22);
\draw[very thick, color=black] (V4) -- (V26);
\draw[very thick, color=black] (V4) -- (V28);
\draw[very thick, color=black] (V4) -- (V31);
\draw[very thick, color=black] (V5) -- (V27);
\draw[very thick, color=black] (V5) -- (V32);
\draw[very thick, color=black] (V5) -- (V34);
\draw[very thick, color=black] (V6) -- (V33);
\draw[very thick, color=black] (V6) -- (V37);
\draw[very thick, color=black] (V6) -- (V39);
\draw[very thick, color=black] (V7) -- (V30);
\draw[very thick, color=black] (V7) -- (V38);
\draw[very thick, color=black] (V8) -- (V36);
\draw[very thick, color=black] (V8) -- (V42);
\draw[very thick, color=black] (V8) -- (V43);
\draw[very thick, color=black] (V9) -- (V41);
\draw[very thick, color=black] (V9) -- (V42);
\draw[very thick, color=black] (V10) -- (V25);
\draw[very thick, color=black] (V10) -- (V31);
\draw[very thick, color=black] (V11) -- (V21);
\draw[very thick, color=black] (V11) -- (V45);
\draw[very thick, color=black] (V12) -- (V13);
\draw[very thick, color=black] (V13) -- (V14);
\draw[very thick, color=black] (V15) -- (V16);
\draw[very thick, color=black] (V16) -- (V17);
\draw[very thick, color=black] (V18) -- (V19);
\draw[very thick, color=black] (V19) -- (V20);
\draw[very thick, color=black] (V23) -- (V24);
\draw[very thick, color=black] (V24) -- (V25);
\draw[very thick, color=black] (V26) -- (V27);
\draw[very thick, color=black] (V28) -- (V29);
\draw[very thick, color=black] (V29) -- (V30);
\draw[very thick, color=black] (V32) -- (V33);
\draw[very thick, color=black] (V34) -- (V35);
\draw[very thick, color=black] (V35) -- (V36);
\draw[very thick, color=black] (V37) -- (V38);
\draw[very thick, color=black] (V39) -- (V40);
\draw[very thick, color=black] (V40) -- (V41);
\draw[very thick, color=black] (V43) -- (V44);
\draw[very thick, color=black] (V44) -- (V45);
\end{tikzpicture}
    }
               \resizebox{0.19\linewidth}{!}{
    \begin{tikzpicture}[scale=1]
\coordinate (V0) at (7.895, 4.939);
\filldraw[fill=gray] (V0) circle [radius=0.1];
\coordinate (V1) at (3.546, 2.642);
\filldraw[fill=gray] (V1) circle [radius=0.1];
\coordinate (V2) at (1.061, 7.103);
\filldraw[fill=gray] (V2) circle [radius=0.1];
\coordinate (V3) at (1.229, 6.821);
\filldraw[fill=gray] (V3) circle [radius=0.1];
\coordinate (V4) at (5.911, 9.372);
\filldraw[fill=gray] (V4) circle [radius=0.1];
\coordinate (V5) at (5.863, 9.500);
\filldraw[fill=gray] (V5) circle [radius=0.1];
\coordinate (V6) at (8.939, 2.775);
\filldraw[fill=gray] (V6) circle [radius=0.1];
\coordinate (V7) at (8.841, 2.785);
\filldraw[fill=gray] (V7) circle [radius=0.1];
\coordinate (V8) at (7.883, 5.137);
\filldraw[fill=gray] (V8) circle [radius=0.1];
\coordinate (V9) at (8.214, 4.741);
\filldraw[fill=gray] (V9) circle [radius=0.1];
\coordinate (V10) at (7.040, 1.769);
\filldraw[fill=gray] (V10) circle [radius=0.1];
\coordinate (V11) at (6.922, 1.695);
\filldraw[fill=gray] (V11) circle [radius=0.1];
\coordinate (V12) at (6.554, 1.841);
\filldraw[fill=gray] (V12) circle [radius=0.1];
\coordinate (V13) at (2.479, 4.453);
\filldraw[fill=gray] (V13) circle [radius=0.1];
\coordinate (V14) at (2.573, 4.739);
\filldraw[fill=gray] (V14) circle [radius=0.1];
\coordinate (V15) at (2.409, 4.354);
\filldraw[fill=gray] (V15) circle [radius=0.1];
\coordinate (V16) at (4.591, 0.657);
\filldraw[fill=gray] (V16) circle [radius=0.1];
\coordinate (V17) at (4.924, 0.500);
\filldraw[fill=gray] (V17) circle [radius=0.1];
\coordinate (V18) at (6.772, 6.875);
\filldraw[fill=gray] (V18) circle [radius=0.1];
\coordinate (V19) at (6.666, 6.880);
\filldraw[fill=gray] (V19) circle [radius=0.1];
\coordinate (V20) at (6.997, 6.897);
\filldraw[fill=gray] (V20) circle [radius=0.1];
\coordinate (V21) at (3.263, 2.748);
\filldraw[fill=gray] (V21) circle [radius=0.1];
\coordinate (V22) at (3.357, 2.530);
\filldraw[fill=gray] (V22) circle [radius=0.1];
\coordinate (V23) at (4.548, 5.879);
\filldraw[fill=gray] (V23) circle [radius=0.1];
\coordinate (V24) at (4.864, 5.442);
\filldraw[fill=gray] (V24) circle [radius=0.1];
\coordinate (V25) at (4.453, 5.416);
\filldraw[fill=gray] (V25) circle [radius=0.1];
\draw[very thick, color=black] (V0) -- (V7);
\draw[very thick, color=black] (V0) -- (V8);
\draw[very thick, color=black] (V0) -- (V23);
\draw[very thick, color=black] (V1) -- (V16);
\draw[very thick, color=black] (V1) -- (V21);
\draw[very thick, color=black] (V1) -- (V24);
\draw[very thick, color=black] (V2) -- (V15);
\draw[very thick, color=black] (V3) -- (V5);
\draw[very thick, color=black] (V4) -- (V18);
\draw[very thick, color=black] (V6) -- (V12);
\draw[very thick, color=black] (V9) -- (V19);
\draw[very thick, color=black] (V10) -- (V25);
\draw[very thick, color=black] (V11) -- (V17);
\draw[very thick, color=black] (V13) -- (V20);
\draw[very thick, color=black] (V14) -- (V22);
\end{tikzpicture}
    }
                  \resizebox{0.19\linewidth}{!}{
    \begin{tikzpicture}[scale=1]
\coordinate (V0) at (0.946, 6.913);
\filldraw[fill=gray] (V0) circle [radius=0.1];
\coordinate (V1) at (2.117, 4.695);
\filldraw[fill=gray] (V1) circle [radius=0.1];
\coordinate (V2) at (7.017, 7.282);
\filldraw[fill=gray] (V2) circle [radius=0.1];
\coordinate (V3) at (5.845, 9.500);
\filldraw[fill=gray] (V3) circle [radius=0.1];
\coordinate (V4) at (7.958, 4.943);
\filldraw[fill=gray] (V4) circle [radius=0.1];
\coordinate (V5) at (4.685, 5.712);
\filldraw[fill=gray] (V5) circle [radius=0.1];
\coordinate (V6) at (6.812, 1.684);
\filldraw[fill=gray] (V6) circle [radius=0.1];
\coordinate (V7) at (9.054, 2.868);
\filldraw[fill=gray] (V7) circle [radius=0.1];
\coordinate (V8) at (3.474, 2.575);
\filldraw[fill=gray] (V8) circle [radius=0.1];
\coordinate (V9) at (4.570, 0.500);
\filldraw[fill=gray] (V9) circle [radius=0.1];
\coordinate (V10) at (8.188, 5.064);
\filldraw[fill=gray] (V10) circle [radius=0.1];
\coordinate (V11) at (3.289, 2.477);
\filldraw[fill=gray] (V11) circle [radius=0.1];
\draw[very thick, color=black] (V0) -- (V1);
\draw[very thick, color=black] (V0) -- (V3);
\draw[very thick, color=black] (V0) -- (V7);
\draw[very thick, color=black] (V0) -- (V2);
\draw[very thick, color=black] (V1) -- (V2);
\draw[very thick, color=black] (V1) -- (V11);
\draw[very thick, color=black] (V1) -- (V8);
\draw[very thick, color=black] (V2) -- (V3);
\draw[very thick, color=black] (V2) -- (V10);
\draw[very thick, color=black] (V2) -- (V6);
\draw[very thick, color=black] (V4) -- (V5);
\draw[very thick, color=black] (V4) -- (V7);
\draw[very thick, color=black] (V4) -- (V10);
\draw[very thick, color=black] (V5) -- (V6);
\draw[very thick, color=black] (V5) -- (V8);
\draw[very thick, color=black] (V6) -- (V7);
\draw[very thick, color=black] (V6) -- (V9);
\draw[very thick, color=black] (V8) -- (V9);
\draw[very thick, color=black] (V8) -- (V11);
\end{tikzpicture}
    }
                      \resizebox{0.19\linewidth}{!}{
    \begin{tikzpicture}[scale=1]
\coordinate (V0) at (0.946, 6.913);
\filldraw[fill=gray] (V0) circle [radius=0.1];
\coordinate (V1) at (2.117, 4.695);
\filldraw[fill=gray] (V1) circle [radius=0.1];
\coordinate (V2) at (7.017, 7.282);
\filldraw[fill=gray] (V2) circle [radius=0.1];
\coordinate (V3) at (5.845, 9.500);
\filldraw[fill=gray] (V3) circle [radius=0.1];
\coordinate (V4) at (7.958, 4.943);
\filldraw[fill=gray] (V4) circle [radius=0.1];
\coordinate (V5) at (4.685, 5.712);
\filldraw[fill=gray] (V5) circle [radius=0.1];
\coordinate (V6) at (6.812, 1.684);
\filldraw[fill=gray] (V6) circle [radius=0.1];
\coordinate (V7) at (9.054, 2.868);
\filldraw[fill=gray] (V7) circle [radius=0.1];
\coordinate (V8) at (3.474, 2.575);
\filldraw[fill=gray] (V8) circle [radius=0.1];
\coordinate (V9) at (4.570, 0.500);
\filldraw[fill=gray] (V9) circle [radius=0.1];
\coordinate (V10) at (8.188, 5.064);
\filldraw[fill=gray] (V10) circle [radius=0.1];
\draw[very thick, color=black] (V0) -- (V1);
\draw[very thick, color=black] (V0) -- (V3);
\draw[very thick, color=black] (V1) -- (V2);
\draw[very thick, color=black] (V2) -- (V3);
\draw[very thick, color=black] (V2) -- (V10);
\draw[very thick, color=black] (V4) -- (V5);
\draw[very thick, color=black] (V4) -- (V7);
\draw[very thick, color=black] (V4) -- (V10);
\draw[very thick, color=black] (V5) -- (V6);
\draw[very thick, color=black] (V5) -- (V8);
\draw[very thick, color=black] (V6) -- (V7);
\draw[very thick, color=black] (V6) -- (V9);
\draw[very thick, color=black] (V8) -- (V9);
\end{tikzpicture}
    }\\
            \resizebox{0.19\linewidth}{!}{
    \begin{tikzpicture}[scale=1]
\coordinate (V0) at (2.316, 5.516);
\filldraw[fill=gray] (V0) circle [radius=0.1];
\coordinate (V1) at (1.848, 3.785);
\filldraw[fill=gray] (V1) circle [radius=0.1];
\coordinate (V2) at (8.564, 1.969);
\filldraw[fill=gray] (V2) circle [radius=0.1];
\coordinate (V3) at (9.032, 3.700);
\filldraw[fill=gray] (V3) circle [radius=0.1];
\coordinate (V4) at (3.074, 7.590);
\filldraw[fill=gray] (V4) circle [radius=0.1];
\coordinate (V5) at (1.443, 8.031);
\filldraw[fill=gray] (V5) circle [radius=0.1];
\coordinate (V6) at (0.971, 6.287);
\filldraw[fill=gray] (V6) circle [radius=0.1];
\coordinate (V7) at (2.603, 5.846);
\filldraw[fill=gray] (V7) circle [radius=0.1];
\coordinate (V8) at (0.500, 4.543);
\filldraw[fill=gray] (V8) circle [radius=0.1];
\coordinate (V9) at (1.947, 4.152);
\filldraw[fill=gray] (V9) circle [radius=0.1];
\coordinate (V10) at (2.341, 5.607);
\filldraw[fill=gray] (V10) circle [radius=0.1];
\coordinate (V11) at (2.363, 5.690);
\filldraw[fill=gray] (V11) circle [radius=0.1];
\coordinate (V12) at (2.547, 5.640);
\filldraw[fill=gray] (V12) circle [radius=0.1];
\coordinate (V13) at (9.500, 5.431);
\filldraw[fill=gray] (V13) circle [radius=0.1];
\coordinate (V14) at (2.900, 7.216);
\filldraw[fill=gray] (V14) circle [radius=0.1];
\coordinate (V15) at (2.503, 5.745);
\filldraw[fill=gray] (V15) circle [radius=0.1];
\draw[very thick, color=black] (V0) -- (V1);
\draw[very thick, color=black] (V1) -- (V2);
\draw[very thick, color=black] (V2) -- (V3);
\draw[very thick, color=black] (V3) -- (V0);
\draw[very thick, color=black] (V4) -- (V5);
\draw[very thick, color=black] (V5) -- (V6);
\draw[very thick, color=black] (V6) -- (V7);
\draw[very thick, color=black] (V7) -- (V4);
\draw[very thick, color=black] (V8) -- (V9);
\draw[very thick, color=black] (V9) -- (V10);
\draw[very thick, color=black] (V10) -- (V11);
\draw[very thick, color=black] (V11) -- (V12);
\draw[very thick, color=black] (V12) -- (V7);
\draw[very thick, color=black] (V6) -- (V8);
\draw[very thick, color=black] (V13) -- (V14);
\draw[very thick, color=black] (V14) -- (V15);
\draw[very thick, color=black] (V11) -- (V0);
\draw[very thick, color=black] (V3) -- (V13);
\end{tikzpicture}
    }
            \resizebox{0.19\linewidth}{!}{
    \begin{tikzpicture}[scale=1]
\coordinate (V0) at (2.613, 5.619);
\filldraw[fill=gray] (V0) circle [radius=0.1];
\coordinate (V1) at (1.635, 3.868);
\filldraw[fill=gray] (V1) circle [radius=0.1];
\coordinate (V2) at (8.226, 2.046);
\filldraw[fill=gray] (V2) circle [radius=0.1];
\coordinate (V3) at (8.828, 4.038);
\filldraw[fill=gray] (V3) circle [radius=0.1];
\coordinate (V4) at (2.944, 7.720);
\filldraw[fill=gray] (V4) circle [radius=0.1];
\coordinate (V5) at (1.362, 7.880);
\filldraw[fill=gray] (V5) circle [radius=0.1];
\coordinate (V6) at (1.093, 6.342);
\filldraw[fill=gray] (V6) circle [radius=0.1];
\coordinate (V7) at (2.672, 6.203);
\filldraw[fill=gray] (V7) circle [radius=0.1];
\coordinate (V8) at (0.547, 4.484);
\filldraw[fill=gray] (V8) circle [radius=0.1];
\coordinate (V9) at (1.611, 4.044);
\filldraw[fill=gray] (V9) circle [radius=0.1];
\coordinate (V10) at (2.198, 5.499);
\filldraw[fill=gray] (V10) circle [radius=0.1];
\coordinate (V11) at (2.406, 5.655);
\filldraw[fill=gray] (V11) circle [radius=0.1];
\coordinate (V12) at (2.534, 5.670);
\filldraw[fill=gray] (V12) circle [radius=0.1];
\coordinate (V13) at (9.500, 5.316);
\filldraw[fill=gray] (V13) circle [radius=0.1];
\coordinate (V14) at (2.874, 7.416);
\filldraw[fill=gray] (V14) circle [radius=0.1];
\coordinate (V15) at (2.541, 5.674);
\filldraw[fill=gray] (V15) circle [radius=0.1];
\coordinate (V16) at (1.959, 4.349);
\filldraw[fill=gray] (V16) circle [radius=0.1];
\coordinate (V17) at (4.775, 4.986);
\filldraw[fill=gray] (V17) circle [radius=0.1];
\coordinate (V18) at (5.592, 4.780);
\filldraw[fill=gray] (V18) circle [radius=0.1];
\coordinate (V19) at (7.778, 4.177);
\filldraw[fill=gray] (V19) circle [radius=0.1];
\coordinate (V20) at (2.675, 5.548);
\filldraw[fill=gray] (V20) circle [radius=0.1];
\coordinate (V21) at (2.486, 5.523);
\filldraw[fill=gray] (V21) circle [radius=0.1];
\coordinate (V22) at (2.963, 3.649);
\filldraw[fill=gray] (V22) circle [radius=0.1];
\coordinate (V23) at (4.220, 3.352);
\filldraw[fill=gray] (V23) circle [radius=0.1];
\coordinate (V24) at (8.623, 3.902);
\filldraw[fill=gray] (V24) circle [radius=0.1];
\coordinate (V25) at (8.989, 4.240);
\filldraw[fill=gray] (V25) circle [radius=0.1];
\coordinate (V26) at (9.009, 4.545);
\filldraw[fill=gray] (V26) circle [radius=0.1];
\coordinate (V27) at (9.164, 4.711);
\filldraw[fill=gray] (V27) circle [radius=0.1];
\coordinate (V28) at (1.157, 7.917);
\filldraw[fill=gray] (V28) circle [radius=0.1];
\coordinate (V29) at (2.744, 7.160);
\filldraw[fill=gray] (V29) circle [radius=0.1];
\coordinate (V30) at (2.808, 6.551);
\filldraw[fill=gray] (V30) circle [radius=0.1];
\coordinate (V31) at (1.322, 7.954);
\filldraw[fill=gray] (V31) circle [radius=0.1];
\coordinate (V32) at (1.236, 6.946);
\filldraw[fill=gray] (V32) circle [radius=0.1];
\coordinate (V33) at (1.094, 6.240);
\filldraw[fill=gray] (V33) circle [radius=0.1];
\coordinate (V34) at (2.347, 6.114);
\filldraw[fill=gray] (V34) circle [radius=0.1];
\coordinate (V35) at (0.604, 5.005);
\filldraw[fill=gray] (V35) circle [radius=0.1];
\coordinate (V36) at (0.662, 4.476);
\filldraw[fill=gray] (V36) circle [radius=0.1];
\coordinate (V37) at (0.500, 4.555);
\filldraw[fill=gray] (V37) circle [radius=0.1];
\coordinate (V38) at (2.610, 6.002);
\filldraw[fill=gray] (V38) circle [radius=0.1];
\coordinate (V39) at (2.715, 6.007);
\filldraw[fill=gray] (V39) circle [radius=0.1];
\coordinate (V40) at (2.510, 5.662);
\filldraw[fill=gray] (V40) circle [radius=0.1];
\coordinate (V41) at (1.089, 4.384);
\filldraw[fill=gray] (V41) circle [radius=0.1];
\coordinate (V42) at (1.325, 4.323);
\filldraw[fill=gray] (V42) circle [radius=0.1];
\coordinate (V43) at (2.235, 5.166);
\filldraw[fill=gray] (V43) circle [radius=0.1];
\coordinate (V44) at (2.258, 5.559);
\filldraw[fill=gray] (V44) circle [radius=0.1];
\coordinate (V45) at (2.326, 5.398);
\filldraw[fill=gray] (V45) circle [radius=0.1];
\coordinate (V46) at (2.250, 5.446);
\filldraw[fill=gray] (V46) circle [radius=0.1];
\coordinate (V47) at (2.367, 5.504);
\filldraw[fill=gray] (V47) circle [radius=0.1];
\coordinate (V48) at (9.332, 5.445);
\filldraw[fill=gray] (V48) circle [radius=0.1];
\coordinate (V49) at (7.976, 5.863);
\filldraw[fill=gray] (V49) circle [radius=0.1];
\coordinate (V50) at (5.711, 6.457);
\filldraw[fill=gray] (V50) circle [radius=0.1];
\coordinate (V51) at (2.754, 6.314);
\filldraw[fill=gray] (V51) circle [radius=0.1];
\draw[very thick, color=black] (V0) -- (V16);
\draw[very thick, color=black] (V0) -- (V17);
\draw[very thick, color=black] (V0) -- (V20);
\draw[very thick, color=black] (V1) -- (V16);
\draw[very thick, color=black] (V1) -- (V22);
\draw[very thick, color=black] (V2) -- (V23);
\draw[very thick, color=black] (V2) -- (V24);
\draw[very thick, color=black] (V3) -- (V19);
\draw[very thick, color=black] (V3) -- (V24);
\draw[very thick, color=black] (V3) -- (V25);
\draw[very thick, color=black] (V4) -- (V28);
\draw[very thick, color=black] (V4) -- (V29);
\draw[very thick, color=black] (V5) -- (V28);
\draw[very thick, color=black] (V5) -- (V31);
\draw[very thick, color=black] (V6) -- (V32);
\draw[very thick, color=black] (V6) -- (V33);
\draw[very thick, color=black] (V6) -- (V35);
\draw[very thick, color=black] (V7) -- (V30);
\draw[very thick, color=black] (V7) -- (V34);
\draw[very thick, color=black] (V7) -- (V38);
\draw[very thick, color=black] (V8) -- (V37);
\draw[very thick, color=black] (V8) -- (V41);
\draw[very thick, color=black] (V9) -- (V42);
\draw[very thick, color=black] (V9) -- (V43);
\draw[very thick, color=black] (V10) -- (V43);
\draw[very thick, color=black] (V10) -- (V44);
\draw[very thick, color=black] (V11) -- (V21);
\draw[very thick, color=black] (V11) -- (V46);
\draw[very thick, color=black] (V11) -- (V47);
\draw[very thick, color=black] (V12) -- (V40);
\draw[very thick, color=black] (V12) -- (V47);
\draw[very thick, color=black] (V13) -- (V27);
\draw[very thick, color=black] (V13) -- (V48);
\draw[very thick, color=black] (V14) -- (V50);
\draw[very thick, color=black] (V14) -- (V51);
\draw[very thick, color=black] (V15) -- (V51);
\draw[very thick, color=black] (V17) -- (V18);
\draw[very thick, color=black] (V18) -- (V19);
\draw[very thick, color=black] (V20) -- (V21);
\draw[very thick, color=black] (V22) -- (V23);
\draw[very thick, color=black] (V25) -- (V26);
\draw[very thick, color=black] (V26) -- (V27);
\draw[very thick, color=black] (V29) -- (V30);
\draw[very thick, color=black] (V31) -- (V32);
\draw[very thick, color=black] (V33) -- (V34);
\draw[very thick, color=black] (V35) -- (V36);
\draw[very thick, color=black] (V36) -- (V37);
\draw[very thick, color=black] (V38) -- (V39);
\draw[very thick, color=black] (V39) -- (V40);
\draw[very thick, color=black] (V41) -- (V42);
\draw[very thick, color=black] (V44) -- (V45);
\draw[very thick, color=black] (V45) -- (V46);
\draw[very thick, color=black] (V48) -- (V49);
\draw[very thick, color=black] (V49) -- (V50);
\end{tikzpicture}
    }
               \resizebox{0.19\linewidth}{!}{
    \begin{tikzpicture}[scale=1]
\coordinate (V0) at (9.000, 3.717);
\filldraw[fill=gray] (V0) circle [radius=0.1];
\coordinate (V1) at (0.500, 4.556);
\filldraw[fill=gray] (V1) circle [radius=0.1];
\coordinate (V2) at (2.570, 5.968);
\filldraw[fill=gray] (V2) circle [radius=0.1];
\coordinate (V3) at (2.462, 5.843);
\filldraw[fill=gray] (V3) circle [radius=0.1];
\coordinate (V4) at (2.554, 5.724);
\filldraw[fill=gray] (V4) circle [radius=0.1];
\coordinate (V5) at (2.675, 5.464);
\filldraw[fill=gray] (V5) circle [radius=0.1];
\coordinate (V6) at (2.420, 5.582);
\filldraw[fill=gray] (V6) circle [radius=0.1];
\coordinate (V7) at (2.892, 7.090);
\filldraw[fill=gray] (V7) circle [radius=0.1];
\coordinate (V8) at (2.816, 7.159);
\filldraw[fill=gray] (V8) circle [radius=0.1];
\coordinate (V9) at (2.164, 5.834);
\filldraw[fill=gray] (V9) circle [radius=0.1];
\coordinate (V10) at (2.324, 5.741);
\filldraw[fill=gray] (V10) circle [radius=0.1];
\coordinate (V11) at (2.490, 5.863);
\filldraw[fill=gray] (V11) circle [radius=0.1];
\coordinate (V12) at (1.786, 4.018);
\filldraw[fill=gray] (V12) circle [radius=0.1];
\coordinate (V13) at (2.109, 4.076);
\filldraw[fill=gray] (V13) circle [radius=0.1];
\coordinate (V14) at (3.131, 7.456);
\filldraw[fill=gray] (V14) circle [radius=0.1];
\coordinate (V15) at (2.917, 7.671);
\filldraw[fill=gray] (V15) circle [radius=0.1];
\coordinate (V16) at (1.343, 8.007);
\filldraw[fill=gray] (V16) circle [radius=0.1];
\coordinate (V17) at (1.595, 8.193);
\filldraw[fill=gray] (V17) circle [radius=0.1];
\coordinate (V18) at (0.916, 6.234);
\filldraw[fill=gray] (V18) circle [radius=0.1];
\coordinate (V19) at (0.912, 6.177);
\filldraw[fill=gray] (V19) circle [radius=0.1];
\coordinate (V20) at (0.929, 6.309);
\filldraw[fill=gray] (V20) circle [radius=0.1];
\coordinate (V21) at (2.432, 5.595);
\filldraw[fill=gray] (V21) circle [radius=0.1];
\coordinate (V22) at (2.341, 5.778);
\filldraw[fill=gray] (V22) circle [radius=0.1];
\coordinate (V23) at (2.191, 5.678);
\filldraw[fill=gray] (V23) circle [radius=0.1];
\coordinate (V24) at (1.961, 3.816);
\filldraw[fill=gray] (V24) circle [radius=0.1];
\coordinate (V25) at (1.674, 3.938);
\filldraw[fill=gray] (V25) circle [radius=0.1];
\coordinate (V26) at (2.246, 5.687);
\filldraw[fill=gray] (V26) circle [radius=0.1];
\coordinate (V27) at (2.380, 5.589);
\filldraw[fill=gray] (V27) circle [radius=0.1];
\coordinate (V28) at (2.290, 5.342);
\filldraw[fill=gray] (V28) circle [radius=0.1];
\coordinate (V29) at (9.414, 5.253);
\filldraw[fill=gray] (V29) circle [radius=0.1];
\coordinate (V30) at (9.500, 5.571);
\filldraw[fill=gray] (V30) circle [radius=0.1];
\coordinate (V31) at (8.561, 1.979);
\filldraw[fill=gray] (V31) circle [radius=0.1];
\coordinate (V32) at (8.720, 1.807);
\filldraw[fill=gray] (V32) circle [radius=0.1];
\draw[very thick, color=black] (V0) -- (V26);
\draw[very thick, color=black] (V0) -- (V29);
\draw[very thick, color=black] (V0) -- (V31);
\draw[very thick, color=black] (V1) -- (V12);
\draw[very thick, color=black] (V1) -- (V18);
\draw[very thick, color=black] (V2) -- (V19);
\draw[very thick, color=black] (V3) -- (V15);
\draw[very thick, color=black] (V4) -- (V6);
\draw[very thick, color=black] (V5) -- (V11);
\draw[very thick, color=black] (V7) -- (V30);
\draw[very thick, color=black] (V8) -- (V21);
\draw[very thick, color=black] (V9) -- (V22);
\draw[very thick, color=black] (V10) -- (V27);
\draw[very thick, color=black] (V13) -- (V23);
\draw[very thick, color=black] (V14) -- (V17);
\draw[very thick, color=black] (V16) -- (V20);
\draw[very thick, color=black] (V24) -- (V28);
\draw[very thick, color=black] (V25) -- (V32);
\end{tikzpicture}
    }
                  \resizebox{0.19\linewidth}{!}{
    \begin{tikzpicture}[scale=1]
\coordinate (V0) at (2.316, 5.516);
\filldraw[fill=gray] (V0) circle [radius=0.1];
\coordinate (V1) at (1.848, 3.785);
\filldraw[fill=gray] (V1) circle [radius=0.1];
\coordinate (V2) at (8.564, 1.969);
\filldraw[fill=gray] (V2) circle [radius=0.1];
\coordinate (V3) at (9.032, 3.700);
\filldraw[fill=gray] (V3) circle [radius=0.1];
\coordinate (V4) at (3.074, 7.590);
\filldraw[fill=gray] (V4) circle [radius=0.1];
\coordinate (V5) at (1.443, 8.031);
\filldraw[fill=gray] (V5) circle [radius=0.1];
\coordinate (V6) at (0.971, 6.287);
\filldraw[fill=gray] (V6) circle [radius=0.1];
\coordinate (V7) at (2.603, 5.846);
\filldraw[fill=gray] (V7) circle [radius=0.1];
\coordinate (V8) at (0.500, 4.543);
\filldraw[fill=gray] (V8) circle [radius=0.1];
\coordinate (V9) at (1.947, 4.152);
\filldraw[fill=gray] (V9) circle [radius=0.1];
\coordinate (V10) at (2.341, 5.607);
\filldraw[fill=gray] (V10) circle [radius=0.1];
\coordinate (V11) at (2.363, 5.690);
\filldraw[fill=gray] (V11) circle [radius=0.1];
\coordinate (V12) at (2.547, 5.640);
\filldraw[fill=gray] (V12) circle [radius=0.1];
\coordinate (V13) at (9.500, 5.431);
\filldraw[fill=gray] (V13) circle [radius=0.1];
\coordinate (V14) at (2.900, 7.216);
\filldraw[fill=gray] (V14) circle [radius=0.1];
\coordinate (V15) at (2.503, 5.745);
\filldraw[fill=gray] (V15) circle [radius=0.1];
\draw[very thick, color=black] (V0) -- (V1);
\draw[very thick, color=black] (V0) -- (V3);
\draw[very thick, color=black] (V0) -- (V11);
\draw[very thick, color=black] (V0) -- (V7);
\draw[very thick, color=black] (V1) -- (V2);
\draw[very thick, color=black] (V2) -- (V3);
\draw[very thick, color=black] (V2) -- (V11);
\draw[very thick, color=black] (V3) -- (V13);
\draw[very thick, color=black] (V4) -- (V5);
\draw[very thick, color=black] (V4) -- (V7);
\draw[very thick, color=black] (V4) -- (V14);
\draw[very thick, color=black] (V5) -- (V6);
\draw[very thick, color=black] (V6) -- (V7);
\draw[very thick, color=black] (V6) -- (V8);
\draw[very thick, color=black] (V7) -- (V12);
\draw[very thick, color=black] (V8) -- (V9);
\draw[very thick, color=black] (V9) -- (V10);
\draw[very thick, color=black] (V10) -- (V11);
\draw[very thick, color=black] (V10) -- (V13);
\draw[very thick, color=black] (V11) -- (V12);
\draw[very thick, color=black] (V13) -- (V14);
\draw[very thick, color=black] (V14) -- (V15);
\end{tikzpicture}
    }
                      \resizebox{0.19\linewidth}{!}{
    \begin{tikzpicture}[scale=1]
\coordinate (V0) at (2.316, 5.516);
\filldraw[fill=gray] (V0) circle [radius=0.1];
\coordinate (V1) at (1.848, 3.785);
\filldraw[fill=gray] (V1) circle [radius=0.1];
\coordinate (V2) at (8.564, 1.969);
\filldraw[fill=gray] (V2) circle [radius=0.1];
\coordinate (V3) at (9.032, 3.700);
\filldraw[fill=gray] (V3) circle [radius=0.1];
\coordinate (V4) at (3.074, 7.590);
\filldraw[fill=gray] (V4) circle [radius=0.1];
\coordinate (V5) at (1.443, 8.031);
\filldraw[fill=gray] (V5) circle [radius=0.1];
\coordinate (V6) at (0.971, 6.287);
\filldraw[fill=gray] (V6) circle [radius=0.1];
\coordinate (V7) at (2.603, 5.846);
\filldraw[fill=gray] (V7) circle [radius=0.1];
\coordinate (V8) at (0.500, 4.543);
\filldraw[fill=gray] (V8) circle [radius=0.1];
\coordinate (V9) at (2.341, 5.607);
\filldraw[fill=gray] (V9) circle [radius=0.1];
\coordinate (V10) at (2.363, 5.690);
\filldraw[fill=gray] (V10) circle [radius=0.1];
\coordinate (V11) at (9.500, 5.431);
\filldraw[fill=gray] (V11) circle [radius=0.1];
\coordinate (V12) at (2.503, 5.745);
\filldraw[fill=gray] (V12) circle [radius=0.1];
\draw[very thick, color=black] (V0) -- (V1);
\draw[very thick, color=black] (V0) -- (V3);
\draw[very thick, color=black] (V0) -- (V10);
\draw[very thick, color=black] (V1) -- (V2);
\draw[very thick, color=black] (V2) -- (V3);
\draw[very thick, color=black] (V3) -- (V11);
\draw[very thick, color=black] (V4) -- (V5);
\draw[very thick, color=black] (V4) -- (V7);
\draw[very thick, color=black] (V5) -- (V6);
\draw[very thick, color=black] (V6) -- (V7);
\draw[very thick, color=black] (V6) -- (V8);
\draw[very thick, color=black] (V9) -- (V10);
\end{tikzpicture}
    }\\
    \caption{Examples of corrupted ground truth wireframes, used for wireframe ranking. Left to right: GT, deformed edges (deform\_medium), vertex duplication and random movement (perturb\_medium), edge addition (add\_low), edge deletion(remove\_low).}
    \label{fig:corruptions}
\vspace{-1.5em}
\end{figure}

\mypara{Pool2 --- Corrupted ground truth}. We apply one of the following operations on the ground-truth wireframes from Pool 1 -- examples are shown in Fig.~\ref{fig:corruptions}:
\begin{itemize}
    \item {(deform\_\{low, medium, high\}) Split the edge into several edges and perturb the positions of the vertices without breaking the topology.}
    \item {(perturb\_\{low, medium, high\}) Split the vertex into several ones. Each of the new vertices is randomly shifted from the ground truth. If the original vertex was connected to multiple neighbors, randomly decide which of the new vertices is connected to which neighbors. }
    \item {(add\_\{low, medium, high\}) Randomly add wrong edges to the model.}
    \item {(remove\_\{low, medium, high\}) Randomly delete some of the vertices and all the edges connected to them.}
\end{itemize}

\mypara{Unit-tests \& Desired properties of dissimilarity scores}: In addition to being aligned with human judgment, we also propose a set of "unit-tests" for the metrics, which we believe are reasonable, and check if the dissimilarity scores satisfy these requirements. We design tests for the formal properties of mathematical metrics as well as additional properties relevant to evaluating dissimilarity in structured reconstruction tasks:
For example, if wrong edges $E_1$ and $E_2$ are added to the GT model, the metric should score the resulting wireframe lower than if $E_1$ or $E_2$ are added separately.

\mysubpara{Identity of Indiscernibles}: This property ensures that identical inputs receive a dissimilarity score of zero, indicating perfect similarity. For any reconstruction \( x \), a metric \( d \) satisfies this property if \( d(x, x) = 0 \). %

\mysubpara{Symmetry}: A symmetric metric produces the same dissimilarity score regardless of the order of the inputs. For reconstructions \( x \) and \( y \), a metric satisfies symmetry if \( d(x, y) = d(y, x) \). %

\mysubpara{Triangle Inequality}: The triangle inequality ensures that for any three reconstructions \( x \), \( y \), and \( z \), the dissimilarity between \( x \) and \( z \) is less than or equal to the sum of dissimilarities between \( x \) and \( y \), and \( y \) and \( z \). This relationship is expressed as \( d(x, z) \leq d(x, y) + d(y, z) \).
   
\mysubpara{Monotonicity}: This property describes how the dissimilarity score behaves when components (such as vertices or edges) are removed from a reconstruction. 
    A metric satisfies monotonicity if the dissimilarity score does not increase when wrong vertices or edges are deleted. Similarly, the dissimilarity must not increase when correct vertices or edges are added.
    
\mysubpara{Quasi-proportionality}: This property holds when the metric changes smoothly under perturbations. This is evaluated by moving random vertices with small increments and checking the variance of the differences in the score.
We use the following perturbations to simulate better or worse reconstructions:
(i) remove correct edges from the ground truth wireframe; (ii) add wrong edges to the ground truth wireframe; (iii) disconnect ground truth edges; (iv) remove correct vertices; (v) move ground truth vertices to the wrong location.
For every perturbation, we apply it 10 times and declare an example monotonic if it is strictly increasing (or decreasing as appropriate) for those continuous 10 perturbations.

\section{Metrics}
\label{sec:metrics}

The following metrics are considered:

\mypara{WED -- Wireframe Edit Distance} was proposed by Liu~\etal\cite{liu2021_pc2wf} as an extension of the Graph Edit Distance (GED)~\cite{sanfeliu1983distance_ged}. 
GED quantifies the distance between two graphs as the minimum number of elementary operations (inserting and deleting edges and vertices) required to transform one graph into another. 
WED extends this to wireframes (graphs with node positions and edge lengths) and proposes a cheap approximation to the NP-Hard problem of computing the optimal sequence of edits. 
Concretely, an assignment is first computed between the predicted and ground-truth vertices, and a cost is paid proportional to the distance between matched vertices. 
Next, unmatched vertices are deleted, and missing vertices are inserted (paying a cost proportional to the number of inserted/deleted vertices). 
Finally, given the vertex assignments, missing edges are inserted and extra edges deleted, paying a cost proportional to their length. 
In order to use WED, one needs to decide on the cost of insertion and deletion of the vertices, as well as the order of operations, and method of computing vertex/edge assignment. WED was used to determine the winner in the Building3D~\cite{Building3D} and \sttdr~\cite{S23DR} CVPR Challenges.

\mypara{ECD -- Edge Chamfer Distance.} 
ECD is commonly used in structured reconstruction papers~\cite{Huang_2024_CVPR, SepicNet, APC2Mesh}.
We consider a family of chamfer-like metrics between two point sets $A$ and $B$ sampled from wireframe edges. The general form is:
\begin{equation}
d(A,B) := \inf_{\pi_{AB}: A \rightarrow B} \E_{a \in A} \left[f(a, \pi_{AB}(a))\right] \, ,
\end{equation}
where $\pi_{AB}$ represents an assignment from elements in $A$ to elements in $B$, and $f$ is typically an $\ell_p$ norm of the difference between the inputs. Different constraints on $\pi_{AB}$ yield different metrics:

\begin{itemize}
\item The classical chamfer distance corresponds to $\pi_{AB}(a) = \argmin_{b \in B} f(a,b)$, \ie nearest neighbor matching.
\item The most constrained version requires $\pi_{AB}$ to be a bijective matching, which can be computed via the Hungarian algorithm and is equivalent to the Earth Mover's Distance when $f$ is the $\ell_p$ norm.
\end{itemize}
\begin{figure*}[tb]
    \centering
\includegraphics[width=0.88\linewidth]{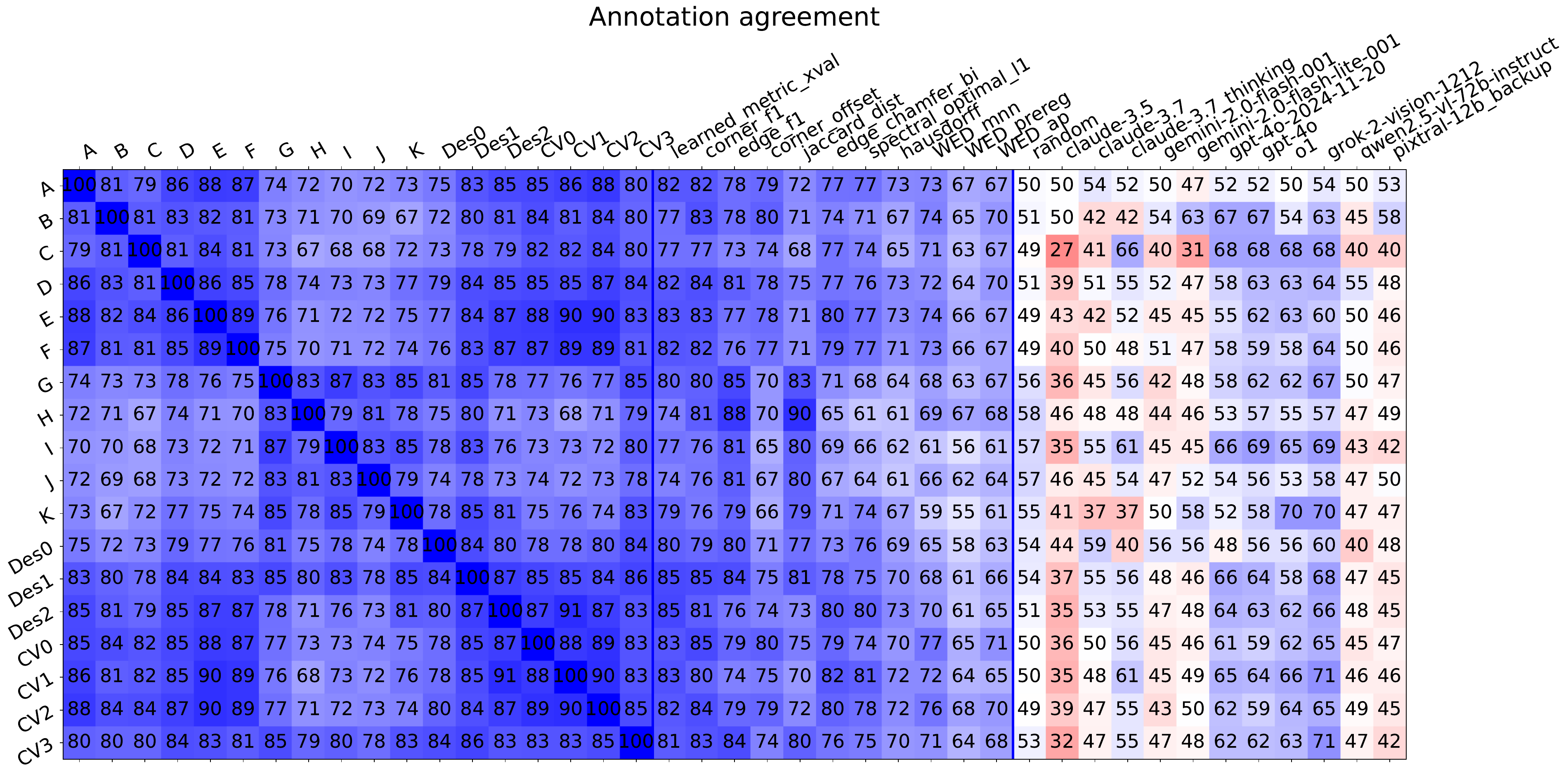}\\
    \caption{Annotator agreement (all pairs). Left to right: annotator agreement with each other, the learned metric, handcrafted metrics, and VLMs. 
    Annotators background: A-K -- 3D modellers, Des[0-2] - designers, CV[0-3] - computer vision engineers. Best zoom-in.}
    \label{fig:human-agreement-all}
    \vspace{-1em}
\end{figure*}
\newcommand{\LWSGD}{SD}

\mypara{Length Weighted Spectral Graph Distance -- \LWSGD} incorporates both topological and geometric information by framing graph (wireframe) distance in terms of distances between the spectra of weighted graph Laplacians. We measure the spectral distance using the 2-Wasserstein metric between the eigenvalue distributions:
\begin{equation}
    \LWSGD(G_1, G_2) := W_2(\lambda(L_1), \lambda(L_2)) \, ,
\end{equation}
where $\lambda(L)$ denotes the spectrum of the Laplacian $L$.
For a graph $G = (V, E)$, the weighted graph Laplacian is defined:
\begin{equation}
    L := D - A
\end{equation}
where $D$ is the weighted degree matrix ($|V| \times |V|$ diagonal matrix with each diagonal entry containing the sum of the lengths of edges incident to that vertex), and $A$ is the weighted adjacency matrix ($|V| \times |V|$ with $A_{ij} = \|\text{coord}(V_i) - \text{coord}(V_j)\|_2$ if $(i,j) \in E$ and $0$ otherwise).

\mypara{Corner and Edge Metrics.} We also compute precision, recall, and F1 scores for both corners and edges. For corners, we consider a prediction correct if it lies within a distance threshold of a ground truth corner. For edges, we use the Hausdorff distance between line segments to determine matches. These metrics provide an intuitive measure of the topological accuracy of the predicted wireframes.

\mypara{Hausdorff Distance} measures the maximum of minimal distances between two sets of points. For wireframes, we sample points along the edges and compute the Hausdorff distance between these point sets, providing a measure of geometric similarity that considers both corner positions and edge geometry.

\mypara{Intersection over Union} is a popular metric in a wide range of fields, \eg segmentation, tracking, object detection.
However, it is rarely used to assess the quality of wireframe reconstructions.
We extend the definition of the wireframe as a set of cylinders with a fixed radius (the only hyperparameter of this metric) and define the metric as an IoU between two sets of cylinders, given by two wireframe reconstructions that need to be compared.
An approximation via point sampling is considered: sample random points from both sets of cylinders and compute the average number of times when the point falls inside of both sets of cylinders. The \textbf{Jaccard distance} is reported between two sets.

\begin{figure*}[!tb]
\centering
\includegraphics[width=0.3\linewidth]{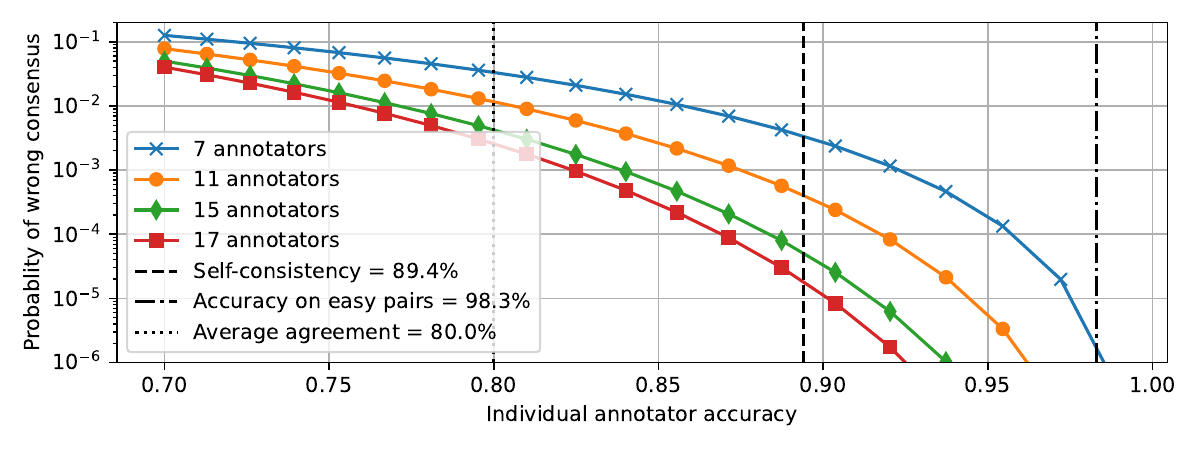}
\includegraphics[width=0.3\linewidth]{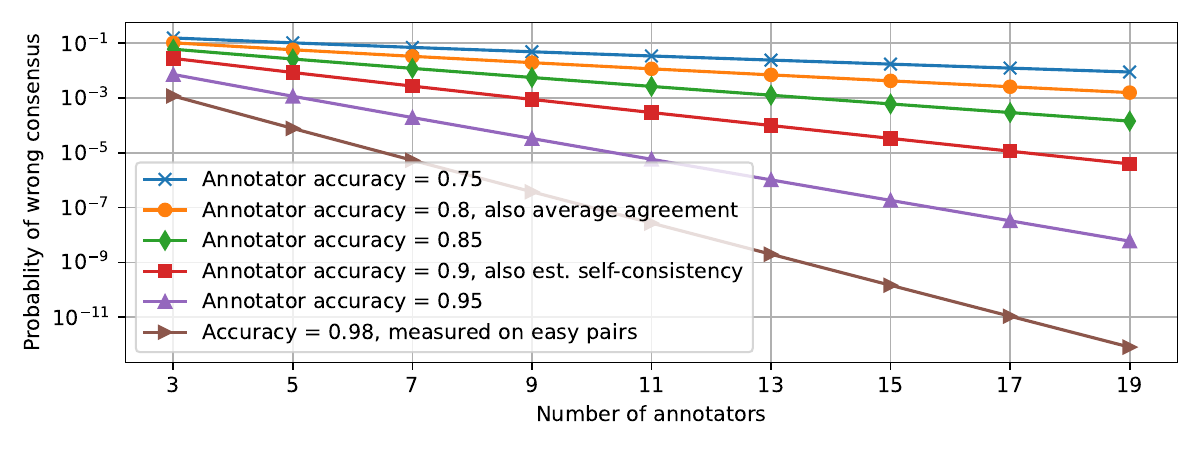}
\includegraphics[width=0.34\linewidth]{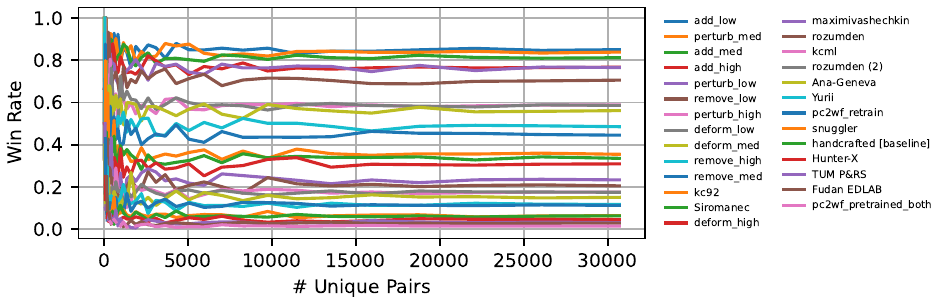}
    \vspace{-1.em}
    \caption{Probability of selecting wrong "winner" depending on number of raters (left), individual accuracy (center), win rate (right).}
    \vspace{-1.6em}
    \label{fig:error-analysis}

    \vspace{-5.4em}
    \scalebox{.4}{
        \hspace{2.155 \linewidth} \includegraphics[width=0.34\linewidth]{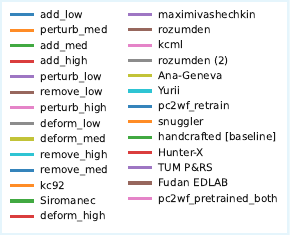}
        }
\end{figure*}

\mypara{Visual-and-language models} could potentially be used for this task. We consider 4o, o1~\cite{openai2024openaio1card}, Grok 2~\cite{grok2}, qwen-2.5~\cite{qwen25}, pixtral 12b~\cite{agrawal2024pixtral12b}, claude 3.5 and 3.7~\cite{claude35}, gemini 2.0 and gemini 2.0-flash~\cite{gemini} models via OpenRouter~\cite{openrouter}.
The prompts are provided in the supplementary. 

\mypara{Learned metric.} We also explore how to distill the human annotations directly into a metric. 
To this end, we propose learning a metric with transfer learning.
First, reconstruction and ground-truth wireframes are plotted in 3D and rendered from a canonical viewpoint (denoted $r_i$). Then, DiNOv2~\cite{oquab2023dinov2} features are extracted from those renderings and an MLP-based regression head is then trained to regress scores based on the extracted features ($g(r_i)$.
A Bradley–Terry~\cite{bradley1952} probability model is assumed and pairwise annotations are used to supervise the training by minimizing a binary cross-entropy loss with a batch size of 16. 10-fold cross validation splits the data such that the sets of ground truth structures and reconstruction methods used in the training and test sets are disjoint. We observe average accuracy across the folds of $76\%$ where a prediction is considered correct if $g(r_\text{winner}) > g(r_\text{loser})$.  %

\section{Experiments}
\label{sec:experiments}
\begin{figure*}[tb]
    \centering
\includegraphics[width=0.24\linewidth]{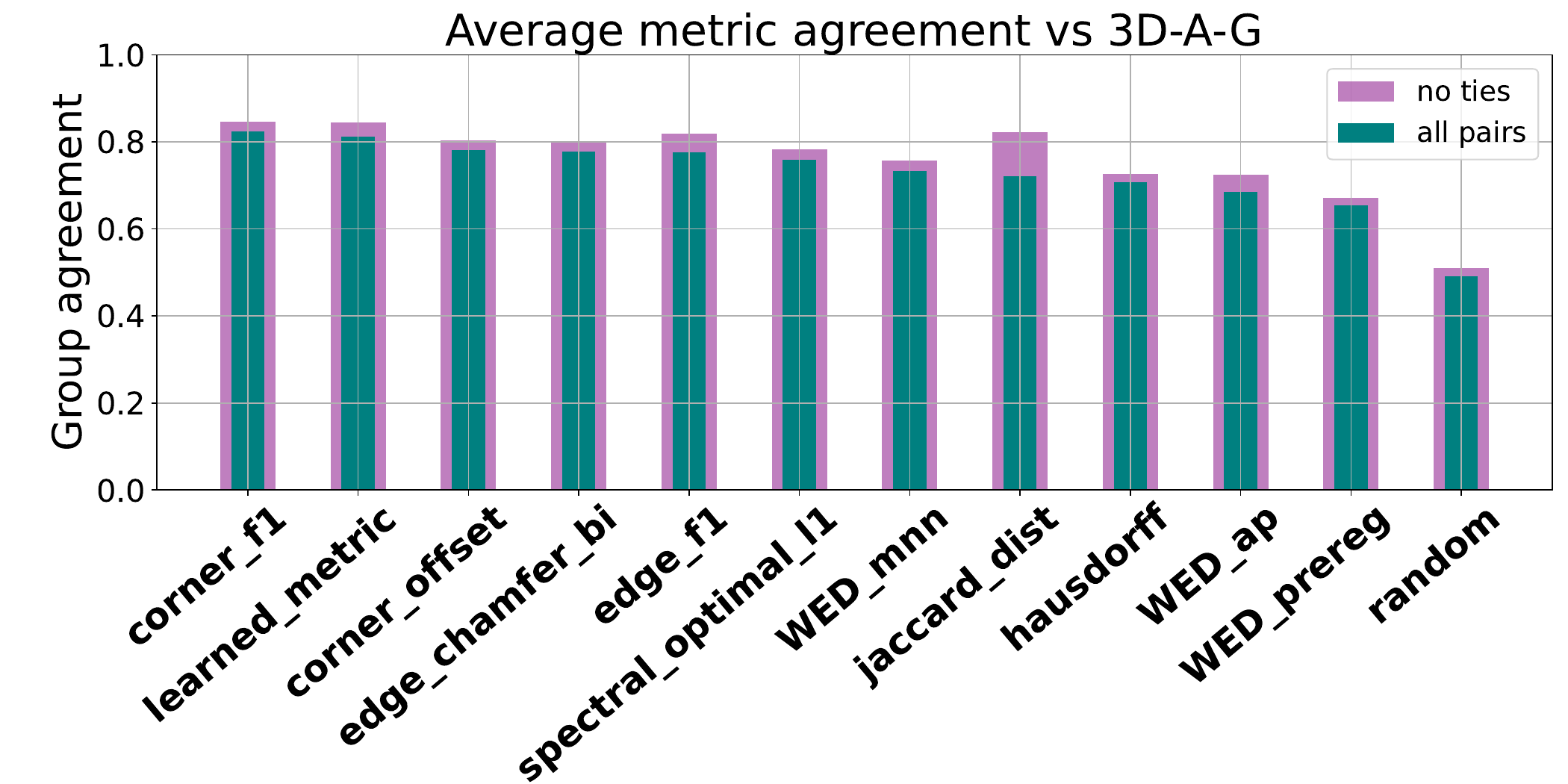}
\includegraphics[width=0.24\linewidth]{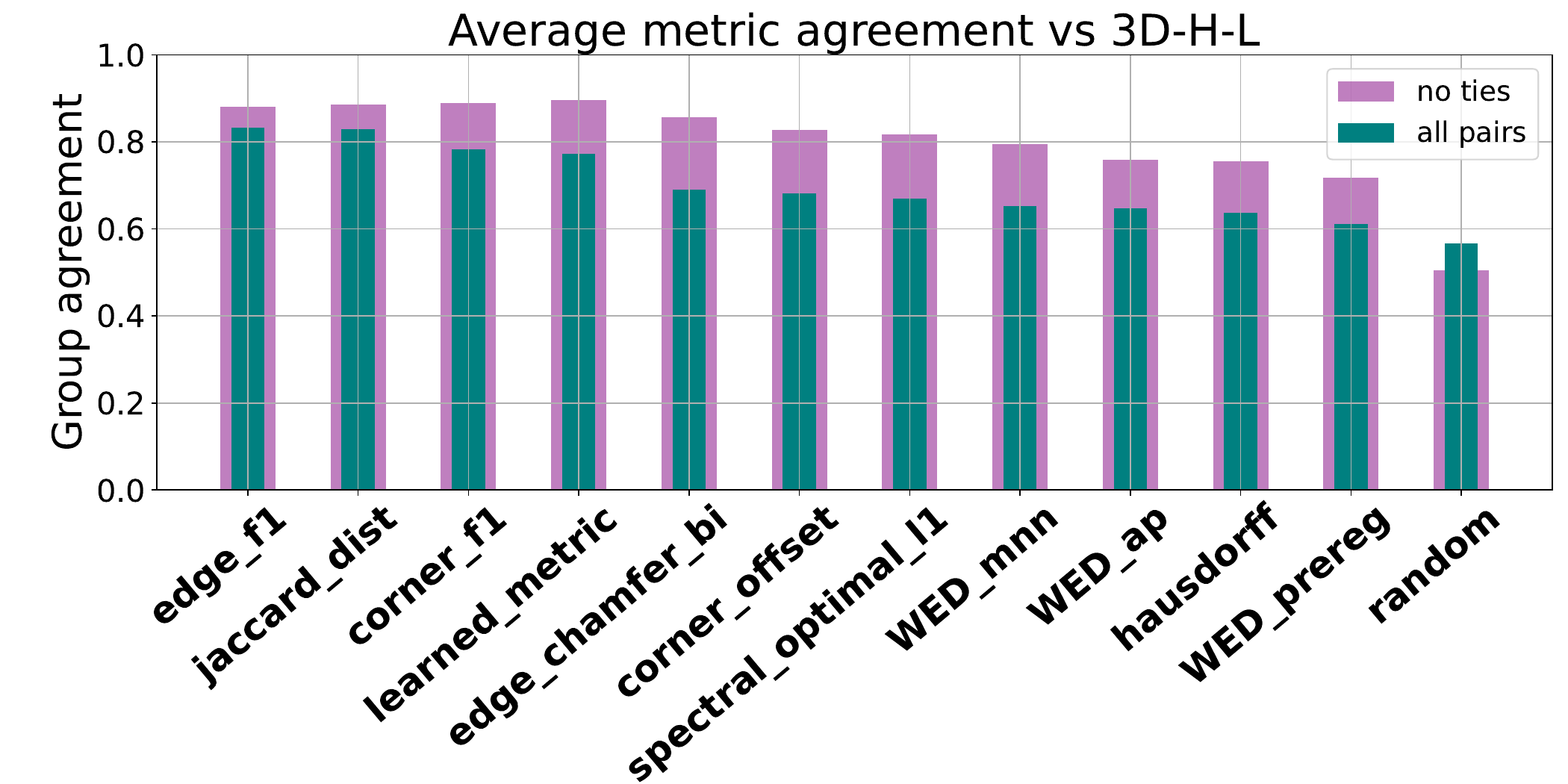}
    \includegraphics[width=0.24\linewidth]{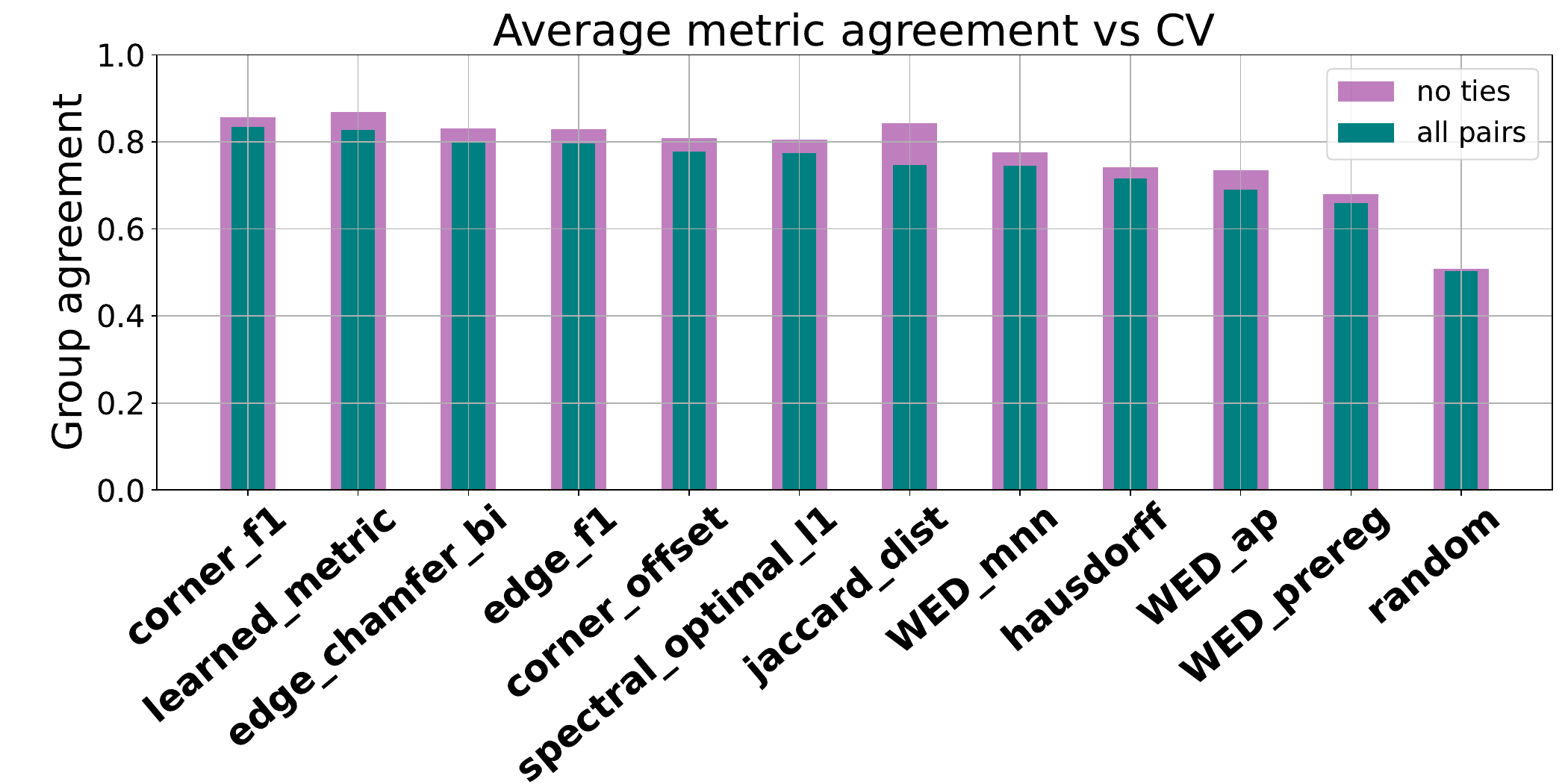}
    \includegraphics[width=0.24\linewidth]{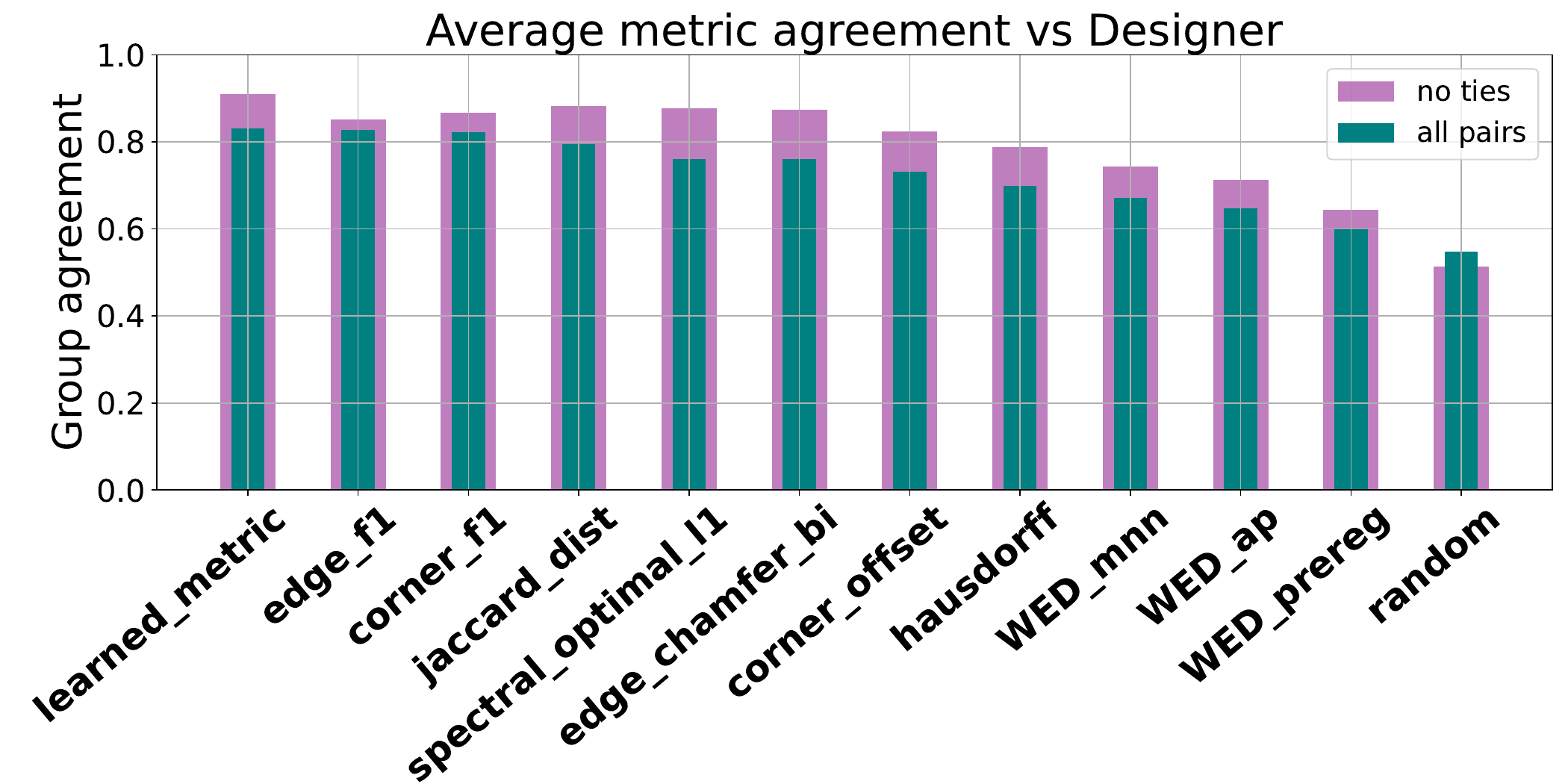}\\
    \caption{Metric ranking by agreement with group in average. Left to right: the first group of raters with more attention to vertices, the second group of raters with more attention to edges, computer vision engineers, and designers.}
    \label{fig:metrics ranking}
\vspace{-1.5em}
\end{figure*}
\mypara{Human wireframe ranking and its consistency.}
To determine which of the metrics under consideration are most appropriate, we employ three groups of people to provide pseudo-ground-truth rankings of the solutions. 
The first and biggest group is made up of human 3D modeling experts who professionally create CAD models of objects from photos. The second group is computer vision researchers, and finally, the third group is people who do not work with 3D modeling in their daily lives (designers). 
There are 11, 4, and 3 people in those groups respectively. 

Annotators were shown pairs of reconstructions of the same structure, and asked to specify which reconstruction most closely matched the superimposed corresponding wireframe. An example of the user interface is shown in Fig.~\ref{fig:ranking-app}. Annotators were able to zoom, pan, and rotate, allowing them to examine the solutions from all sides if needed; the viewpoint of both solutions is synchronized to ease the comparison. All 27 methods are compared exhaustively by every rater, meaning ${27 \choose 2} = 351$ method pairs per house $\times$ 10 houses $=$ 3510 pairs \textit{for every rater}. Most raters rate a few more pairs because of the self-consistency checks. See the Suppl. for the additional information.

\mypara{Rater Reliability.}
We quantify rater reliability using two complementary methods: self-consistency and correctness on synthetic samples.

\mypara{Correctness on Synthetic Samples.} We introduced synthetic pairs of wireframes with known ground-truth rankings based on systematically applied "corruptions."  Each corruption type featured "Low," "Medium," and "High" severity levels. We treat the "low" vs "high" per each corruption type to be obvious enough that if annotators rank them differently, it can be treated as a labeling error, \eg because of wrong clicks. Average rater accuracy on these "easy" pairs is 98.3\%.

\mypara{Self-consistency.} We assessed intra-rater reliability by measuring how consistently annotators rated repeated pairs of wireframes, occasionally reversing pair order to mitigate order biases. These repeated evaluations constituted only a small subset of the total ratings (self-consistency checks are performed with 5\% probability). The average self-consistency score is 89.4\%, which could be partially attributed to the labeling mistakes and partially to the changing preferences during the annotation process, which we experienced ourselves.

\mypara{Do we label enough?}
Under the assumption that there is some latent “correct” winner in any given pair, and $N$ raters each independently select this ``correct'' winner with probability $p$, we can compute the probability that the majority is “wrong” (see Fig.~\ref{fig:error-analysis}). Therefore, the estimated panel error rate per pair is $\approx1\%$ for 11 (expert) raters, and $0.25\%$ for 17 (all) raters (assuming a significant $20\%$ individual error rate). 
We also analyze the stability of our results and present adequacy analysis for the number of raters, comparisons, and houses. For each, we sweep a range of subsample sizes and resample 500 times at each size. For each subsample of a given size, we compute ranking implied by the win rates for that subset and rank correlation (Kendall $\tau$) between the subset ranks and the rankings using the full dataset. We then construct a 95\% Bootstrap CI across the 500 iterates for $\tau$. The minimum number of raters/comparisons/houses needed for $\tau \geq 0.95$: comparisons $\geq$ 3350, houses $\geq$ 4, raters $\geq$ 8 (in all cases we have more than that). 

\mypara{Finding agreement.}
We compute an agreement score for each pair for annotators using the following simple rule: the same ranking gets 1 point, decisive ranking vs "equal" gets 0.5 points, and the opposite ranking gets zero.

The agreement table between human annotators, metrics, and VLMs is shown in Fig.~\ref{fig:human-agreement-all}. 
The agreement table with all "equal" rankings excluded, is shown in the supplementary. %
\begin{obsbox}
When accounting for ties, there is a moderate global consensus among the human annotators; however, they form two distinct clusters that do not seem to depend on the annotator's background. 
One group assigns more weights to the edge accuracy -- correlated with edge F1 and Jaccard distance, and the second -- vertex accuracy, correlated with corner F1 score. 
\end{obsbox}
The average agreement score is around 80\% across all annotators, but within clusters, it increases to 85\%. 
When annotators are decisive (select one of the reconstructions as clearly better rather than selecting "equal"), then the average agreement increases to 91\%, and no clusters are observed.
Agreement with the metrics (middle part of Fig.~\ref{fig:human-agreement-all}) suggests an explanation of preferences.
\vspace{-0.5em}
\begin{obsbox}
Human annotators pay more attention to correct parts of the reconstruction than the incorrect parts. Regardless of whether edges or vertices are considered, recall metrics agree more with human preferences than precision ones. 
\end{obsbox}
\vspace{-0.5em}
Cluster 1 (raters A-G, CV, Des1-2) mostly correlates with corner-based metrics, such as corner recall and corner F1-score, whereas Cluster 2 (raters H-K, Des0) correlates more with edge-based metrics, such as edge recall, edge F1-score and Jaccard distance. The second difference between clusters is that Cluster 2 is more likely to give an "equal" score for the low-quality reconstruction, whereas Cluster 1 tried to rank reconstruction more decisively. 

The metrics rankings w.r.t. different groups of raters is shown in Fig.~\ref{fig:metrics ranking}. 
\vspace{-0.5em}
\begin{obsbox}
 The average agreement with human preferences of the top handcrafted metrics does not vary significantly.
 WED-based scores correlate with annotators the least.
\end{obsbox}
\vspace{-0.5em}
The supplementary material shows the full agreement table. 
Furthermore, the ranking setup changes the human preferences for different metrics; for example, Jaccard distance performs much better for the decisive pairs compared to all other metrics. 
WED with pre-registration, which was used in the \sttdr{} challenge, shows the worst correlation and is just slightly better than random chance. 
Other flavors of the WED metric, namely WED\_mnn and WED\_AP (used in Building3D Challenge), perform better but still worse than the rest. Graph structure metrics are in the middle.
We also consider the agreement with VLMs. Despite initial success with individual examples, in our tests they did not perform meaningfully better than chance.
\begin{obsbox}
VLMs do not show significant agreement with human preferences in the wireframe ranking. The only exceptions are OpenAI models, as well as Grok2; yet those are only slightly better than chance. VLMs agree the most with the WED metrics family. 
\end{obsbox}

\mypara{Finding the best reconstruction.}
To determine if there is a single "quality" factor which explains the human expert judgments, we employ three distinct approaches to map these pairwise comparisons to a single ranking of the methods: simple win rates, a Bradley-Terry probability model, and a factor analysis based approach. While the methods differ, they point to highly concordant conclusions.
\begin{obsbox}
Human raters tend to rank equally all solutions that are below some quality threshold. Having no solution often ranks better, compared to a totally wrong reconstruction. 
\end{obsbox}

\mypara{Simple Win Rate}
The first approach is similar to chess scoring -- first a win-count table is computed. For each rated pair, each method receives 1 point for a win, 0.5 for a tie, and 0 for a loss. This scoring does not break ties but distributes the points evenly. 
Results with selected reconstruction methods are shown in Fig.~\ref{fig:metrics ranking}.
\begin{figure}[tb]
    \centering
  \includegraphics[width=0.8\linewidth]{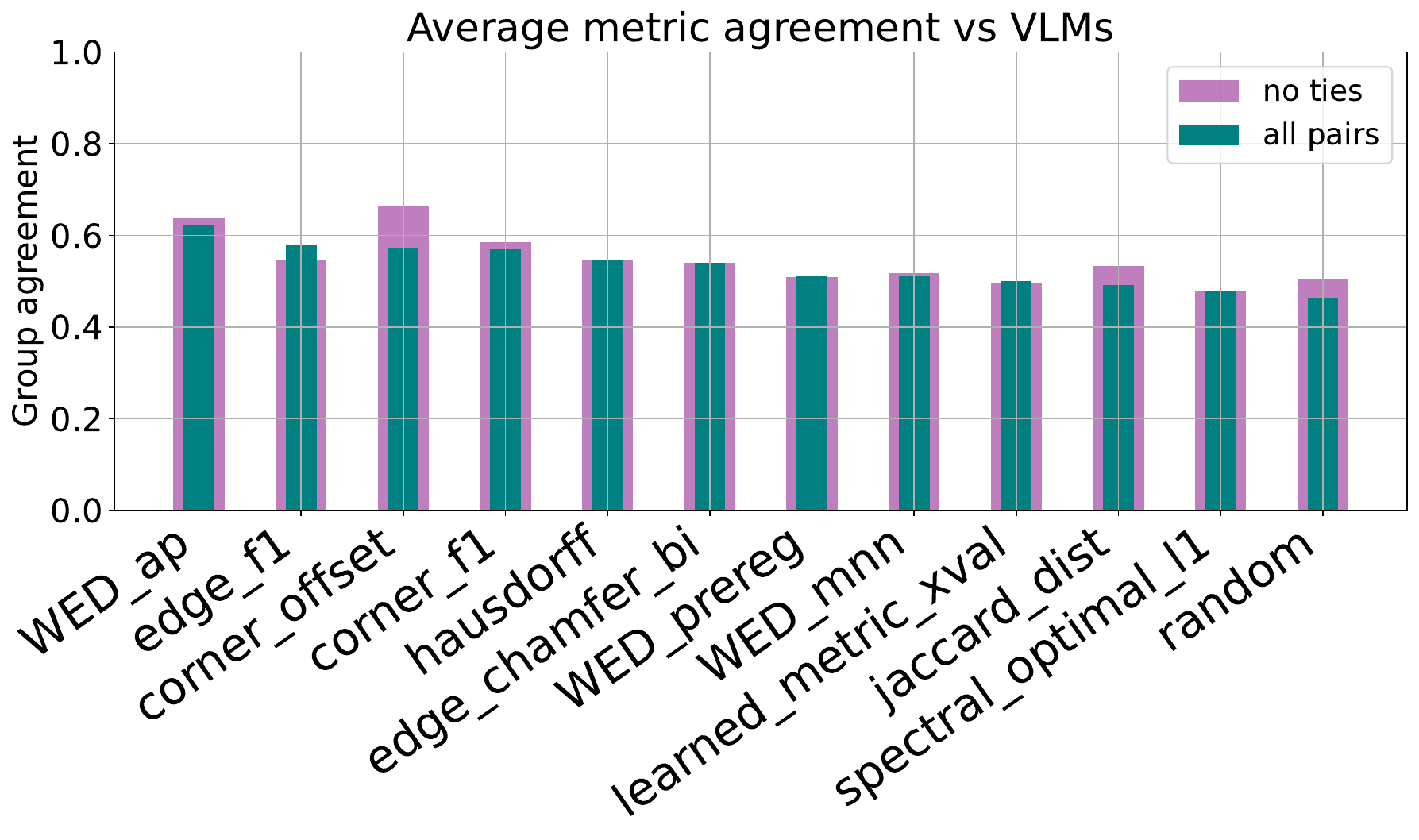}
  \vspace{-1em}
    \caption{Metric ranking by agreement with  VLMs}
    \label{fig:metrics ranking}
\vspace{-1em}
\end{figure}
Consistent with the metric-human agreements, recall plays a more important role, and the wireframes with perfect recall -- "add\_low" and "add\_med" are among the leaders in all groups. 
In other words, extra (erroneous) edges are considered less of an issue when compared to missing an edge or a vertex -- the "remove\_*" family. One possible explanation resides in the information contained in the reconstruction: if all correct edges are present, we may identify and remove any erroneous ones. 
However, in many cases, inferring the exact position of an edge or vertex is not possible given the reconstruction alone if key information is missing.

The next set of solutions contains roughly correct but slightly noisy reconstructions - "perturb\_*" and "deform*" family. 
One of the best solutions from the S23DR Challenge ranked better than highly deformed ground truth but worse than the less invasive corruptions. 

The rest of the reconstructions get almost equal scores due to the high proportion of draws and lack of wins among themselves. 

\mypara{Bradley-Terry Model} We have also modeled the quality of each solution using a Bradley-Terry (BT)~\cite{zermelo1929,bradley1952,elo1978,glickman2013} preference model on the expressed preferences of the annotators (using the BT-Abilities as scores).
The Bradley-Terry model defines the probability that item \(i\) is preferred over item \(j\) as:
\begin{equation}
P(i > j) = \frac{a_i}{a_i + a_j} \, ,
\end{equation}
where \(a_i, a_j\) are positive real numbers representing the latent strength of each item.

Following standard practice, we reparameterize these latent strengths as exponentials of real-valued parameters \(\theta_i\), giving \(a_i = e^{\theta_i}\). This yields:
\begin{align}
    P(i > j) &= \frac{e^{\theta_i}}{e^{\theta_i} + e^{\theta_j}} = \frac{1}{1 + e^{-(\theta_i - \theta_j)}} 
    = \sigma(\theta_i - \theta_j) \, ,
\end{align}
where \(\sigma(x)\) is the sigmoid function \(\sigma(x) = \frac{1}{1 + e^{-x}}\).

This formulation is further generalized by introducing a scale parameter \(s\) and offset \(o\):
\begin{equation}
p_{ij} = \sigma\left(\frac{\theta_i - \theta_j}{s} + o\right) \, .
\end{equation}
With \(s=1\) and \(o=0\), this reduces to the standard Bradley-Terry formulation. Alternatively, setting \(s=400\) and \(o=800\) yields the Elo scoring system familiar to chess players.

To estimate the latent abilities \(\theta\), we initialize each \(\theta_i\) by sampling from an independent Gaussian distribution and then iteratively minimize the expectation of the following binary cross-entropy loss using stochastic gradient descent (SGD) with the Adam optimizer:
\begin{equation}
\mathcal{L} = \mathbb{E}_{(i,j)}\left[-y\log(p_{ij}) - (1 - y)\log(1 - p_{ij})\right] \, ,
\end{equation}
where \(y = 1\) if item \(i\) was chosen over item \(j\), and \(y = 0\) otherwise.

\mypara{Factor Analysis} To investigate whether the data reflect a single underlying dimension of quality, we additionally peruse a factor analysis based approach. We form the methods-by-raters table $M$ such that $M_{kl}$ is the rate at which rater $k$ chose method $l$ when they saw it. 
We hypothesize the empirical log-odds of these win-rates (rate $l$ wins according to $k$), $\eta = \log\frac{M}{1 - M}$
possess a low-rank structure (rank one in the ideal case); this would indicate a single dominant factor ("quality") governing outcomes. We apply singular value decomposition (SVD) to factorize $\eta = U\Sigma V^T$ and 
extract the first left singular vector of $\eta$ containing the estimated quality scores. 

\begin{obsbox}
We find a Kendall correlation coefficient $\mathord{>}0.7$ between the rankings implied by SVD and those implied by BT. This lends additional evidence to the hypothesis that there is a true "quality" factor driving the raters' views. The result is shown in Table~\ref{tab:elo}.
\end{obsbox}

\begin{table}
\scriptsize
\begin{tabular}{l@{}c@{\hspace{.5em}}c@{\hspace{.5em}}cc@{\hspace{.5em}}c@{\hspace{.5em}}c}
\toprule
Method & \makecell{Empirical\\Win Rate} & \makecell{Implied Win\\Rate (BT)} & \makecell{Implied Win\\Rate (Elo)} & \makecell{BT\\Ability} & \makecell{Elo\\Score} & \makecell{Quality\\Factor} \\
\midrule
add\_low & 0.89 & 0.89 & 0.89 & 2.79 & 1937 & 0.03 \\
add\_med & 0.86 & 0.86 & 0.86 & 2.47 & 1769 & 0.02 \\
perturb\_med & 0.85 & 0.85 & 0.85 & 2.33 & 1739 & 0.02 \\
add\_high & 0.82 & 0.82 & 0.82 & 2.04 & 1604 & -0.02 \\
perturb\_low & 0.79 & 0.79 & 0.79 & 1.79 & 1510 & -0.01 \\
remove\_low & 0.79 & 0.78 & 0.79 & 1.71 & 1498 & -0.02 \\
perturb\_high & 0.67 & 0.67 & 0.68 & 0.83 & 1144 & -0.09 \\
deform\_med & 0.67 & 0.67 & 0.66 & 0.83 & 1107 & -0.09 \\
deform\_low & 0.66 & 0.66 & 0.66 & 0.78 & 1094 & -0.10 \\
remove\_high & 0.65 & 0.65 & 0.65 & 0.67 & 1077 & -0.10 \\
remove\_med & 0.63 & 0.64 & 0.64 & 0.60 & 1027 & -0.11 \\
kc92 & 0.51 & 0.51 & 0.51 & -0.25 & 698 & -0.18 \\
Siromanec & 0.50 & 0.50 & 0.49 & -0.34 & 645 & -0.19 \\
deform\_high & 0.50 & 0.50 & 0.50 & -0.34 & 669 & -0.19 \\
maximivashechkin & 0.39 & 0.39 & 0.39 & -1.01 & 386 & -0.27 \\
rozumden & 0.38 & 0.38 & 0.38 & -1.10 & 373 & -0.28 \\
kcml & 0.35 & 0.35 & 0.35 & -1.28 & 291 & -0.26 \\
rozumden & 0.34 & 0.34 & 0.33 & -1.35 & 236 & -0.26 \\
Ana-Geneva & 0.32 & 0.32 & 0.32 & -1.47 & 221 & -0.25 \\
pc2wf\_retrain & 0.29 & 0.29 & 0.30 & -1.65 & 157 & -0.23 \\
Yurii & 0.29 & 0.29 & 0.29 & -1.63 & 146 & -0.25 \\
snuggler & 0.25 & 0.25 & 0.24 & -1.92 & 15 & -0.25 \\
baseline & 0.25 & 0.25 & 0.25 & -1.91 & 19 & -0.25 \\
Hunter-X & 0.22 & 0.22 & 0.22 & -2.10 & -43 & -0.25 \\
TUM & 0.22 & 0.22 & 0.22 & -2.13 & -48 & -0.26 \\
Fudan EDLAB & 0.21 & 0.21 & 0.21 & -2.18 & -76 & -0.26 \\
pc2wf\_pretrained & 0.20 & 0.20 & 0.20 & -2.28 & -117 & -0.25 \\
\bottomrule
\end{tabular}
\vspace{-1em}
\caption{Win rates for selected wireframe, according to the pairwise human annotations and estimated Elo.}
\label{tab:elo}
\vspace{-1em}
\end{table}

\begin{figure*}[tb]
    \centering
    \includegraphics[width=0.249\linewidth]{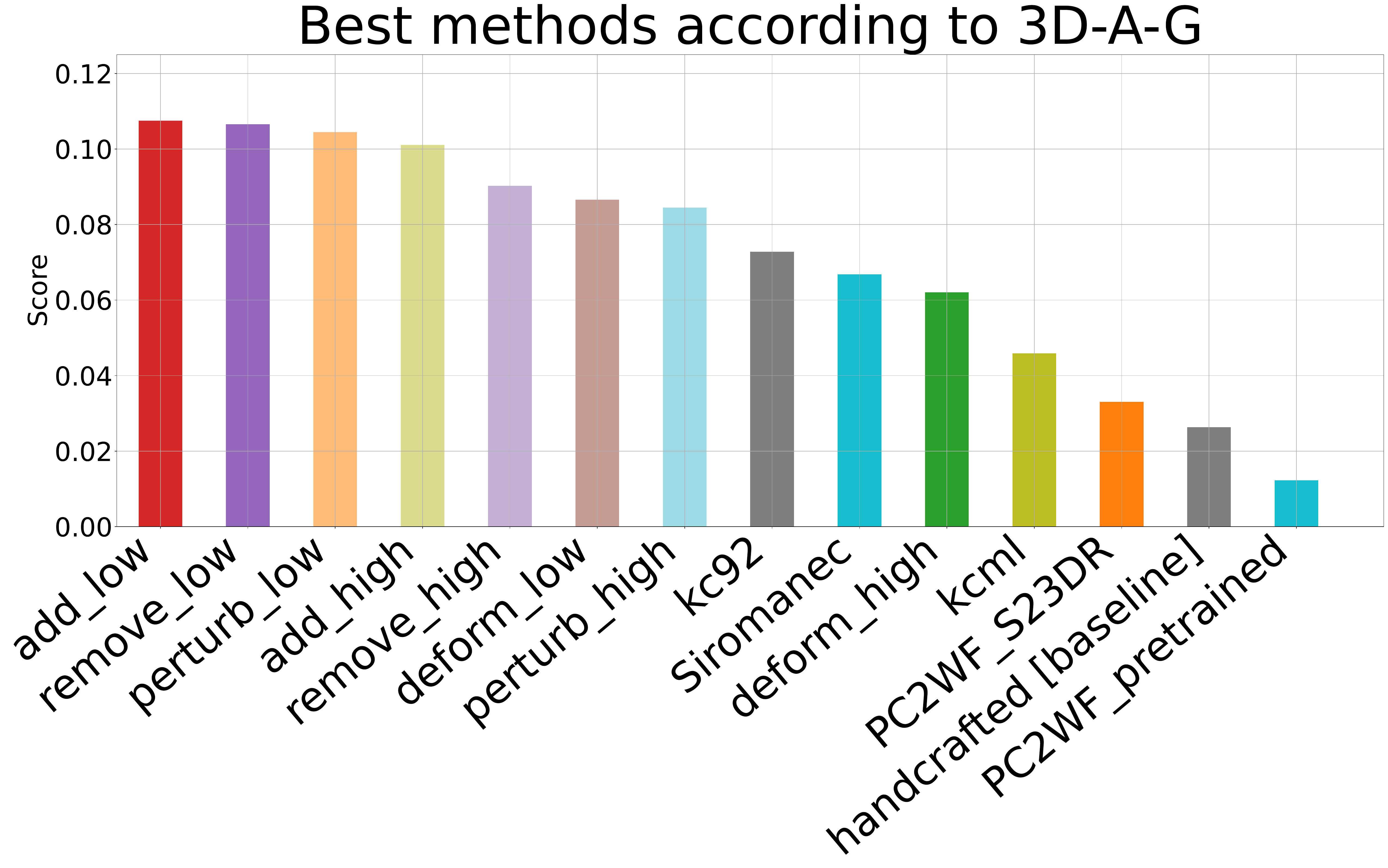}
    \hspace{-0.02\linewidth} \includegraphics[width=0.25\linewidth]{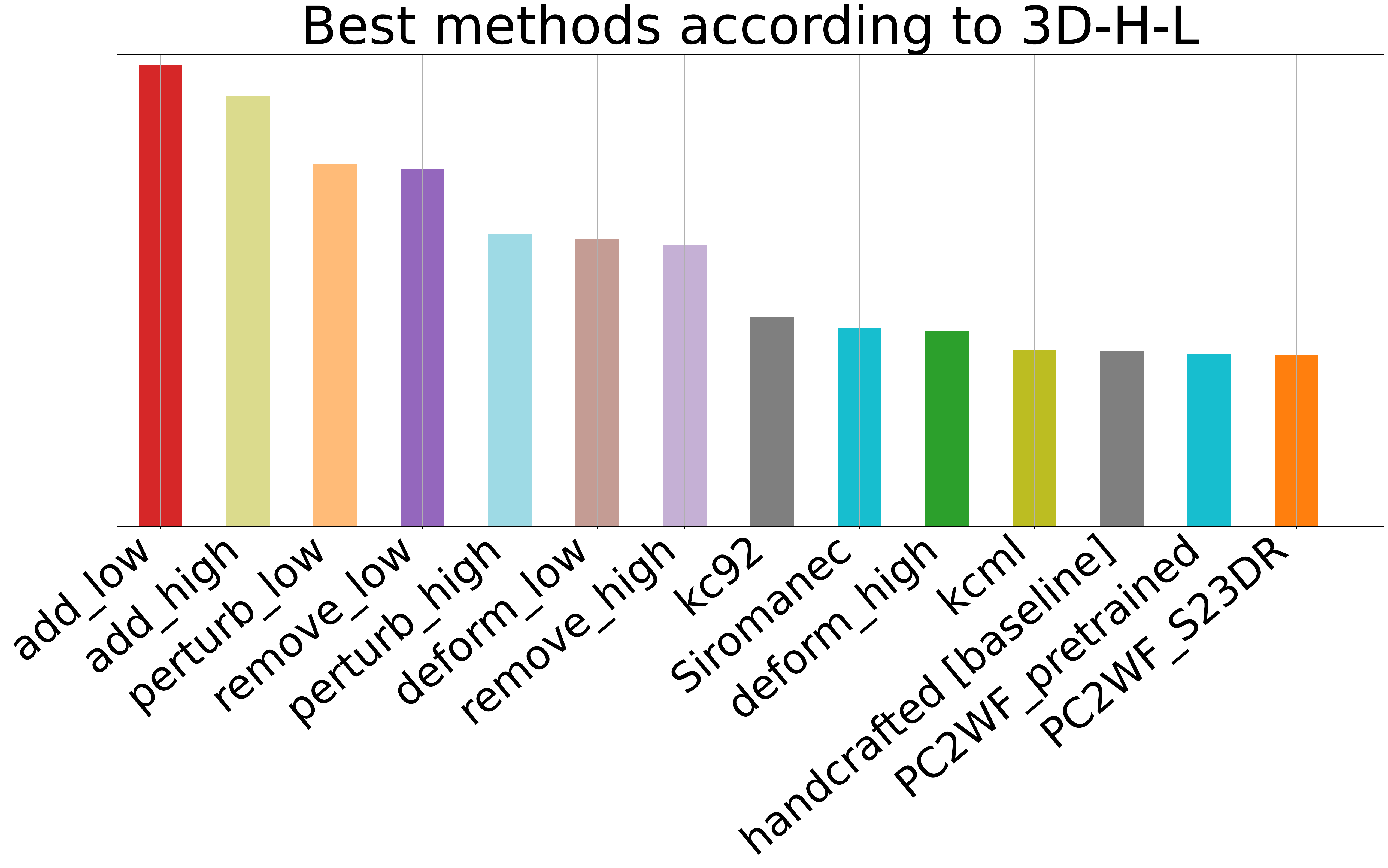}
    \hspace{-0.02\linewidth} \includegraphics[width=0.25\linewidth]{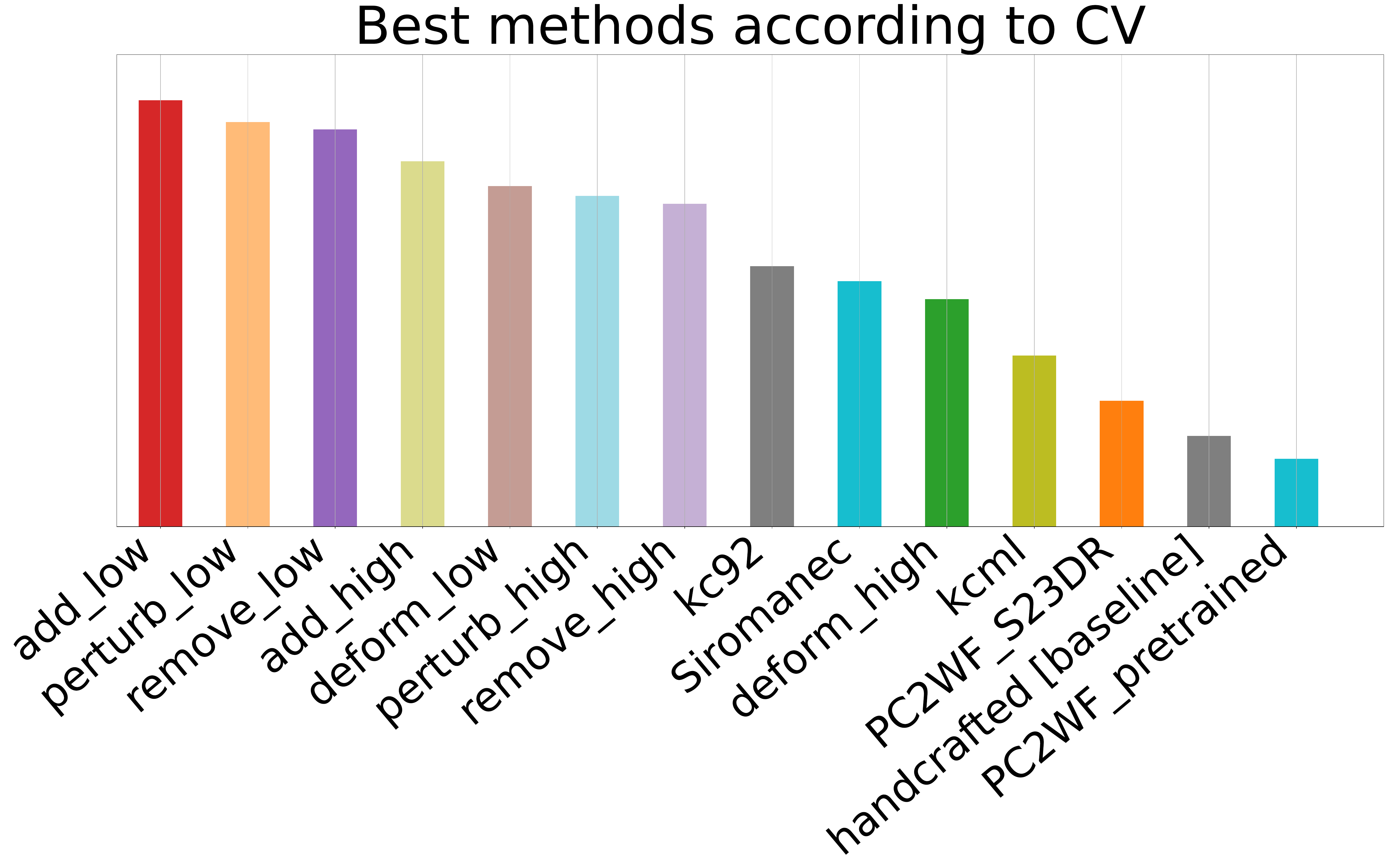}
    \hspace{-0.02\linewidth} \includegraphics[width=0.25\linewidth]{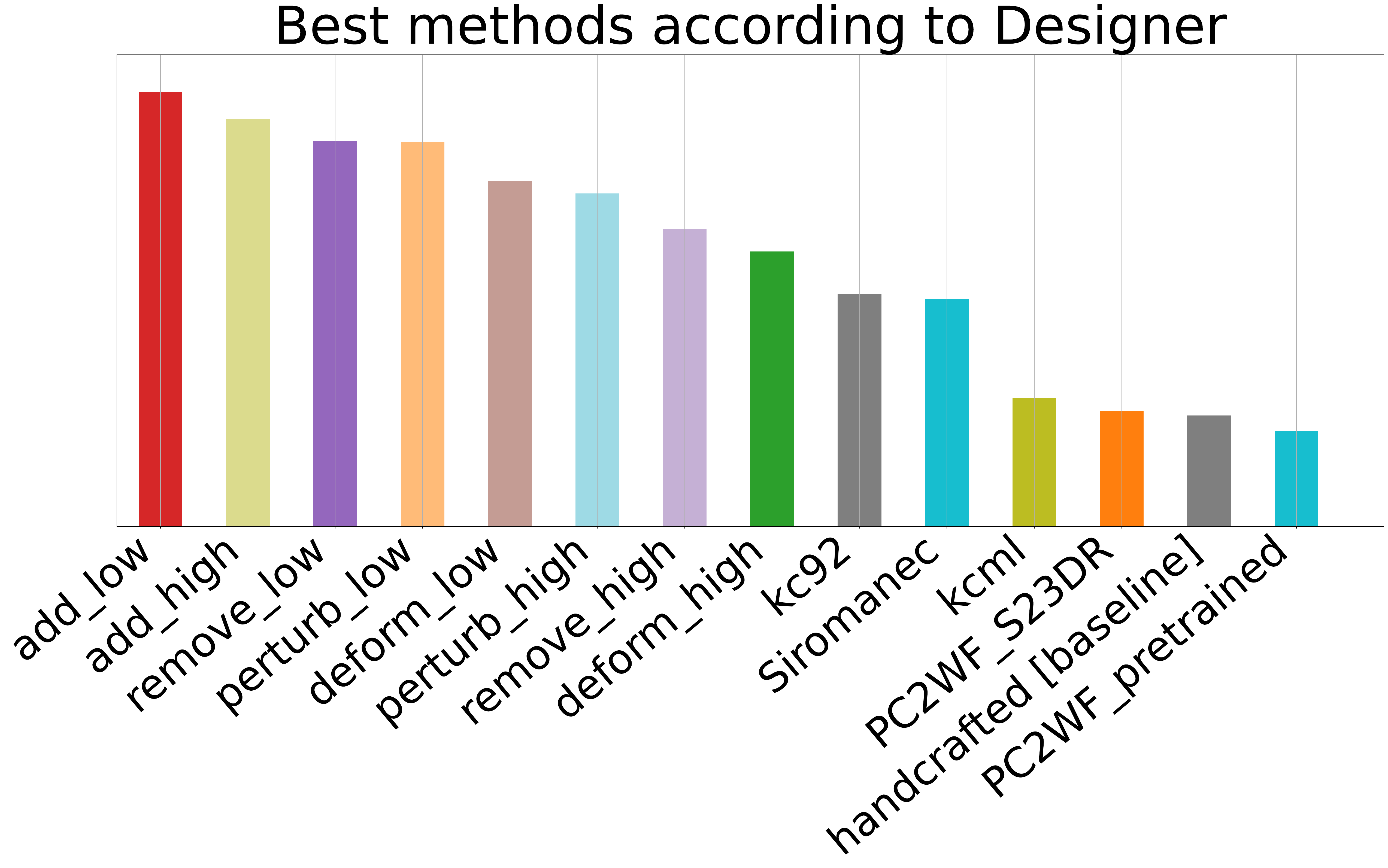}%
\vspace{-0.5em}
    \caption{Methods scores according to different groups.}
\vspace{-1em}
    \label{fig:metrics ranking}
\end{figure*}
\newcommand{\winner}[1]{\textbf{#1}}
\newcommand{\szz}{\scriptsize}
\begin{table*}
\centering
\scriptsize
\setlength{\tabcolsep}{2pt} %
\newcommand{\MySkip}{\hskip 1.3em}  %
\begin{tabular}{lcccccccccccccccc}
\toprule
 & \multicolumn{4}{c}{Corner} & \multicolumn{3}{c}{Edge} & \multicolumn{4}{c}{WED} & \multicolumn{2}{c}{Spectral} & \multicolumn{1}{c}{IoU} & \multicolumn{1}{c}{Hausdorff} & \multicolumn{1}{c}{Chamfer} \\
Test &Prec & Rec & F1 & offset & Prec & Rec & F1 & prereg & MNN & nearest & AP & L1 & L2 & Jaccard & dist. & edge \\
\cmidrule(lr){2-5} \cmidrule(lr){6-8} \cmidrule(lr){9-12} \cmidrule(lr){13-14} \cmidrule(lr){15-15} \cmidrule(lr){16-16} \cmidrule(lr){17-17}
\midrule
\multicolumn{17}{l}{\textbf{Monotonic}} \\
Monotonic (wrong edges) & \textcolor{red}{0.02} & \textcolor{red}{0.01} & \textcolor{red}{0.02} & 1.00 & 1.00 & \textcolor{red}{0.00} & 1.00 & \textcolor{red}{0.00} & \textcolor{red}{0.00} & \textcolor{red}{0.00} & 0.98 & \textcolor{red}{0.63} & \textcolor{red}{0.14} & \textcolor{red}{0.01} & \textcolor{red}{0.09} & \textcolor{red}{0.00} \\
Monotonic (deform/split) & 1.00 & \textcolor{red}{0.86} & 1.00 & \textcolor{red}{0.65} & 1.00 & 0.98 & 1.00 & \textcolor{red}{0.74} & \textcolor{red}{0.79} & \textcolor{red}{0.67} & \textcolor{red}{0.60} & \textcolor{red}{0.00} & \textcolor{red}{0.06} & \textcolor{red}{0.16} & \textcolor{red}{0.05} & \textcolor{red}{0.40} \\
Monotonic (moving vertex) & 1.00 & 1.00 & 1.00 & 0.98 & 1.00 & 1.00 & 1.00 & 0.99 & 1.00 & 0.99 & 1.00 & 0.99 & 0.99 & \textcolor{red}{0.45} & 0.99 & 1.00 \\
Monotonic (disconnect edges) & \textcolor{red}{0.52} & \textcolor{red}{0.00} & \textcolor{red}{0.50} & \textcolor{red}{0.31} & \textcolor{red}{0.00} & \textcolor{red}{0.00} & \textcolor{red}{0.00} & \textcolor{red}{0.26} & 1.00 & 1.00 & 1.00 & \textcolor{red}{0.00} & \textcolor{red}{0.01} & 1.00 & \textcolor{red}{0.00} & \textcolor{red}{0.02} \\
Monotonic (delete vertices) & \textcolor{red}{0.00} & 1.00 & 1.00 & 1.00 & \textcolor{red}{0.00} & 1.00 & 1.00 & \textcolor{red}{0.75} & 1.00 & 1.00 & 1.00 & 0.90 & \textcolor{red}{0.33} & \textcolor{red}{0.67} & \textcolor{red}{0.14} & 0.99 \\
Monotonic (delete edges) & \textcolor{red}{0.00} & \textcolor{red}{0.00} & \textcolor{red}{0.00} & 1.00 & \textcolor{red}{0.00} & 1.00 & 1.00 & 1.00 & 1.00 & 1.00 & 1.00 & 1.00 & \textcolor{red}{0.31} & 1.00 & \textcolor{red}{0.07} & \textcolor{red}{0.83} \\
\midrule
\multicolumn{17}{l}{\textbf{Identity}} \\
Identity of indiscernibles & 1.00 & 1.00 & 1.00 & 1.00 & 1.00 & 1.00 & 1.00 & \textcolor{red}{0.02} & 1.00 & 1.00 & 1.0& 1.00 & 1.00 & 1.00 & \textcolor{red}{0.00} & \textcolor{red}{0.00} \\
Near identity of indiscernibles & 1.00 & 1.00 & 1.00 & 1.00 & 1.00 & 1.00 & 1.00 & 1.00 & 1.00 & 1.00 & 1.0 & 1.00 & 1.00 & 1.00 & \textcolor{red}{0.00} & \textcolor{red}{0.00} \\
\midrule
\multicolumn{17}{l}{\textbf{Symmetry}} \\
Symmetry (0 mean, weighted) & 1.00 & 1.00 & 1.00 & \textcolor{red}{0.88} & 1.00 & 1.00 & 1.00 & \textcolor{red}{0.44} & \textcolor{red}{0.44} & \textcolor{red}{0.44} & \textcolor{red}{0.44} & \textcolor{red}{0.44} & 0.90 & 1.00 & \textcolor{red}{0.84} & 1.00 \\
Near symmetry (0 mean, weighted) & 1.00 & 1.00 & 1.00 & \textcolor{red}{0.87} & 1.00 & 1.00 & 1.00 & \textcolor{red}{0.58} & \textcolor{red}{0.54} & \textcolor{red}{0.54} & \textcolor{red}{0.45} & 0.90 & 0.92 & 1.00 & 1.00 & 1.00 \\
Symmetry (shift, weighted) & 1.00 & 1.00 & 1.00 & 1.00 & 1.00 & 1.00 & 1.00 & \textcolor{red}{0.88} & 0.99 & 0.92 & 0.98 & 0.99 & 1.00 & 1.00 & \textcolor{red}{0.82} & 1.00 \\
Near symmetry (shift, weighted) & 1.00 & 1.00 & 1.00 & 1.00 & 1.00 & 1.00 & 1.00 & 1.00 & 1.00 & 0.92 & 0.99 & 1.00 & 1.00 & 1.00 & 1.00 & 1.00 \\
\midrule
\multicolumn{17}{l}{\textbf{Quasi-proportionality}} \\
Quasi-proportionality (shift, far) & \textcolor{red}{0.00} & \textcolor{red}{0.00} & \textcolor{red}{0.00} & \textcolor{red}{0.00} & \textcolor{red}{0.00} & \textcolor{red}{0.00} & \textcolor{red}{0.00} & 1.00 & \textcolor{red}{0.00} & \textcolor{red}{0.00} & \textcolor{red}{0.00} & 1.00 & \textcolor{red}{0.72} & \textcolor{red}{0.05} & \textcolor{red}{0.00} & \textcolor{red}{0.00} \\
Quasi-proportionality (shift, close) & \textcolor{red}{0.00} & \textcolor{red}{0.00} & \textcolor{red}{0.00} & \textcolor{red}{0.00} & \textcolor{red}{0.00} & \textcolor{red}{0.00} & \textcolor{red}{0.00} & 1.00 & \textcolor{red}{0.16} & \textcolor{red}{0.16} & 0.91 & 1.00 & \textcolor{red}{0.72} & \textcolor{red}{0.06} & \textcolor{red}{0.00} & \textcolor{red}{0.00} \\
\midrule
\multicolumn{17}{l}{\textbf{Triangle ineq}} \\
Triangle ineq. (rand other) & 0.99 & 1.00 & 1.00 & 1.00 & 1.00 & 0.98 & 1.00 & 1.00 & 1.00 & 0.99 & 0.95 & 0.92 & 0.97 & 1.00 & 1.00 & 0.96 \\
Triangle ineq. (add noise) & 1.00 & 1.00 & 1.00 & 0.91 & 1.00 & 1.00 & 1.00 & \textcolor{red}{0.84} & \textcolor{red}{0.81} & \textcolor{red}{0.83} & 1.00 & 0.90 & \textcolor{red}{0.69} & 1.00 & 1.00 & 1.00 \\
Triangle ineq. (del1/del2) & 1.00 & 1.00 & 1.00 & 1.00 & 1.00 & 1.00 & 1.00 & 0.90 & 0.98 & 0.98 & 1.00 & 1.00 & \textcolor{red}{0.66} & 1.00 & 1.00 & \textcolor{red}{0.43} \\
 \midrule
\textbf{Pass count} & 11/17 & 11/17 & 12/17 & 11/17 & 12/17 & 13/17 & 14/17 & 8/17 & 10/17 & 10/17 & 13/17 & 13/17 & 8/17 & 11/17 & 6/17 & 8/17 \\
\bottomrule
\end{tabular}

\caption{Properties "unit-test" results for metrics, percentage of tests passed. The test is passed if the result is $\geq90\%$ (black).}
\label{tab:properties}
\vspace{-1.5em}
\end{table*}
\mypara{Metric properties and "unit tests".} 
We define a range of properties a good metric should have, implement tests for these properties, and report the results for all metrics. We use a dataset of 128 ground truth wireframes, which we disturb or alter and check the behavior of each metric.
We report the percentage of wireframes where the property was valid for a given metric. 
Results are shown in Table~\ref{tab:properties}. Most of the metrics pass the triangle inequality test, the identity of indiscernibles, and symmetry. None of the metrics is perfectly monotonic w.r.t. wireframe changes, which makes sense; for example, the precision metrics are insensitive to the number of predicted vertices/edges as long as each predicted element is aligned with an element of the ground truth. Hausdorff and WED with pre-registration pass the fewest tests, which is in line with their poor performance w.r.t. human preferences. Conversely -- corner F1, edge F1, and Jaccard perform well and align well with the preferences of annotators. The spectral L1 distance and Chamfer edge distance are exceptions -- the spectral distance scores well on properties, but not human alignment, and Chamfer one -- vice versa. 

\mypara{Additional considerations.} After speaking to the human annotators, we note the following.
First, the 3D reconstruction experts are considering how the estimated reconstruction could help them in 3D modeling. There are two main approaches to using such help. The first (in case of noisy reconstructions) is to use the wireframe as a guide to create their own clean reconstruction. The second, if the reconstruction is already close enough, is to fix the wireframe to produce the final result. 
The Wireframe Edit Distance is designed to estimate how costly it would be to modify the predicted wireframe to match the ground truth. 
The issue with it is the set of operations -- WED only considers vertex/edge deletion/insertion and vertex movement. 
In practice, one could fit a single edge to multiple noisy ones, bulk-delete a lot of wrong edges or vertices, and apply a rigid transform on the whole model. All of these operations are commonly used in 3D editing software, and WED could benefit from them. 

Finally, if the wireframe is totally wrong, then it is better to have no reconstruction at all and start from scratch. In this "low quality regime", most human annotators see no difference between reconstructions if both are wrong. 

Considering usage in competition and benchmarks, we would recommend the use of F1-score, despite the fact that recall-based metrics are more in line with human preferences, as it is less easy to game.
For example, a dense grid of vertices and edges could score perfectly on recall but be useless in practice and score poorly on precision-based metrics. 
\vspace{-0.3em}
\section{Conclusion} 
\label{sec:conclusion}
\vspace{-0.3em}
We have studied how human preferences in structured reconstruction evaluation are explained via a wide range of metrics. We show that human preferences can be learned from a small number of examples by transferring from pretrained models which can subsequently be used to score unseen reconstructions. However, we conclude that additional study is warranted prior to relying on such learned metrics as the sole adjudication mechanism for competitions (especially those with strong incentives), because of the potential for reward hacking, gradient-based adversarial attacks, and the like. Based on our study, we recommend using a combination of edge-based (edge F1 or Jaccard score) and corner-based metrics (F1) for benchmarks and competitions. They better explain human preferences in ranking structured reconstructions than more complex and fragile graph-based metrics such as WED or spectral distances. %
\newpage
\mypara{Acknowledgements}
We thank Sherwin Dale Sotto, Rosalie Cellacay, John Carlo Castro, Johsua Vernon Awog, Romeo Sapitanan, Francis Angel Viloria, Joenard Sarmiento, Danica B. Mercado, Zander Hechanova, Michael James Dulay, Joshua Muñoz, Jonah Franz R. De Leon, Viktor Rusov, Olha Mishkina, Yihe Wang and Tyler Nguyen for their help with wireframe annotations. We also thank Tolga Birdal, Caner Korkmaz, Hanzhi Chen, Daoyi Gao, and Ilke Demir for helpful discussions and dataset work at the early stage of the paper. Dmytro Mishkin is supported by MPO 60273/24/21300/21000 CEDMO 2.0 NPO project.

{
    \small
    \bibliographystyle{ieeenat_fullname}
    \bibliography{main}

\begin{thebibliography}{36}
\providecommand{\natexlab}[1]{#1}
\providecommand{\url}[1]{\texttt{#1}}
\expandafter\ifx\csname urlstyle\endcsname\relax
  \providecommand{\doi}[1]{doi: #1}\else
  \providecommand{\doi}{doi: \begingroup \urlstyle{rm}\Url}\fi

\bibitem[USM(2024)]{USM3D2024}
1st workshop on urban scene modeling: Where vision meets photogrammetry and graphics.
\newblock In \emph{Proceedings of the IEEE/CVF Conference on Computer Vision and Pattern Recognition Workshops (CVPRW)}, 2024.

\bibitem[Agrawal et~al.(2024)Agrawal, Antoniak, Hanna, Bout, Chaplot, Chudnovsky, Costa, Monicault, Garg, Gervet, Ghosh, Héliou, Jacob, Jiang, Khandelwal, Lacroix, Lample, Casas, Lavril, Scao, Lo, Marshall, Martin, Mensch, Muddireddy, Nemychnikova, Pellat, Platen, Raghuraman, Rozière, Sablayrolles, Saulnier, Sauvestre, Shang, Soletskyi, Stewart, Stock, Studnia, Subramanian, Vaze, Wang, and Yang]{agrawal2024pixtral12b}
Pravesh Agrawal, Szymon Antoniak, Emma~Bou Hanna, Baptiste Bout, Devendra Chaplot, Jessica Chudnovsky, Diogo Costa, Baudouin~De Monicault, Saurabh Garg, Theophile Gervet, Soham Ghosh, Amélie Héliou, Paul Jacob, Albert~Q. Jiang, Kartik Khandelwal, Timothée Lacroix, Guillaume Lample, Diego~Las Casas, Thibaut Lavril, Teven~Le Scao, Andy Lo, William Marshall, Louis Martin, Arthur Mensch, Pavankumar Muddireddy, Valera Nemychnikova, Marie Pellat, Patrick~Von Platen, Nikhil Raghuraman, Baptiste Rozière, Alexandre Sablayrolles, Lucile Saulnier, Romain Sauvestre, Wendy Shang, Roman Soletskyi, Lawrence Stewart, Pierre Stock, Joachim Studnia, Sandeep Subramanian, Sagar Vaze, Thomas Wang, and Sophia Yang.
\newblock Pixtral 12b, 2024.

\bibitem[Anthropic(2024)]{claude35}
AI Anthropic.
\newblock Claude 3.5 sonnet model card addendum.
\newblock \emph{Claude-3.5 Model Card}, 2024.

\bibitem[Araujo et~al.(2021)Araujo, Cao, Askew, Sim, Maggie, Weyand, and Cukierski]{GLD}
Andre Araujo, Bingyi Cao, Cam Askew, Jack Sim, Maggie, Tobias Weyand, and Will Cukierski.
\newblock Google landmark retrieval 2021.
\newblock \url{https://kaggle.com/competitions/landmark-retrieval-2021}, 2021.
\newblock Kaggle.

\bibitem[Avetisyan et~al.(2024)Avetisyan, Xie, Howard-Jenkins, Yang, Aroudj, Patra, Zhang, Frost, Holland, Orme, Engel, Miller, Newcombe, and Balntas]{SceneScript2024}
Armen Avetisyan, Christopher Xie, Henry Howard-Jenkins, Tsun-Yi Yang, Samir Aroudj, Suvam Patra, Fuyang Zhang, Duncan Frost, Luke Holland, Campbell Orme, Jakob Engel, Edward Miller, Richard Newcombe, and Vasileios Balntas.
\newblock Scenescript: Reconstructing scenes with an autoregressive structured language model.
\newblock In \emph{European Conference on Computer Vision (ECCV)}, 2024.

\bibitem[Brachmann et~al.(2024)Brachmann, Wynn, Chen, Cavallari, Monszpart, Turmukhambetov, and Prisacariu]{acezero2024}
Eric Brachmann, Jamie Wynn, Shuai Chen, Tommaso Cavallari, {\'{A}}ron Monszpart, Daniyar Turmukhambetov, and Victor~Adrian Prisacariu.
\newblock Scene coordinate reconstruction: Posing of image collections via incremental learning of a relocalizer.
\newblock In \emph{ECCV}, 2024.

\bibitem[Bradley and Terry(1952)]{bradley1952}
Ralph~A. Bradley and Milton~E. Terry.
\newblock Rank analysis of incomplete block designs: I. the method of paired comparisons.
\newblock \emph{Biometrika}, 39\penalty0 (3/4):\penalty0 324--345, 1952.

\bibitem[Cao et~al.(2023)Cao, Xu, Guo, and Liu]{WireframeNet}
Li Cao, Yike Xu, Jianwei Guo, and Xiaoping Liu.
\newblock Wireframenet: A novel method for wireframe generation from point cloud.
\newblock \emph{Computers and Graphics}, 115:\penalty0 226--235, 2023.

\bibitem[Chen et~al.(2022)Chen, Qian, and Furukawa]{HEAT}
Jiacheng Chen, Yiming Qian, and Yasutaka Furukawa.
\newblock Heat: Holistic edge attention transformer for structured reconstruction.
\newblock In \emph{IEEE Conference on Computer Vision and Pattern Recognition (CVPR)}, 2022.

\bibitem[Cherenkova et~al.(2023)Cherenkova, Dupont, Kacem, Arzhannikov, Gusev, and Aouada]{SepicNet}
Kseniya Cherenkova, Elona Dupont, Anis Kacem, Ilya Arzhannikov, Gleb Gusev, and Djamila Aouada.
\newblock Sepicnet: Sharp edges recovery by parametric inference of curves in 3d shapes.
\newblock In \emph{2023 IEEE/CVF Conference on Computer Vision and Pattern Recognition Workshops (CVPRW)}, pages 2727--2735, 2023.

\bibitem[Elo(1978)]{elo1978}
Arpad~E. Elo.
\newblock \emph{The Rating of Chessplayers, Past and Present}.
\newblock Arco Publishing, 1978.
\newblock Accessed: 2025-03-06.

\bibitem[et~al.(2024{\natexlab{a}})]{gemini}
Gemini~Team et al.
\newblock Gemini: A family of highly capable multimodal models, 2024{\natexlab{a}}.

\bibitem[et~al.(2024{\natexlab{b}})]{openai2024openaio1card}
OpenAI et al.
\newblock Openai o1 system card, 2024{\natexlab{b}}.

\bibitem[Glickman(2013)]{glickman2013}
Mark~E Glickman.
\newblock Introductory note to 1928, 2013.

\bibitem[Hoda{\v{n}} et~al.(2024)Hoda{\v{n}}, Sundermeyer, Labb{\'e}, Nguyen, Wang, Brachmann, Drost, Lepetit, Rother, and Matas]{BOP}
Tom{\'a}{\v{s}} Hoda{\v{n}}, Martin Sundermeyer, Yann Labb{\'e}, Van~Nguyen Nguyen, Gu Wang, Eric Brachmann, Bertram Drost, Vincent Lepetit, Carsten Rother, and Ji{\v{r}}{\'i} Matas.
\newblock {BOP} challenge 2023 on detection, segmentation and pose estimation of seen and unseen rigid objects.
\newblock \emph{Conference on Computer Vision and Pattern Recognition Workshops (CVPRW)}, 2024.

\bibitem[{Hope Akwensi} et~al.(2024){Hope Akwensi}, Bharadwaj, and Wang]{APC2Mesh}
Perpetual {Hope Akwensi}, Akshay Bharadwaj, and Ruisheng Wang.
\newblock Apc2mesh: Bridging the gap from occluded building façades to full 3d models.
\newblock \emph{ISPRS Journal of Photogrammetry and Remote Sensing}, 211:\penalty0 438--451, 2024.

\bibitem[Huang et~al.(2024)Huang, Wang, Guo, and Yang]{Huang_2024_CVPR}
Shangfeng Huang, Ruisheng Wang, Bo Guo, and Hongxin Yang.
\newblock Pbwr: Parametric-building-wireframe reconstruction from aerial lidar point clouds.
\newblock In \emph{Proceedings of the IEEE/CVF Conference on Computer Vision and Pattern Recognition (CVPR)}, pages 27778--27787, 2024.

\bibitem[Hui et~al.(2024)Hui, Yang, Cui, Yang, Liu, Zhang, Liu, Zhang, Yu, Dang, et~al.]{qwen25}
Binyuan Hui, Jian Yang, Zeyu Cui, Jiaxi Yang, Dayiheng Liu, Lei Zhang, Tianyu Liu, Jiajun Zhang, Bowen Yu, Kai Dang, et~al.
\newblock Qwen2. 5-coder technical report.
\newblock \emph{arXiv preprint arXiv:2409.12186}, 2024.

\bibitem[Inc.(2022)]{AppleRoomPlan}
Apple Inc.
\newblock Roomplan: Create 3d floor plans with iphone and ipad.
\newblock \url{https://developer.apple.com/augmented-reality/roomplan/}, 2022.

\bibitem[Jin et~al.(2020)Jin, Mishkin, Mishchuk, Matas, Fua, Yi, and Trulls]{IMC}
Yuhe Jin, Dmytro Mishkin, Anastasiia Mishchuk, Jiri Matas, Pascal Fua, Kwang~Moo Yi, and Eduard Trulls.
\newblock {Image Matching across Wide Baselines: From Paper to Practice}.
\newblock \emph{International Journal of Computer Vision}, 2020.

\bibitem[Kristan et~al.(2016)Kristan, Matas, Leonardis, Vojir, Pflugfelder, Fernandez, Nebehay, Porikli, and \v{C}ehovin]{VOT}
Matej Kristan, Jiri Matas, Ale\v{s} Leonardis, Tomas Vojir, Roman Pflugfelder, Gustavo Fernandez, Georg Nebehay, Fatih Porikli, and Luka \v{C}ehovin.
\newblock A novel performance evaluation methodology for single-target trackers.
\newblock \emph{IEEE Transactions on Pattern Analysis and Machine Intelligence}, 38\penalty0 (11):\penalty0 2137--2155, 2016.

\bibitem[Langerman et~al.(2024)Langerman, Korkmaz, Chen, Gao, Demir, Mishkin, and Birdal]{S23DR}
Jack Langerman, Caner Korkmaz, Hanzhi Chen, Daoyi Gao, Ilke Demir, Dmytro Mishkin, and Tolga Birdal.
\newblock S23dr competition at 1st workshop on urban scene modeling @ cvpr 2024.
\newblock \url{https://huggingface.co/usm3d}, 2024.

\bibitem[Leal-Taix\'{e} et~al.(2015)Leal-Taix\'{e}, Milan, Reid, Roth, and Schindler]{MOT}
L. Leal-Taix\'{e}, A. Milan, I. Reid, S. Roth, and K. Schindler.
\newblock {MOTC}hallenge 2015: {T}owards a benchmark for multi-target tracking.
\newblock \emph{arXiv:1504.01942 [cs]}, 2015.
\newblock arXiv: 1504.01942.

\bibitem[Liu et~al.(2021)Liu, D'Aronco, Schindler, and Wegner]{liu2021_pc2wf}
Yujia Liu, Stefano D'Aronco, Konrad Schindler, and Jan~Dirk Wegner.
\newblock Pc2wf: 3d wireframe reconstruction from raw point clouds.
\newblock In \emph{International Conference on Learning Representations}, 2021.

\bibitem[Luo et~al.(2022)Luo, Ren, Zhe, Kang, Xu, Wonka, and Bao]{luo2022LC2WF}
Yicheng Luo, Jing Ren, Xuefei Zhe, Di Kang, Yajing Xu, Peter Wonka, and Linchao Bao.
\newblock Lc2wf:learning to construct 3d building wireframes from 3d line clouds.
\newblock In \emph{Proceedings of the British Machine Vision Conference (BMVC)}, 2022.

\bibitem[Ma et~al.(2022)Ma, Tan, Xue, Wu, Zheng, and Xia]{ma2022HoW3D}
Wenchao Ma, Bin Tan, Nan Xue, Tianfu Wu, Xianwei Zheng, and Gui-Song Xia.
\newblock How-3d: Holistic 3d wireframe perception from a single image.
\newblock In \emph{International Conference on 3D Vision}, 2022.

\bibitem[Menze and Geiger(2015)]{KITTY}
Moritz Menze and Andreas Geiger.
\newblock Object scene flow for autonomous vehicles.
\newblock In \emph{Conference on Computer Vision and Pattern Recognition (CVPR)}, 2015.

\bibitem[OpenRouter(2024)]{openrouter}
OpenRouter.
\newblock Openrouter: Unified api for ai models, 2024.
\newblock Accessed: 2025-03-07.

\bibitem[Oquab et~al.(2023)Oquab, Darcet, Moutakanni, Vo, Szafraniec, Khalidov, Fernandez, Haziza, Massa, El-Nouby, Howes, Huang, Xu, Sharma, Li, Galuba, Rabbat, Assran, Ballas, Synnaeve, Misra, Jegou, Mairal, Labatut, Joulin, and Bojanowski]{oquab2023dinov2}
Maxime Oquab, Timothée Darcet, Theo Moutakanni, Huy~V. Vo, Marc Szafraniec, Vasil Khalidov, Pierre Fernandez, Daniel Haziza, Francisco Massa, Alaaeldin El-Nouby, Russell Howes, Po-Yao Huang, Hu Xu, Vasu Sharma, Shang-Wen Li, Wojciech Galuba, Mike Rabbat, Mido Assran, Nicolas Ballas, Gabriel Synnaeve, Ishan Misra, Herve Jegou, Julien Mairal, Patrick Labatut, Armand Joulin, and Piotr Bojanowski.
\newblock Dinov2: Learning robust visual features without supervision, 2023.

\bibitem[Radenovi\'{c} et~al.(2018)Radenovi\'{c}, Iscen, Tolias, Avrithis, and Chum]{rOxford5k}
Filip Radenovi\'{c}, Ahmet Iscen, Giorgos Tolias, Yannis Avrithis, and Ond{\v{r}}ej Chum.
\newblock Revisiting oxford and paris: Large-scale image retrieval benchmarking.
\newblock In \emph{CVPR}, 2018.

\bibitem[Russakovsky et~al.(2015)Russakovsky, Deng, Su, Krause, Satheesh, Ma, Huang, Karpathy, Khosla, Bernstein, Berg, and Fei-Fei]{ImageNet}
Olga Russakovsky, Jia Deng, Hao Su, Jonathan Krause, Sanjeev Satheesh, Sean Ma, Zhiheng Huang, Andrej Karpathy, Aditya Khosla, Michael Bernstein, Alexander~C. Berg, and Li Fei-Fei.
\newblock {ImageNet Large Scale Visual Recognition Challenge}.
\newblock \emph{International Journal of Computer Vision (IJCV)}, 115\penalty0 (3):\penalty0 211--252, 2015.

\bibitem[Sanfeliu and Fu(1983)]{sanfeliu1983distance_ged}
Alberto Sanfeliu and King-Sun Fu.
\newblock A distance measure between attributed relational graphs for pattern recognition.
\newblock \emph{IEEE Transactions on Systems, Man, and Cybernetics}, SMC-13\penalty0 (3):\penalty0 353--362, 1983.

\bibitem[Wang et~al.(2023)Wang, Huang, and Yang]{Building3D}
Ruisheng Wang, Shangfeng Huang, and Hongxin Yang.
\newblock { Building3D: An Urban-Scale Dataset and Benchmarks for Learning Roof Structures from Point Clouds }.
\newblock In \emph{ICCV}, pages 20019--20029, Los Alamitos, CA, USA, 2023. IEEE Computer Society.

\bibitem[xAI(2024)]{grok2}
xAI.
\newblock Grok-2: Advancements in ai language modeling.
\newblock 2024.
\newblock Accessed: 2025-03-07.

\bibitem[Zermelo(1929)]{zermelo1929}
Ernst Zermelo.
\newblock Die berechnung der turnier-ergebnisse als ein maximumproblem der wahrscheinlichkeitsrechnung.
\newblock \emph{Mathematische Zeitschrift}, 29:\penalty0 436--460, 1929.

\bibitem[Zhou et~al.(2019)Zhou, Qi, and Ma]{zhou2019end}
Yichao Zhou, Haozhi Qi, and Yi Ma.
\newblock End-to-end wireframe parsing.
\newblock In \emph{ICCV 2019}, 2019.

\end{thebibliography}
}
\clearpage
\setcounter{page}{1}
\maketitlesupplementary
\section{Ranking and Annotator workflow}
\label{sec:additional}
Annotators were selected from the same pool of 3D modelers who created the ground truth wireframes. 
The experiment is conducted through their employer, and compensated at their usual fair hourly rate.

Most raters rate a few more pairs than minimally required 3500 because of the self-consistency checks. 
In prior experiments, we observed a drop-off in self-consistency after around 400 pairs. %
Therefore, we inserted a following pop-up every 350 pairs:
“Time for a Break! Taking regular breaks helps maintain rating quality. We recommend: Stand up and stretch, Look away from the screen, Take a short walk if possible.” We found this to be effective in maintaining rating quality. 
One rater completed all ratings (over 3500) in under 7 hours.  All other raters spread the task over two or more (up to 27) days. Mean and median ratings per day are 1327 and 207, respectively.

\section{Ranking triplets}
\label{sec:other-ranks}
\begin{figure}[tb]
    \centering
    \includegraphics[width=0.8\linewidth]{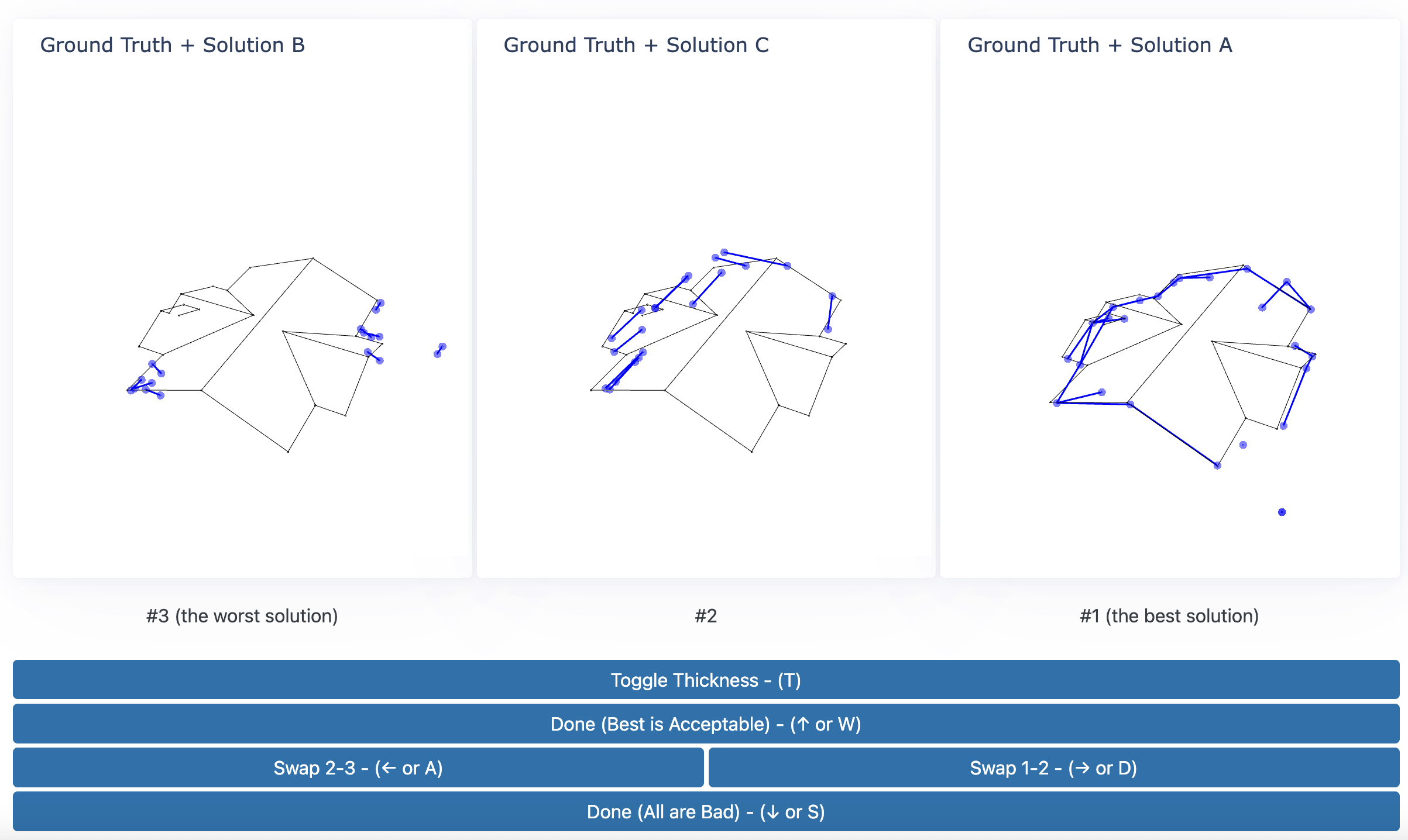}
    \caption{First iteration wireframe ranking interface for human annotators }
    \label{fig:ranking-app-old}
\end{figure}

We have experimented with different forms of rankind the reconstructions, one of which is shown in Figure~\ref{fig:ranking-app-old}. The annotators were sorting triplets, and also marking if the best solution is acceptable. 
However, it took people much longer time to process one triplet, and in addition to that, the ranks  2-3 was much less reliable, than 1-2. In the end, we opted for simplicity and also added a message asking annotators to take a break after each 350 pairs. 

\section{Length Weighted Spectral Graph Distances} incorporate both topological and geometric information by framing graph (wireframe) distance in terms of distances between the spectra of weighted graph Laplacians. We measure the spectral distance using the 2-Wasserstein metric between the eigenvalue distributions:
\begin{equation}
    \LWSGD(G_1, G_2) := W_2(\lambda(L_1), \lambda(L_2))
\end{equation}
where $\lambda(L)$ denotes the spectrum of the Laplacian $L$.

For a graph $G = (V, E)$, the weighted graph Laplacian is defined:
\begin{equation}
     L := D - A
\end{equation}
where $D$ is the weighted degree matrix ($|V| \times |V|$ diagonal matrix with each diagonal entry containing the sum of the lengths of edges incident to that vertex), and $A$ is the weighted adjacency matrix ($|V| \times |V|$ with $A_{ij} = \|V_i - V_j\|_2$ iff $(i,j) \in E$ and $0$ otherwise).

\section{Full agreement tables}
\begin{figure}[tb]
    \centering
    \includegraphics[width=0.9\linewidth]{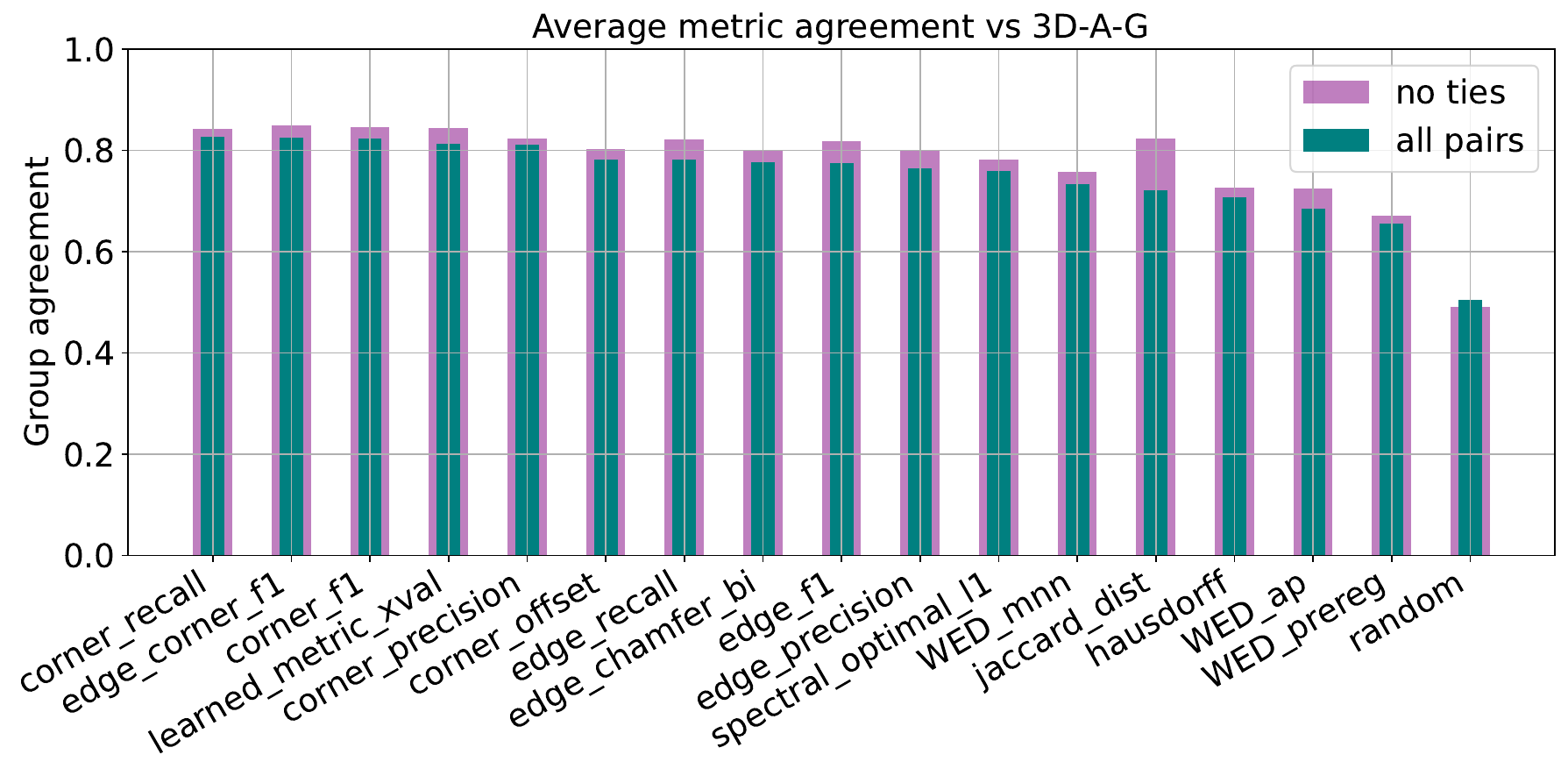}\\
    \includegraphics[width=0.9\linewidth]{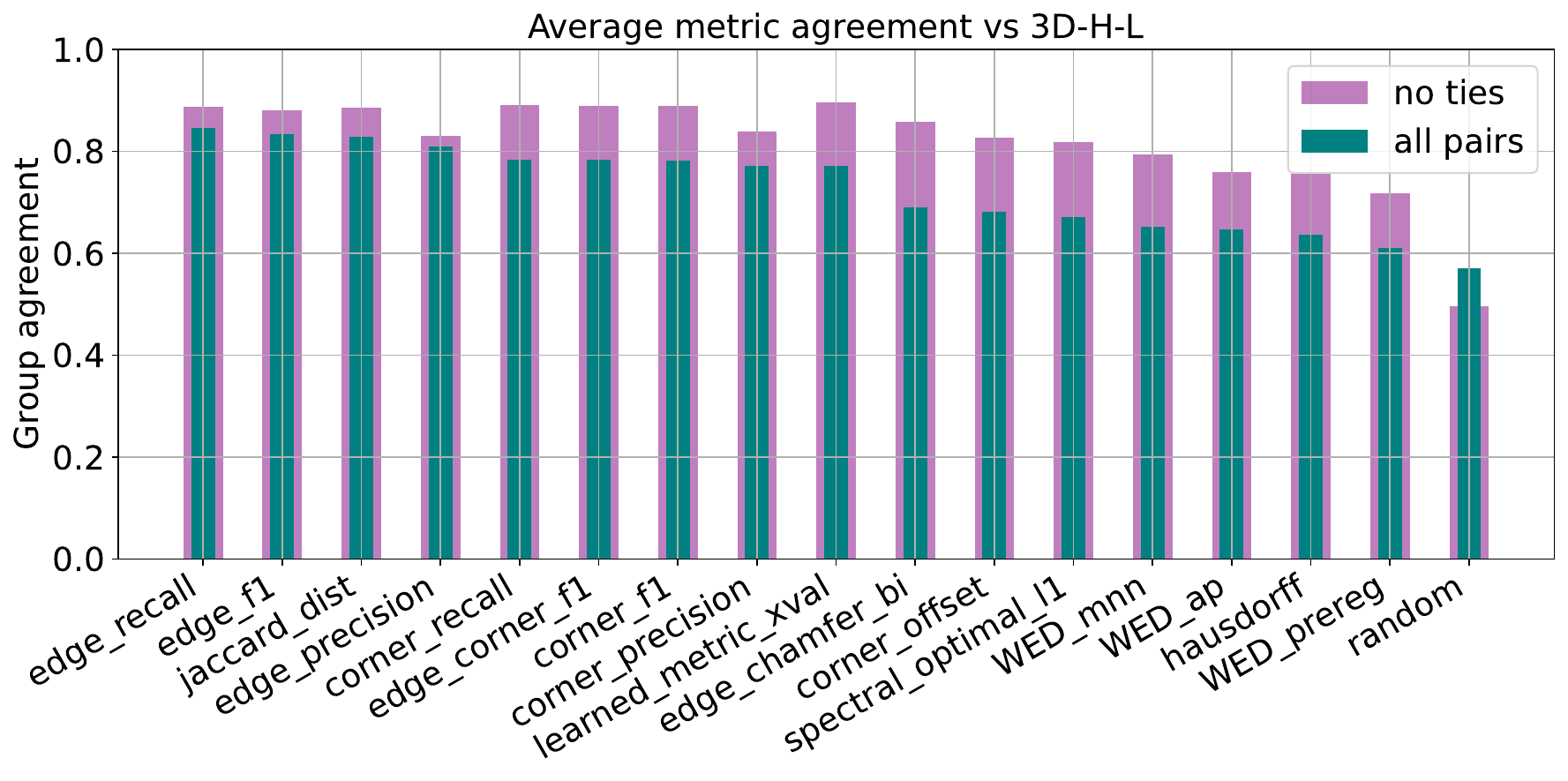}\\

    \includegraphics[width=0.9\linewidth]{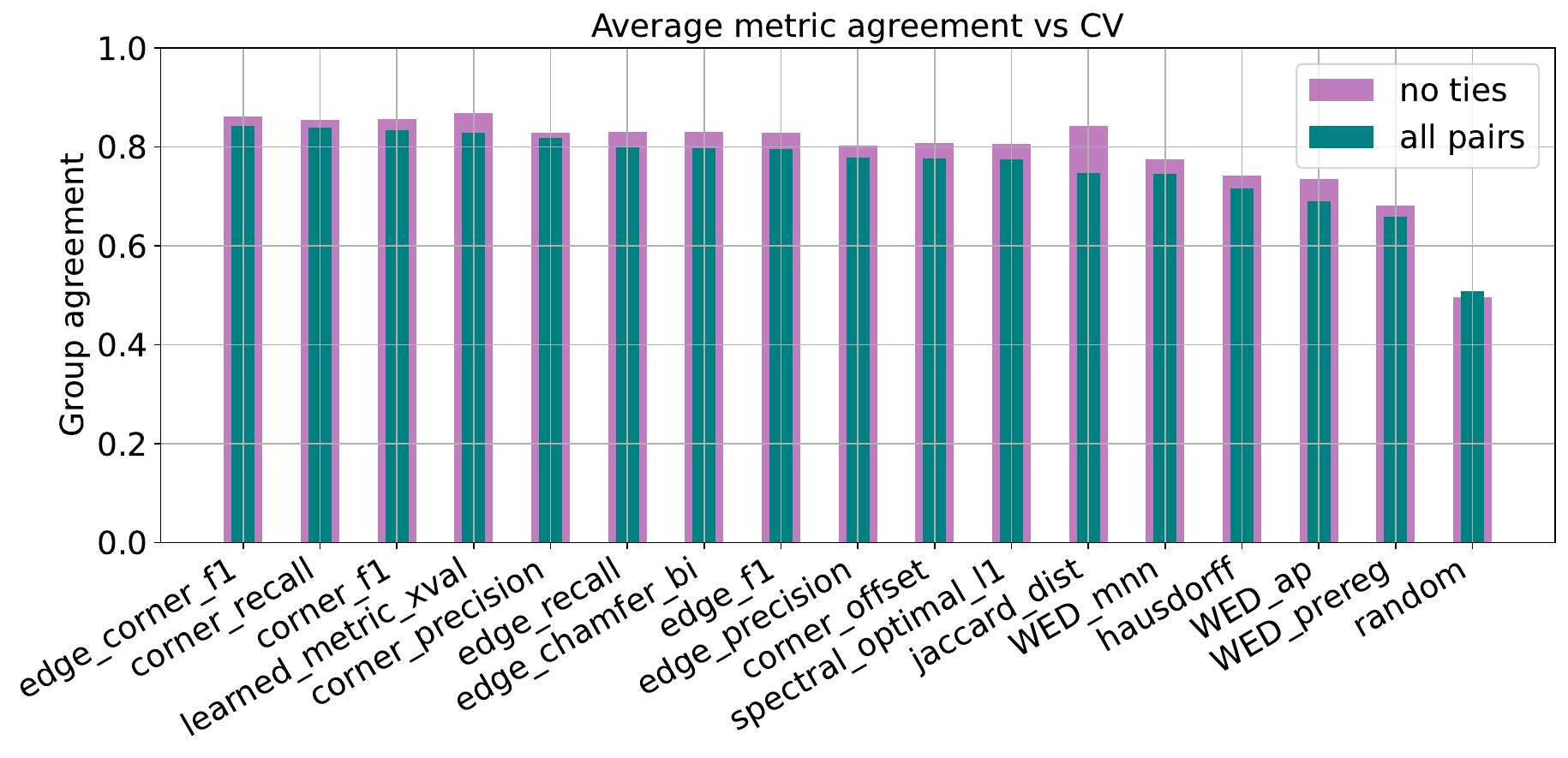}\\
    \includegraphics[width=0.9\linewidth]{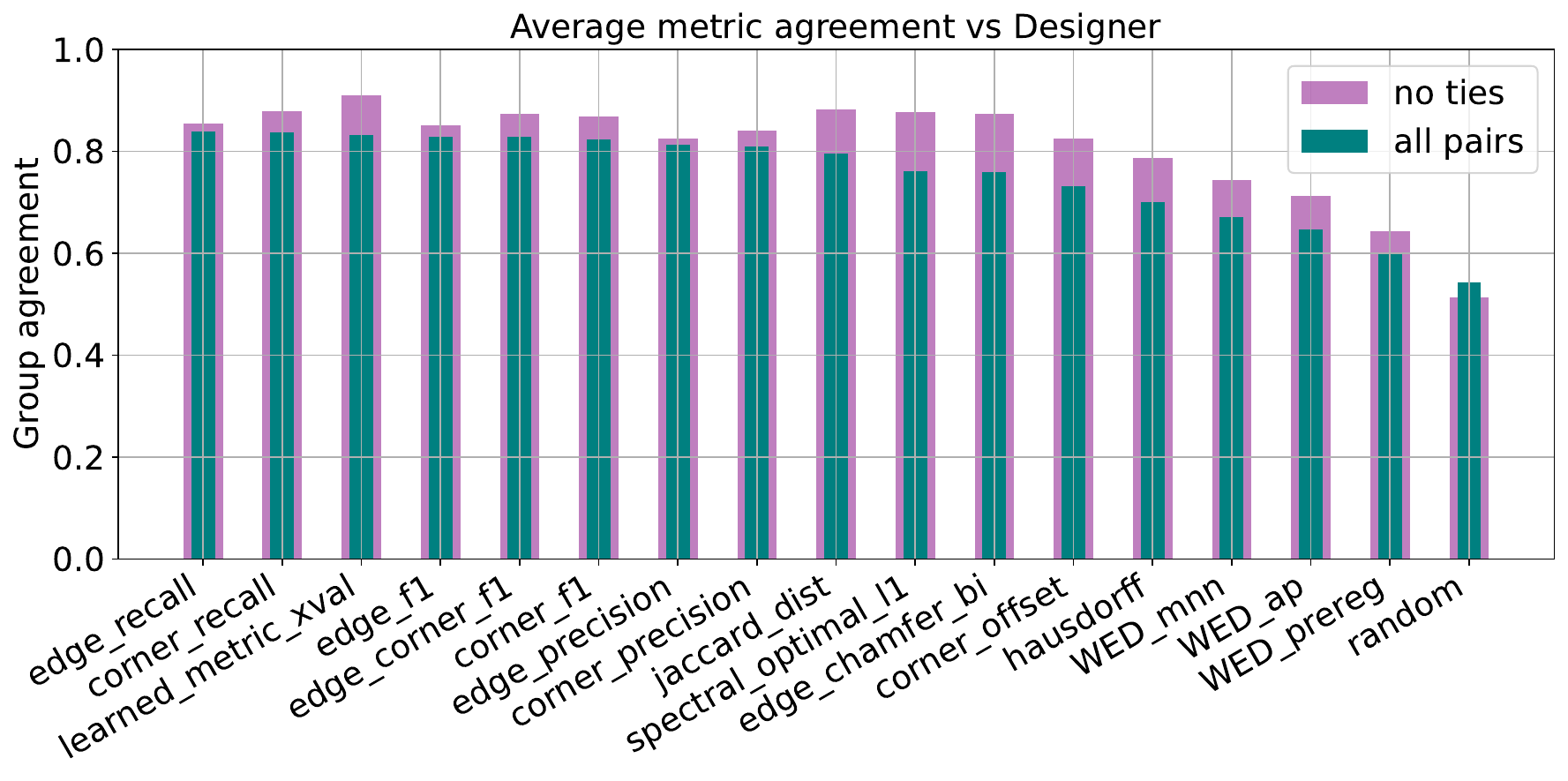}\\

    \caption{Metric ranking by agreement with group in average. From top to bottom: group of raters 1 with more attention to vertices, group of raters 2 with more attention to edges, computer vision engineers}
    \label{fig:supp-metrics-ranking}
\end{figure}

\begin{figure*}[tb]
    \centering
    \includegraphics[width=0.98\linewidth]{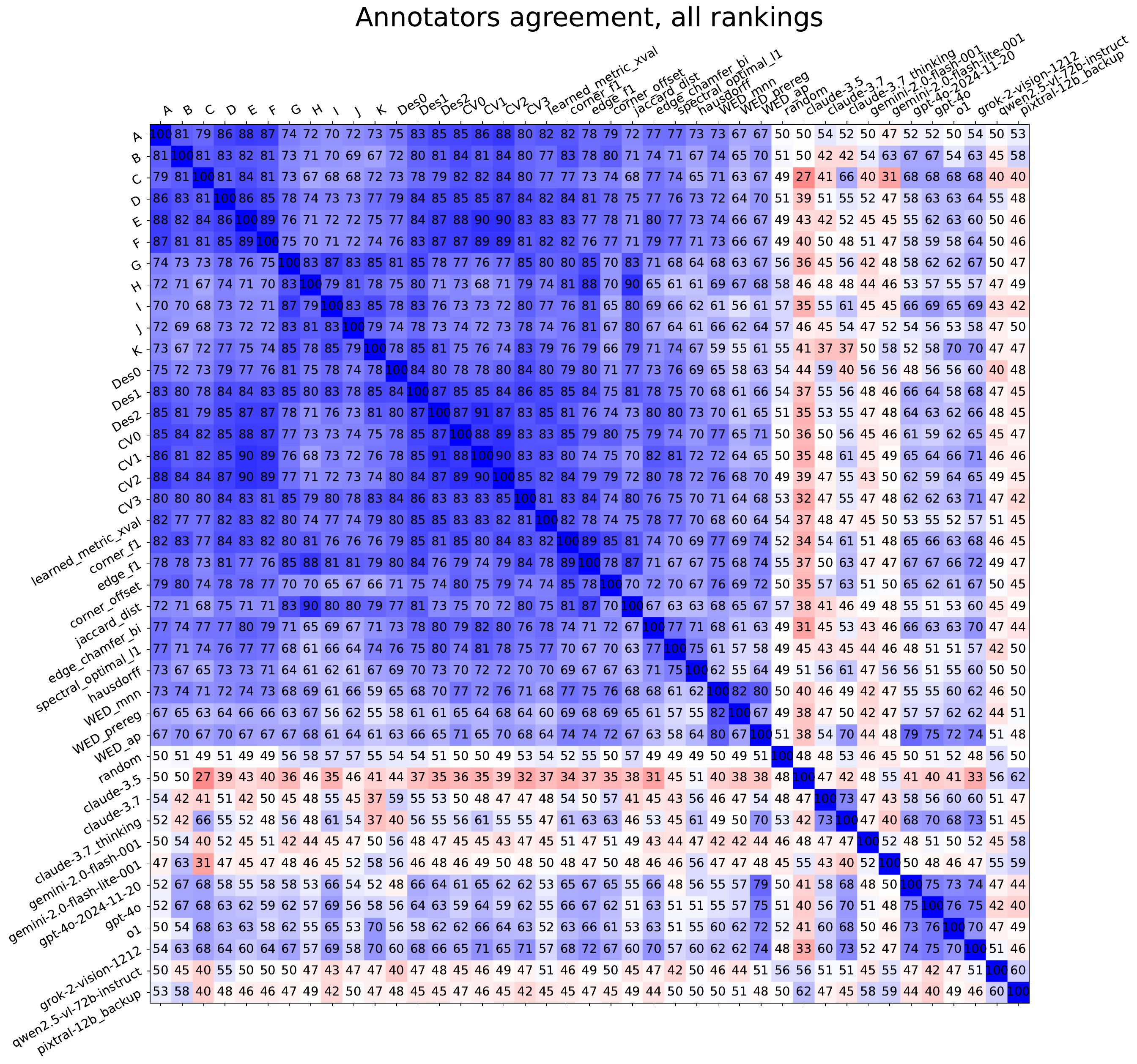}\\
    \caption{Annotator-metrics-LLM agreement (all pairs). 
    Annotators background: A-K -- 3D modellers, Des[0-2] - designers, CV[0-3] - computer vision engineers. Best zoom-in.}
    \label{fig:supp-human-agreement-all}
\end{figure*}

\begin{figure*}[tb]
    \centering
    \includegraphics[width=0.98\linewidth]{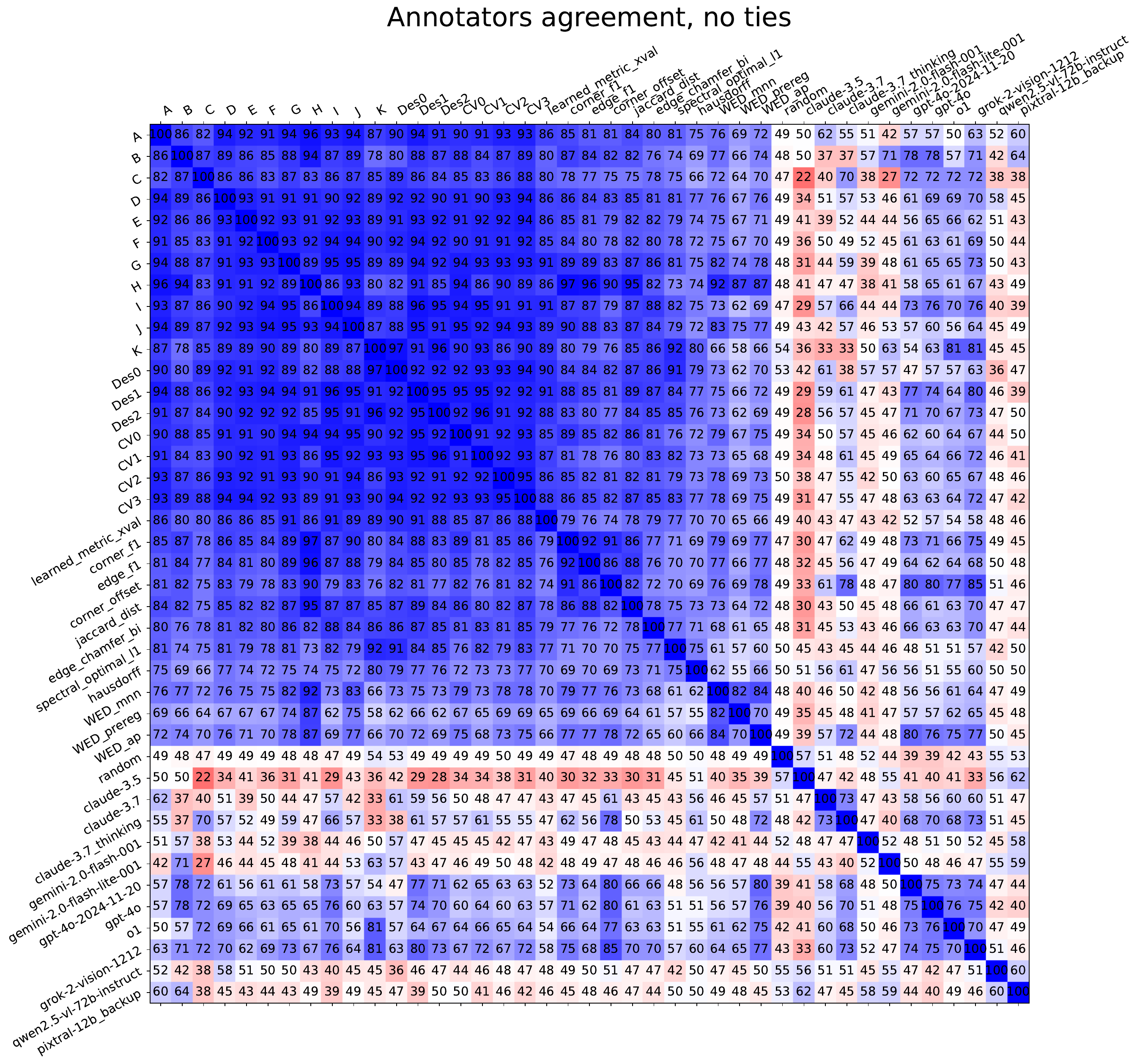}\\
    \caption{Annotator-metrics-LLM agreement (excluding ties pairs). 
annotators agreement to each other, handcrafted metrics, and visual language models.     Annotators background: A-K -- 3D modellers, Des[0-2] - designers, CV[0-3] - computer vision engineers. Best zoom-in.}
    \label{fig:supp-human-agreement-conf}
\end{figure*}

The full annotator-metric-VLM agreement table is shown in Figures~\ref{fig:supp-human-agreement-all},~\ref{fig:supp-human-agreement-conf}, and the metric rankings are shown in Figure~\ref{fig:supp-metrics-ranking}.

\section{VLM Prompts}
All the VLMs are prompted with the following text. 
\begin{quote}
\footnotesize
\texttt{Here we see two possible wireframe reconstructions of houses (shown in blue) \\
superimposed on top of the ground truth wireframe (shown in black). \\
Please describe the quality of each of the two reconstructions (Left and Right). \\
If you don't see any blue lines it is because the reconstruction is incomplete. \\
Which reconstruction most closely matches the ground truth \\
(end by printing the final answer in all caps: "LEFT" or "RIGHT")?}
\end{quote}

\end{document}